\documentclass[10pt,journal,compsoc]{IEEETran}
\usepackage[utf8]{inputenc}
\usepackage{times}
\usepackage{epsfig}
\usepackage{graphicx}
\usepackage{amsmath}
\usepackage{amssymb}
\usepackage{microtype}
\usepackage{subcaption}
\usepackage{booktabs}
\usepackage{placeins} 
\usepackage{xspace}
\usepackage{tabularx}
\usepackage{rotating}
\usepackage[dvipsnames]{xcolor}
\usepackage{tikz}
\usetikzlibrary{arrows.meta}
\usepackage{xtab, afterpage}
\usepackage{comment}
\usepackage[shortlabels]{enumitem}
\usepackage[backend=biber]{biblatex}

\makeatletter
\DeclareRobustCommand\onedot{\futurelet\@let@token\@onedot}
\def\@onedot{\ifx\@let@token.\else.\null\fi\xspace}

\newcommand{\R}{\ensuremath{\mathbb{R}}}

\def\eg{\emph{e.g}\onedot} 
\def\ie{\emph{i.e}\onedot} 
\def\cf{\emph{c.f}\onedot} 
 
\def\wrt{w.r.t\onedot} 
\def\etal{\emph{et al}\onedot}
\makeatother

\usepackage[]{hyperref}
\usepackage[capitalise]{cleveref}

\crefname{figure}{Fig.}{Fig.}
\title{Review of Serial and Parallel Min-Cut/Max-Flow Algorithms for Computer Vision}

\author{Patrick~M.~Jensen,~Niels~Jeppesen,~Anders~B.~Dahl,~and~Vedrana~A.~Dahl%
\IEEEcompsocitemizethanks{%
\IEEEcompsocthanksitem P. M. Jensen, N. Jeppesen, A. B. Dahl, and V. A. Dahl are with the Department of Applied Mathematics and Computer Science, Technical University of Denmark, Kongens Lyngby, Denmark.\protect\\
E-mail: \{patmjen, niejep, abda, vand\}@dtu.dk}}

\newcommand{\maxflow}{min-cut/max-flow\xspace}
\newcommand{\mincut}{\maxflow}
\newcommand{\Maxflow}{Min-cut/max-flow\xspace}
\newcommand{\Mincut}{\Maxflow}

\definecolor{darkgreen}{rgb}{0,0.5,0}
\newcommand{\newtext}[1]{{#1}} 

\begin{document}


\newcommand{\bbox}{\path (-0.7*\s, -3.3 *\s) rectangle (10.5*\s, 3.85*\s);}
\newcommand{\bboxtop}{\path (-0.7*\s, -3.3 *\s) rectangle (10.5*\s, 3.8*\s);}
\newcommand{\bboxelse}{\path (-0.7*\s, -3.0 *\s) rectangle (10.5*\s, 3.5*\s);}

\definecolor{heightgreen}{RGB}{0, 136, 53}
\definecolor{excessred}{RGB}{232, 63, 72}
\definecolor{deficitblue}{RGB}{47, 62, 234}
\definecolor{residualcol}{RGB}{121, 35, 142}
\definecolor{terminalcol}{RGB}{0, 136, 53}
\definecolor{floworange}{RGB}{252, 118, 52}
\definecolor{nodegray}{RGB}{191, 191, 191}

\newcommand{\fflow}[2]{
	\ifnum0=#1\relax$#2$ 
	\else${\color{floworange}#1/}#2$  
	\fi}

\definecolor{drawiogreen}{HTML}{D2F6E6}
\definecolor{drawiored}{HTML}{EBCCCC}
\definecolor{drawioblue}{HTML}{D5D8FB}
\definecolor{drawiogul}{HTML}{FDF6DB}


\tikzstyle{vertex} = [draw=black, circle, minimum size=8pt, inner sep=0pt, fill=nodegray]
\tikzstyle{vertexindex} = [vertex, minimum size=10pt, inner sep=0pt, font=\boldmath \footnotesize, text=white]
\tikzstyle{vertexterminal} = [vertex, fill=terminalcol, text=white, minimum size=13pt, font=\boldmath]
\tikzstyle{edge} = [->, thick, >={Latex[length=2mm]}] 
\tikzstyle{edgeres} = [edge, draw=residualcol, fill=residualcol]


\def\lw{10pt}
\tikzstyle{edgeshaded} = [->, line width = \lw, >={Triangle[width=\lw, length=\lw]}, shorten >=-0.5pt]	

\tikzstyle{edgeset} = [line width = 25pt, rounded corners = 5 pt, line cap = round]	
\tikzstyle{weight} = [draw=none, fill=white,  inner sep=0.5pt, pos=0.45]
\tikzstyle{weightres} = [weight, text=residualcol, outer sep=1pt, pos=0.5]
\tikzstyle{kosi} = [inner sep=1pt, outer sep=0.5pt]
\tikzstyle{vertexexcess} = [vertex, fill=excessred, text=white, minimum size=10pt, font=\boldmath \scriptsize]
\tikzstyle{height} = [draw=heightgreen, thick, circle, text=heightgreen, minimum size=10pt, font=\boldmath \footnotesize, inner sep = 0pt]
\def\shading{0.1}

\IEEEtitleabstractindextext{%
\begin{abstract}
   \newtext{Minimum cut/maximum flow} (\mincut) algorithms solve a variety of problems in computer vision and thus significant effort has been put into developing fast \mincut algorithms. As a result, it is difficult to choose an ideal \newtext{algorithm} for a given problem. Furthermore, parallel algorithms have not been thoroughly compared. In this paper, we evaluate the state-of-the-art serial and parallel \mincut algorithms on the largest set of computer vision problems yet. We focus on \newtext{generic} algorithms\newtext{, \ie,} for unstructured graphs, but also compare with the specialized GridCut implementation. When applicable, GridCut performs best. Otherwise,
   \newtext{the two pseudoflow algorithms, Hochbaum pseudoflow and excesses incremental breadth first search, achieves the overall best performance.}
   The most memory efficient implementation tested is the Boykov-Kolmogorov algorithm. Amongst generic parallel algorithms, we find the bottom-up merging approach by Liu and Sun to be best, but no method is dominant. Of the generic parallel methods, only the parallel preflow push-relabel algorithm is able to efficiently scale with many processors across problem sizes, and no generic parallel method consistently outperforms serial algorithms. Finally, we provide and evaluate strategies for algorithm selection to obtain good expected performance. We make our dataset and implementations publicly available for further research.
\end{abstract}

}

\maketitle

\IEEEdisplaynontitleabstractindextext
\IEEEpeerreviewmaketitle

\IEEEraisesectionheading{\section{Introduction}\label{sec:introduction}}
\IEEEPARstart{M}{in-cut/max-flow} algorithms are ubiquitous in computer vision, since a large variety of computer vision problems can be formulated as \maxflow problems. Example applications include image segmentation \cite{Boykov2004, Chen2018, Isack2017, Ishikawa2003, Jeppesen2020, Peng2013}, stereo matching \cite{Boykov1998,Kolmogorov2001}, surface reconstruction \cite{Lempitsky2007}, surface fitting \cite{Egger2010, Kashyap2017, Lee2020, Li2005, Wang2010, Yin2010}, graph matching \cite{Hutschenreiter2021}, and texture restoration \cite{Rother2007}. In recent years, \maxflow algorithms have also found use in conjunction with deep learning methods --- for example, to quickly generate training labels \cite{Khoreva2017} or in combination with convolutional neural networks (CNNs) \cite{Guo2018, Minaee2021, Tajbakhsh2020}.

Greig \etal \cite{Greig1989} were the first to use \maxflow algorithms to solve maximum a posterior Markov random field (MRF) problems in computer vision. Later, \newtext{Boykov and Jolly} \cite{Boykov2001} showed how this could be generalized and \newtext{Boykov and Kolmogorov} \cite{Boykov2004} proposed a fast \maxflow algorithm for computer vision problems. \Maxflow algorithms in computer vision are used to solve a large family of energy minimization problems, and the most commonly used energy function is of the form
\begin{equation}
    \mathcal{E}(\textbf{x}) = \sum_{i \in \mathcal{P}} \mathcal{E}_i(x_i) + \sum_{(i,j) \in \mathcal{N}} \mathcal{E}_{ij}(x_i, x_j),
    \label{Eq:energy}
\end{equation}
where $\mathcal{P}$ is a set of \newtext{indices for the binary variables $x_i \in \{0, 1\}$}, and $\mathcal{N}$ is a set of \newtext{index} pairs. A unary term \newtext{$\mathcal{E}_i : \{ 0, 1 \} \to \R$} is associated with variable $x_i$, and a pairwise term \newtext{$\mathcal{E}_{ij} : \{ 0, 1\}^2 \to \R$} is associated with the pair of variables $x_i, x_j$. \newtext{As the inputs to the energy terms are binary, the terms are often represented as lookup tables.} In a typical application, such as binary segmentation with MRFs \cite{Boykov2006}, $\mathcal{P}$ represents pixels in an image and $x_i$ represents the assignment of pixel $i$. However, variables can also describe more abstract things \newtext{\cite{Egger2010, Hutschenreiter2021, Jeppesen2020, Li2005, Wang2010, Wu2002}, \eg, candidate positions for mesh vertices}.

For energy functions which are \emph{submodular}, meaning that all pairwise energy terms satisfy the condition
\begin{equation}
    \mathcal{E}_{ij}(0, 0) + \mathcal{E}_{ij}(1, 1) \leq \mathcal{E}_{ij}(0, 1) + \mathcal{E}_{ij}(1, 0),
    \label{Eq:submodular}
\end{equation}
the minimization can be solved directly as a \maxflow problem \cite{Freedman2005, Kolmogorov2004}. Submodular energies favor neighbors that have the same label, \ie, $(x_i,x_j)$ having the labels $(0,0)$ or $(1,1)$ rather than $(0,1)$ or $(1,0)$. Therefore, submodularity imposes a local smoothness of the solution, which is useful in many computer vision problems. However, some important vision problems are not submodular. In such cases, one can use either a submodular approximation or an approach based on quadratic pseudo-Boolean optimization (QPBO) as described in \cite{Boros1991, Hammer1984, Kolmogorov2007, Rother2007}.

Due to the wide applicability of \maxflow in computer vision, several fast generic \maxflow algorithms have been developed \cite{Boykov2004, Goldberg2015, Hochbaum2008}. In addition, more specialized algorithms have been created that exploit the grid structure of images to reduce memory usage and run time \cite{Delong2008, Gridcut2015, Jamriska2012, Peng2015, Vineet2008}. Furthermore, methods for dynamic problems \cite{Goldberg2015, Kohli2007, Yu2017}, where a series of similar \maxflow problems are solved in succession, have been proposed. For such problems, the result of the first solution can be reused to speed up computations of subsequent solutions. Finally, some papers \cite{Halim2011, Shekhovtsov2013, Strandmark2010, Yu2015} have explored methods that also allow for distributed computation of \maxflow problems across several computational nodes. This approach is primarily suited for graphs too large to fit in physical memory.

In this paper, we focus on generic \maxflow algorithms, which do \emph{not} make assumptions about the graph structure (\eg, requiring a grid structure). However, for comparison, we include the GridCut algorithm \cite{Jamriska2012} in our evaluation on grid-based graphs. Furthermore, we consider only static problems where a solution is calculated once, without access to a previous solution (as opposed to dynamic problems). Finally, \newtext{for parallel algorithms,} we do not consider whether the algorithm works well in a distributed setting, but focus \newtext{on the shared memory case where the complete graph can be loaded into the memory of one machine.}

The goal is that our experimental results can help researchers understand the strengths and weaknesses of the current state-of-the-art \maxflow algorithms and help practitioners when choosing a \maxflow algorithm to use for a given problem.

\subsection{Related Work}

\textbf{Serial Algorithms \enspace}
Several papers \cite{Chandran2009,Fishbain2016,Goldberg2015,Verma2012} provide comparisons of different serial \maxflow algorithms on a variety of standard benchmark problems. However, many of these benchmark problems are small \wrt the scale of \maxflow problems that can be solved today --- especially when it comes to grid graphs. Also, graphs in which nodes are not based on an image grid are severely underrepresented.
Furthermore, \cite{Chandran2009, Fishbain2016, Verma2012} do not include all current state-of-the-art algorithms, while other papers do not include initialization times for the min-cut computation. As shown by Verma and Batra \cite{Verma2012}, it is important for practical use to include the initialization time, as
\newtext{algorithm implementations}
may spend as much time on initialization as on the min-cut computation. Additionally, existing papers only compare reference implementations (\ie, the implementation released by the authors) of algorithms --- the exception being that an optimized version of the BK algorithm is sometimes included, \eg, in \cite{Goldberg2015}. However, as implementation details --- \ie, choices that are left unspecified by the algorithm description --- can significantly impact performance \cite{Verma2012}, a systematic investigation of their effect is also important. Finally, existing comparisons focus on determining the overall best algorithm, even though, as we show in this work,
\newtext{the best algorithm depends on the features of the given graph.}

\textbf{Parallel Algorithms \enspace}
To our knowledge, parallel \maxflow algorithms have not been systematically compared. Papers introducing parallel algorithms only compare with serial algorithms \cite{Liu2010,Strandmark2010,Yu2015} or a single parallel algorithm \cite{Baumstark2015}. The most comprehensive comparison so far was made by Shekhovtsov and Hlav{\'{a}}{\v{c}} \cite{Shekhovtsov2013} who included a generic and grid-based parallel algorithm.
\newtext{However, no paper compares with the approach by Liu and Sun \cite{Liu2010}, as no public implementation is available,}
even though it is expected to be the fastest \cite{Shekhovtsov2013,Strandmark2010}.
\newtext{Additionally}, all papers use the same set of computer vision problems used to benchmark serial algorithms. This is not ideal, as the set lacks larger problems \newtext{which we expect} to benefit the most from parallelization \cite{Jeppesen2021}.
\newtext{Therefore, how big the performance benefits of parallelization are, and when to expect them, is still to be determined.}

\subsection{Contributions}
We evaluate current state-of-the-art generic serial and parallel \maxflow algorithms on the largest set of computer vision problems so far. We compare the algorithms on a wide range of graph problems including commonly used benchmarks problems, as well as many new problem instances from recent papers --- some of which are significantly larger than previous problems and expose weaknesses in the algorithms not seen with previous datasets. Since the performance of the algorithms varies between problems, we also provide concrete strategies on algorithm selection and evaluate the expected performance of these.

For the serial algorithms, we evaluate the reference implementations of the Hochbaum pseudoflow (HPF) \cite{Hochbaum2008,Hochbaum2013}, the preflow push-relabel (PPR) \cite{Goldberg1988}, and the GridCut \cite{Gridcut2015} algorithms. Moreover, to reduce the influence of implementation details, we evaluate different versions (including our own) of the Excesses Incremental Breadth First Search (EIBFS) \cite{Goldberg2015} and the Boykov-Kolmogorov (BK) \cite{Boykov2004} algorithm. We chose these for an extended evaluation, as EIBFS is the most recent \maxflow algorithm and BK is still widely used in the computer vision community.

For the parallel algorithms, we provide the first comprehensive comparison of all major approaches. This includes our own implementation of the bottom-up merging algorithm by Liu and Sun \cite{Liu2010}, our own version of the dual decomposition algorithm by Strandmark and Kahl \cite{Strandmark2010}, the reference implementation of the region discharge algorithm by Shekhovtsov and Hlav{\'a}{\v{c}} \cite{Shekhovtsov2013}, an implementation of the parallel preflow push-relabel algorithm by Baunstark et al. \cite{Baumstark2015}, and the parallel implementation of GridCut (P-GridCut) \cite{Gridcut2015}. In our comparison, we evaluate not just the run time --- including both the initialization time and the time for the \maxflow computations --- but also the memory use of the implementations. Memory usage has not received much attention in the literature, despite it often being a limiting factor when working with large problems. \newtext{Finally, we show that the current parallel algorithm implementations have unpredictable performance and unfortunately often perform worse than serial algorithms.}

All tested C++ implementations (except GridCut \cite{Gridcut2015}), including our new implementations of several algorithms, are available at \url{https://github.com/patmjen/maxflow_algorithms} and are archived at \href{https://doi.org/10.5281/zenodo.4903945}{DOI:10.5281/zenodo.4903945} \cite{Code}. We also provide Python wrapper packages for several of the algorithms (including BK and HPF), which can be found at \url{https://github.com/skielex/shrdr}. All of our benchmark problems are available at \href{https://doi.org/10.11583/DTU.17091101}{DOI:10.11583/DTU.17091101}.

\def\s{0.7} 
\def\wthird{0.45\linewidth} 
\begin{figure}
	\centering
	\begin{subfigure}[b]{\wthird}
	    \setlength\abovecaptionskip{0.2\baselineskip}
		\resizebox{\linewidth}{!}{\begin{tikzpicture}
	
	\bboxtop
	
	\foreach \n/\x/\y in {s/0/0, t/10/0, v1/0.8/1.9, v2/2.7/3, 
		v3/5/3.2, v4/7.3/3, v5/9.2/1.9,
		v6/1.2/-1.9, v7/3.6/-2.7, v8/6.4/-2.7, v9/8.8/-1.9,
		v10/3.8/0.7, v11/6.7/0.7, v12/5.2/-1}{
		\coordinate(\n) at (\x*\s, \y*\s);}
	
	\node[vertexterminal](s) at (s) {$s$};
	\node[vertexterminal](t) at (t) {$t$};
	\foreach \i in{1, ..., 12}{
		\node[vertexindex](v\i) at (v\i) {$\i$};}
		
	\draw[edge] (s)--node[weight]{\fflow{0}{10}} (v1);
	\draw[edge] (v1)--node[weight]{\fflow{0}{20}} (v2);
	\draw[edge] (v2)--node[weight]{\fflow{0}{15}} (v3);
	\draw[edge] (v3)--node[weight]{\fflow{0}{9}} (v4);
	\draw[edge] (v4)--node[weight]{\fflow{0}{8}} (v5);
	\draw[edge] (v5)--node[weight]{\fflow{2}{10}} (t);
	
	\draw[edge] (s)--node[weight]{\fflow{8}{10}} (v6);
	\draw[edge] (v6)--node[weight]{\fflow{3}{3}} (v7);
	\draw[edge] (v7)--node[weight]{\fflow{0}{9}} (v8);
	\draw[edge] (v8)--node[weight]{\fflow{0}{4}} (v9);
	\draw[edge] (v9)--node[weight]{\fflow{6}{10}} (t);
	
	\draw[edge] (v6)--node[weight]{\fflow{5}{15}} (v10);
	\draw[edge] (v7)--node[weight]{\fflow{3}{5}} (v12);
	
	\draw[edge] (v10)--node[weight]{\fflow{0}{7}} (v2);
	\draw[edge] (v10)--node[weight]{\fflow{0}{2}} (v4);
	\draw[edge] (v10)--node[weight]{\fflow{2}{20}} (v11);
	\draw[edge] (v10)--node[weight]{\fflow{3}{6}} (v12);
	
	\draw[edge] (v11)--node[weight]{\fflow{2}{12}} (v5);
	\draw[edge] (v12)--node[weight]{\fflow{6}{6}} (v9);

\end{tikzpicture}}%
		\caption{Graph and flow\label{fig:graphflow}}
	\end{subfigure}\hfil
	\begin{subfigure}[b]{\wthird}
	    \setlength\abovecaptionskip{0.2\baselineskip}
		\resizebox{\linewidth}{!}{\begin{tikzpicture}
	
	\bboxtop
	
	\foreach \n/\x/\y in {s/0/0, t/10/0, v1/0.8/1.9, v2/2.7/3, 
		v3/5/3.2, v4/7.3/3, v5/9.2/1.9,
		v6/1.2/-1.9, v7/3.6/-2.7, v8/6.4/-2.7, v9/8.8/-1.9,
		v10/3.8/0.7, v11/6.7/0.7, v12/5.2/-1}{
		\coordinate(\n) at (\x*\s, \y*\s);}
	
	\node[vertexterminal](s) at (s) {$s$};
	\node[vertexterminal](t) at (t) {$t$};
	\foreach \i in{1, ..., 12}{
		\node[vertex](v\i) at (v\i) {};}
		
\draw[edge] (s)--node[weight]{\fflow{8}{10}} (v1);
\draw[edge] (v1)--node[weight]{\fflow{8}{20}} (v2);
\draw[edge] (v2)--node[weight]{\fflow{8}{15}} (v3);
\draw[edge] (v3)--node[weight]{\fflow{8}{9}} (v4);
\draw[edge] (v4)--node[weight]{\fflow{8}{8}} (v5);
\draw[edge] (v5)--node[weight]{\fflow{10}{10}} (t);

\draw[edge] (s)--node[weight]{\fflow{10}{10}} (v6);
\draw[edge] (v6)--node[weight]{\fflow{3}{3}} (v7);
\draw[edge] (v7)--node[weight]{\fflow{3}{9}} (v8);
\draw[edge] (v8)--node[weight]{\fflow{3}{4}} (v9);
\draw[edge] (v9)--node[weight]{\fflow{8}{10}} (t);

\draw[edge] (v6)--node[weight]{\fflow{7}{15}} (v10);
\draw[edge] (v7)--node[weight]{\fflow{0}{5}} (v12);

\draw[edge] (v10)--node[weight]{\fflow{0}{7}} (v2);
\draw[edge] (v10)--node[weight]{\fflow{0}{2}} (v4);
\draw[edge] (v10)--node[weight]{\fflow{2}{20}} (v11);
\draw[edge] (v10)--node[weight]{\fflow{5}{6}} (v12);

\draw[edge] (v11)--node[weight]{\fflow{2}{12}} (v5);
\draw[edge] (v12)--node[weight]{\fflow{5}{6}} (v9);		
	

\begin{scope}[transparency group, opacity = \shading]
	\draw[heightgreen, edgeset] (s.center)--(v1.center)--(v2.center)--(v3.center)--(v4.center);
	\draw[heightgreen, edgeset] (s.center)--(v2.center)--(v4.center);
	\draw[heightgreen, edgeset] (v1.center)--(v3.center);
	\draw[residualcol, fill, edgeset] (t.center)--(v9.center)--(v8.center)--(v7.center)--(v6.center)--
	(v10.center)--(v11.center)--(v5.center)--cycle;
\end{scope}

\end{tikzpicture}}
		\caption{\Mincut\label{fig:graphcut}}
	\end{subfigure}
	\\[6pt]
	\begin{subfigure}[b]{\wthird}
	    \setlength\abovecaptionskip{0.2\baselineskip}
		\resizebox{\linewidth}{!}{\begin{tikzpicture}
	
	\bboxelse
	
	\foreach \n/\x/\y in {s/0/0, t/10/0, v1/0.8/1.9, v2/2.7/3, 
		v3/5/3.2, v4/7.3/3, v5/9.2/1.9,
		v6/1.2/-1.9, v7/3.6/-2.7, v8/6.4/-2.7, v9/8.8/-1.9,
		v10/3.8/0.7, v11/6.7/0.7, v12/5.2/-1}{
		\coordinate(\n) at (\x*\s, \y*\s);}
	
	\node[vertexterminal](s) at (s) {$s$};
	\node[vertexterminal](t) at (t) {$t$};
	\foreach \i in{1, ..., 12}{
		\node[vertex](v\i) at (v\i) {};}
		
	\draw[edgeres] (s)--node[weightres]{$10$} (v1);
	\draw[edgeres] (v1)--node[weightres]{$20$} (v2);
	\draw[edgeres] (v2)--node[weightres]{$15$} (v3);
	\draw[edgeres] (v3)--node[weightres]{$9$} (v4);
	\draw[edgeres] (v4)--node[weightres]{$8$} (v5);

	\draw[edgeres] (v7)--node[weightres]{$3$} (v6);
	\draw[edgeres] (v7)--node[weightres]{$9$} (v8);
	\draw[edgeres] (v8)--node[weightres]{$4$} (v9);
	
	\draw[edgeres] (v7)--node[weightres]{$5$} (v12);
	\draw[edgeres] (v10)--node[weightres]{$7$} (v2);
	\draw[edgeres] (v10)--node[weightres]{$2$} (v4);

	\draw[edgeres] (v9)--node[weightres]{$6$} (v12);		
	
   \draw[edgeres] (v5.335)--node[weightres, right,kosi]{$8$} (t.115);
	\draw[edgeres] (t.155)--node[weightres, left,kosi]{$2$} (v5.295);
	
 	\draw[edgeres] (s.335)--node[weightres, right,kosi]{$2$} (v6.115);
	\draw[edgeres] (v6.155)--node[weightres, left,kosi]{$8$} (s.295);
	
	\draw[edgeres] (v9.65)--node[weightres, left,kosi]{$4$} (t.205);
	\draw[edgeres] (t.245)--node[weightres, right,kosi]{$6$} (v9.25);
	
	\draw[edgeres] (v10.335)--node[weightres,right,kosi]{$3$} (v12.115);
	\draw[edgeres] (v12.155)--node[weightres,left,kosi]{$3$} (v10.295);
	
	\draw[edgeres] (v6.65)--node[weightres, left,kosi]{$10$} (v10.205);
	\draw[edgeres] (v10.245)--node[weightres, right,kosi]{$5$} (v6.25);
	
	\draw[edgeres] (v10.20)--node[weightres, above]{$18$} (v11.160);
	\draw[edgeres] (v11.200)--node[weightres, below]{$2$} (v10.340);
	
	\draw[edgeres] (v11.50)--node[weightres, above, kosi]{$10$} (v5.190);
	\draw[edgeres] (v5.230)--node[weightres, below, kosi]{$2$} (v11.10);
		
\end{tikzpicture}}
		\caption{Residual graph\label{fig:graphresidual}}
	\end{subfigure}\hfil
	\begin{subfigure}[b]{\wthird}
	    \setlength\abovecaptionskip{0.2\baselineskip}
		\resizebox{\linewidth}{!}{\begin{tikzpicture}
	
	\bboxelse
	
	\foreach \n/\x/\y in {s/0/0, t/10/0, v1/0.8/1.9, v2/2.7/3, 
		v3/5/3.2, v4/7.3/3, v5/9.2/1.9,
		v6/1.2/-1.9, v7/3.6/-2.7, v8/6.4/-2.7, v9/8.8/-1.9,
		v10/3.8/0.7, v11/6.7/0.7, v12/5.2/-1}{
		\coordinate(\n) at (\x*\s, \y*\s);}
	
	\node[vertexterminal](s) at (s) {$s$};
	\node[vertexterminal](t) at (t) {$t$};
	\foreach \i in{1, ..., 12}{
		\node[vertex](v\i) at (v\i) {};}

	\def\lw{4pt}
	\def\shading{0.2} 
	\begin{scope}[transparency group, opacity = \shading]
		\draw[deficitblue, edgeshaded] (s)--(v1);
		\draw[deficitblue, edgeshaded] (v1)--(v2);
		\draw[deficitblue, edgeshaded] (v2)--(v3);
		\draw[deficitblue, edgeshaded] (v3)--(v4);
		\draw[deficitblue, edgeshaded] (v4)--(v5);
		\draw[deficitblue, edgeshaded] (v5.180)--(v11.40);
		\draw[deficitblue, edgeshaded] (v11.160)--(v10.20);
		\draw[deficitblue, edgeshaded] (v10)--(v12);
		\draw[deficitblue, edgeshaded] (v11)--(v10);
		\draw[deficitblue, edgeshaded] (v10)--(v12);
		\draw[deficitblue, edgeshaded] (v12)--(v9);	
		\draw[deficitblue, edgeshaded] (v9)--(t);
	\end{scope}

	\begin{scope}[transparency group, opacity = \shading]
		\draw[floworange, edgeshaded] (v6)--(v10);
		\draw[orange, edgeshaded] (v10.340)--(v11.200);
		\draw[floworange, edgeshaded] (v11.0)--(v5.220);
	\end{scope}

	\begin{scope}[transparency group, opacity = 2*\shading]
	\draw[excessred, edgeshaded] (s)--(v6);
	\draw[excessred, edgeshaded] (v5)--(t);
	\end{scope}

	\draw[edge] (s)--node[weight]{\fflow{0}{10}} (v1);
	\draw[edge] (v1)--node[weight]{\fflow{0}{20}} (v2);
	\draw[edge] (v2)--node[weight]{\fflow{0}{15}} (v3);
	\draw[edge] (v3)--node[weight]{\fflow{0}{9}} (v4);
	\draw[edge] (v4)--node[weight]{\fflow{0}{8}} (v5);
	\draw[edge] (v5)--node[weight]{${\color{excessred}10/}10$} (t);
	
	\draw[edge] (s)--node[weight]{${\color{excessred}10/}10$} (v6);
	\draw[edge] (v6)--node[weight]{\fflow{0}{3}} (v7);
	\draw[edge] (v7)--node[weight]{\fflow{0}{9}} (v8);
	\draw[edge] (v8)--node[weight]{\fflow{0}{4}} (v9);
	\draw[edge] (v9)--node[weight]{\fflow{0}{10}} (t);
	
	\draw[edge] (v6)--node[weight]{\fflow{10}{15}} (v10);
	\draw[edge] (v7)--node[weight]{\fflow{0}{5}} (v12);
	
	\draw[edge] (v10)--node[weight]{\fflow{0}{7}} (v2);
	\draw[edge] (v10)--node[weight]{\fflow{0}{2}} (v4);
	\draw[edge] (v10)--node[weight]{\fflow{10}{20}} (v11);
	\draw[edge] (v10)--node[weight]{\fflow{0}{6}} (v12);
	
	\draw[edge] (v11)--node[weight]{\fflow{10}{12}} (v5);
	\draw[edge] (v12)--node[weight]{\fflow{0}{6}} (v9);		
	
	\def\shading{0.05} 
	\begin{scope}[transparency group, opacity = \shading]
	\draw[deficitblue, edgeshaded] (s)--(v1);
	\draw[deficitblue, edgeshaded] (v1)--(v2);
	\draw[deficitblue, edgeshaded] (v2)--(v3);
	\draw[deficitblue, edgeshaded] (v3)--(v4);
	\draw[deficitblue, edgeshaded] (v4)--(v5);
	\draw[deficitblue, edgeshaded] (v5.180)--(v11.40);
	\draw[deficitblue, edgeshaded] (v11.160)--(v10.20);
	\draw[deficitblue, edgeshaded] (v10)--(v12);
	\draw[deficitblue, edgeshaded] (v11)--(v10);
	\draw[deficitblue, edgeshaded] (v10)--(v12);
	\draw[deficitblue, edgeshaded] (v12)--(v9);	
	\draw[deficitblue, edgeshaded] (v9)--(t);
	\end{scope}

	\def\shading{0.1} 
	\begin{scope}[transparency group, opacity = \shading]
	\draw[floworange, edgeshaded] (v6)--(v10);
	\draw[floworange, edgeshaded] (v10.340)--(v11.200);
	\draw[floworange, edgeshaded] (v11.0)--(v5.220);
	\end{scope}
	
	\def\shading{0.3} 
	\begin{scope}[transparency group, opacity = 1.5*\shading]
	\draw[excessred, edgeshaded] (s)--(v6);
	\draw[excessred, edgeshaded] (v5)--(t);
	\end{scope}

\end{tikzpicture}}%
		\caption{Augmenting paths\label{fig:graphAP}}
	\end{subfigure}
	\\[6pt]
	\begin{subfigure}[b]{\wthird}
	    \setlength\abovecaptionskip{0.2\baselineskip}
		\resizebox{\linewidth}{!}{\begin{tikzpicture}
	
	\bboxelse

	\foreach \n/\x/\y in {s/0/0, t/10/0, v1/0.8/1.9, v2/2.7/3, 
		v3/5/3.2, v4/7.3/3, v5/9.2/1.9,
		v6/1.2/-1.9, v7/3.6/-2.7, v8/6.4/-2.7, v9/8.8/-1.9,
		v10/3.8/0.7, v11/6.7/0.7, v12/5.2/-1}{
		\coordinate(\n) at (\x*\s, \y*\s);}

	\node[height] at ([shift=({135:11 pt})]s) {$14$};
	\node[height] at ([shift=({155:11 pt})]t) {$0$};
	\node[height] at ([shift=({135:8 pt})]v1) {$1$};
	\node[height] at ([shift=({135:10 pt})]v2) {$1$};
	\node[height] at ([shift=({125:10 pt})]v3) {$2$};
	\node[height] at ([shift=({125:8 pt})]v4) {$3$};
	\node[height] at ([shift=({110:8 pt})]v5) {$1$};
	\node[height] at ([shift=({200:8 pt})]v6) {$1$};
	\node[height] at ([shift=({115:10 pt})]v7) {$0$};
	\node[height] at ([shift=({125:8 pt})]v8) {$0$};
	\node[height] at ([shift=({110:8 pt})]v9) {$0$};
	\node[height] at ([shift=({170:8 pt})]v10) {$1$};
	\node[height] at ([shift=({125:8 pt})]v11) {$1$};
	\node[height] at ([shift=({175:10 pt})]v12) {$0$};
	
	\node[vertexterminal](s) at (s) {$s$};
	\node[vertexterminal](t) at (t) {$t$};
	
	\foreach \i in{1, 4, 5, 6, 8, 9, 10, 11}{
		\node[vertex](v\i) at (v\i) {};}
	
	\node[vertexexcess](v2) at (v2){$+1$};
	\node[vertexexcess](v3) at (v3){$+1$};
	\node[vertexexcess](v7) at (v7){$+3$};	
	\node[vertexexcess](v12) at (v12){$+6$};	
	
	\draw[edge] (s)--node[weight]{\fflow{10}{10}} (v1);
	\draw[edge] (v1)--node[weight]{\fflow{10}{20}} (v2);
	\draw[edge] (v2)--node[weight]{\fflow{9}{15}} (v3);
	\draw[edge] (v3)--node[weight]{\fflow{8}{9}} (v4);
	\draw[edge] (v4)--node[weight]{\fflow{8}{8}} (v5);
	\draw[edge] (v5)--node[weight]{\fflow{9}{10}} (t);
	
	\draw[edge] (s)--node[weight]{\fflow{10}{10}} (v6);
	\draw[edge] (v6)--node[weight]{\fflow{3}{3}} (v7);
	\draw[edge] (v7)--node[weight]{\fflow{0}{9}} (v8);
	\draw[edge] (v8)--node[weight]{\fflow{0}{4}} (v9);
	\draw[edge] (v9)--node[weight]{\fflow{0}{10}} (t);
	
	\draw[edge] (v6)--node[weight]{\fflow{7}{15}} (v10);
	\draw[edge] (v7)--node[weight]{\fflow{0}{5}} (v12);
	\draw[edge] (v10)--node[weight]{\fflow{0}{7}} (v2);
	\draw[edge] (v10)--node[weight]{\fflow{0}{2}} (v4);
	\draw[edge] (v10)--node[weight]{\fflow{1}{20}} (v11);
	\draw[edge] (v10)--node[weight]{\fflow{6}{6}} (v12);
	\draw[edge] (v11)--node[weight]{\fflow{1}{12}} (v5);
	\draw[edge] (v12)--node[weight]{$6$} (v9);		
	
\end{tikzpicture}}
		\caption{Preflow push-relabel\label{fig:graphPPR}}
	\end{subfigure}\hfil
	\begin{subfigure}[b]{\wthird}
	    \setlength\abovecaptionskip{0.2\baselineskip}
		\resizebox{\linewidth}{!}{\begin{tikzpicture}
	
	\bboxelse
		
	\foreach \n/\x/\y in {s/0/0, t/10/0, v1/0.8/1.9, v2/2.7/3, 
		v3/5/3.2, v4/7.3/3, v5/9.2/1.9,
		v6/1.2/-1.9, v7/3.6/-2.7, v8/6.4/-2.7, v9/8.8/-1.9,
		v10/3.8/0.7, v11/6.7/0.7, v12/5.2/-1}{
		\coordinate(\n) at (\x*\s, \y*\s);}
	
	\node[vertexterminal](s) at (s) {$s$};
	\node[vertexterminal](t) at (t) {$t$};
	
	\node[height] at ([shift=({135:8 pt})]v1) {$1$};
	\node[height] at ([shift=({135:8 pt})]v2) {$1$};
	\node[height] at ([shift=({125:10 pt})]v3) {$1$};
	\node[height] at ([shift=({125:10 pt})]v4) {$1$};
	\node[height] at ([shift=({110:10 pt})]v5) {$0$};
	\node[height] at ([shift=({200:8 pt})]v6) {$1$};
	\node[height] at ([shift=({115:10 pt})]v7) {$0$};
	\node[height] at ([shift=({125:8 pt})]v8) {$0$};
	\node[height] at ([shift=({125:10 pt})]v9) {$0$};
	\node[height] at ([shift=({170:8 pt})]v10) {$1$};
	\node[height] at ([shift=({125:8 pt})]v11) {$0$};
	\node[height] at ([shift=({175:8 pt})]v12) {$0$};

	\foreach \i in{1, 2,  6, 8, 10, 11, 12}{
		\node[vertex](v\i) at (v\i) {};}
	
	\node[vertexexcess](v3) at (v3){$+2$};
	\node[vertexexcess](v4) at (v4){$+1$};
	\node[vertexexcess, fill=deficitblue](v5) at (v5){$-2$};		
	\node[vertexexcess](v7) at (v7){$+3$};
	\node[vertexexcess, fill=deficitblue](v9) at (v9){$-4$};	
	
	\def\lw{10pt}
	\begin{scope}[transparency group, opacity = 2*\shading]
		\draw[heightgreen, edgeshaded] (v1)--(v2);
		\draw[heightgreen, edgeshaded] (v2)--(v3);
		\draw[heightgreen, edgeshaded] (v6)--(v10);
		\draw[heightgreen, edgeshaded] (v10)--(v2);
		\draw[heightgreen, edgeshaded] (v12)--(v9);
	\end{scope}
	
	\draw[edge] (s)--node[weight]{\fflow{10}{10}} (v1);
	\draw[edge] (v1)--node[weight]{\fflow{10}{20}} (v2);
	\draw[edge] (v2)--node[weight]{\fflow{11}{15}} (v3);
	\draw[edge] (v3)--node[weight]{\fflow{9}{9}} (v4);
	\draw[edge] (v4)--node[weight]{\fflow{8}{8}} (v5);
	\draw[edge] (v5)--node[weight]{\fflow{10}{10}} (t);
	
	\draw[edge] (s)--node[weight]{\fflow{10}{10}} (v6);
	\draw[edge] (v6)--node[weight]{\fflow{3}{3}} (v7);
	\draw[edge] (v7)--node[weight]{\fflow{0}{9}} (v8);
	\draw[edge] (v8)--node[weight]{\fflow{0}{4}} (v9);
	\draw[edge] (v9)--node[weight]{\fflow{10}{10}} (t);
	
	\draw[edge] (v6)--node[weight]{\fflow{7}{15}} (v10);
	\draw[edge] (v7)--node[weight]{\fflow{0}{5}} (v12);
	\draw[edge] (v10)--node[weight]{\fflow{1}{7}} (v2);
	\draw[edge] (v10)--node[weight]{\fflow{0}{2}} (v4);
	\draw[edge] (v10)--node[weight]{\fflow{0}{20}} (v11);
	\draw[edge] (v10)--node[weight]{\fflow{6}{6}} (v12);
	\draw[edge] (v11)--node[weight]{\fflow{0}{12}} (v5);
	\draw[edge] (v12)--node[weight]{\fflow{6}{6}} (v9);		
	
	\begin{scope}[transparency group, opacity = 2*\shading]
		\draw[heightgreen, edgeshaded] (v1)--(v2);
		\draw[heightgreen, edgeshaded] (v2)--(v3);
		\draw[heightgreen, edgeshaded] (v6)--(v10);
		\draw[heightgreen, edgeshaded] (v10)--(v2);
		\draw[heightgreen, edgeshaded] (v12)--(v9);
	\end{scope}
	
\end{tikzpicture}}
		\caption{Pseudoflow\label{fig:graphHPS}}
	\end{subfigure}
\caption{\textbf{Graph basics and serial algorithms.} (a) An example of the graph and a feasible \newtext{(non-maximal)} flow. The flow and capacity for each arc is written as $f_{ij}/c_{ij}$, and (to reduce clutter) zero-values of the flow are omitted. \newtext{The flow is 8, which is not maximal, so no $s$-$t$ cut is evident.} (b) \newtext{The \mincut with a value of 18, which all \mincut algorithms will eventually arrive at.} (c) Residual graph for the flow from (a). (d) An intermediate flow while running the AP algorithm. In the first iteration, 10 units are pushed along the path highlighted in orange \newtext{and red}, saturating two terminal arcs (red). In the next iteration, flow is pushed along the residual path highlighted in blue. (e) A \newtext{pre}flow \newtext{at an} intermediate stage of a PPR algorithm. Nodes with excess are shown in red, and a label in green is attached to every node. (f) A \newtext{pseudo}flow \newtext{at an} intermediate stage of the HPF algorithm. Nodes with surplus/deficit are shown in red/blue, a label is attached to every node, and arcs of the tree structure are highlighted in green. \label{fig:graphs}}
\end{figure}
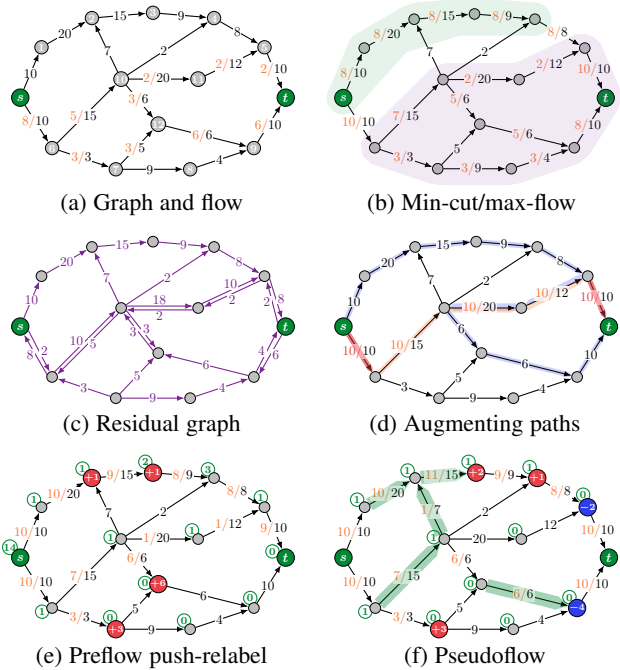

\section{Min-Cut/Max-Flow Algorithms for Computer Vision} 

To illustrate the use of \maxflow, we will sketch how a vision problem, image segmentation, can be solved using \maxflow. We start by introducing our notation and defining the \maxflow problem.

We define a directed graph $G = (V, E)$ by a set of \emph{nodes}, $V$, and a set of directed \emph{arcs}, $E$. We let $n$ and $m$ refer to the number of nodes and arcs, respectively. Each arc $(i, j) \in E$ is assigned a non-negative \emph{capacity} $c_{ij}$. For \maxflow problems, we define two special \emph{terminal nodes}, $s$ and $t$, which are referred to as the \emph{source} and \emph{sink}, respectively. The source has only outgoing arcs, while the sink has only incoming arcs. Arcs to and from the terminal nodes are known as \emph{terminal arcs}.

A \emph{feasible flow} in the graph $G$ is an assignment of non-negative numbers (flows), $f_{ij}$, to each arc $(i, j) \in E$. A feasible flow must satisfy the following two types of constraints: \emph{capacity constraints}, $f_{ij} \leq c_{ij}$, and \emph{conservation constraints}, $\sum_{i| (i,j)\in E} f_{ij} = \sum_{k|(j,k)\in E} f_{jk}$ for all nodes $j \in V \setminus \{s, t\}$. Capacity constraints ensure that the flow along an arc does not exceed its capacity. Conservation constraints ensure that the flow going into a node equals the flow coming out. See \cref{fig:graphflow} for an example of the graph and a feasible flow. The \emph{value} of the flow is the total flow out of the source or, equivalently, into the sink, and the \emph{maximum flow} problem refers to finding a feasible flow that maximizes the flow value.

An \emph{$s$-$t$ cut} is a partition of the nodes into two disjoint sets $S$ and $T$ such that $s \in S$ and $t \in T$. The sets $S$ and $T$ are referred to as the source and sink set, respectively. The \emph{capacity} of the cut is the sum of capacities of the arcs going from $S$ to $T$. And the \emph{minimum cut} problem refers to finding a cut that minimizes the cut capacity. Often, this partition of the nodes is all that is needed for computer vision applications. Therefore, some algorithms only compute the minimum cut, and an additional step would be needed to extract the flow value for every arc.

Finally, the max-flow min-cut theorem 
states that the value of the maximum flow is exactly the capacity of the minimum cut. \cref{fig:graphcut} shows \mincut on a small graph. This can be \newtext{shown} by formulating both problems as linear programs, \newtext{which reveals} that max-flow is the \newtext{strong} dual of the min-cut.

\subsection{Image Segmentation}
\def\s{0.8} 
\def\wsub{0.18\linewidth} 
\definecolor{one}{RGB}{153, 153, 153}
\definecolor{two}{RGB}{163, 163, 163}
\definecolor{three}{RGB}{172, 172, 172}
\definecolor{four}{RGB}{182, 182, 182}
\definecolor{five}{RGB}{192, 192, 192}
\definecolor{six}{RGB}{201, 201, 201}
\definecolor{seven}{RGB}{211, 211, 211}
\definecolor{eight}{RGB}{220, 220, 220}
\definecolor{nine}{RGB}{230, 230, 230}

\definecolor{residualcol}{RGB}{121, 35, 142}
\definecolor{terminalcol}{RGB}{0, 136, 53}

\definecolor{excessred}{RGB}{232, 63, 72}
\definecolor{floworange}{RGB}{252, 118, 52}
	
\tikzstyle{vertexnew} = [draw=black, fill=nodegray, circle, minimum size=8pt, inner sep=0pt]
\tikzstyle{vertexterminal} = [circle, draw=black, fill=terminalcol, text=white, font=\boldmath \footnotesize, , minimum size=10pt, inner sep=0pt]

\tikzstyle{superpix} = [terminalcol, line width=3pt]

\tikzstyle{vertexlast} = [draw=black, fill=nodegray, circle, minimum size=6pt, inner sep=0pt]
\tikzstyle{edgelast} = [black, ->, >={Latex[length=1.2mm]}]
\tikzstyle{edgelastboth} = [edgelast, <->]
\tikzstyle{inf} = [residualcol]
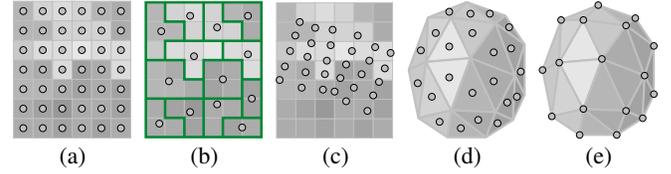
\begin{figure}
	\centering
	\begin{subfigure}[b]{\wsub}
	    \setlength\abovecaptionskip{0.2\baselineskip}
		\resizebox{\linewidth}{!}{\begin{tikzpicture}
		
	\foreach \x/\y/\col in {0/0/three, 1/0/three, 2/0/two, 3/0/three, 4/0/two, 5/0/three, 0/1/three, 1/1/two, 2/1/one, 3/1/three, 4/1/three, 5/1/two, 0/2/four, 1/2/four, 2/2/three, 3/2/two, 4/2/three, 5/2/two, 0/3/four, 1/3/three, 2/3/eight, 3/3/two, 4/3/three, 5/3/eight, 0/4/six, 1/4/eight, 2/4/eight, 3/4/nine, 4/4/eight, 5/4/eight, 0/5/four, 1/5/four, 2/5/seven, 3/5/seven, 4/5/seven, 5/5/three, 0/6/four, 1/6/seven, 2/6/eight, 3/6/seven, 4/6/four, 5/6/three}{%
		\draw[five, line width = 1pt, fill=\col] (\s*\x, \s*\y) rectangle (\s*\x+\s*1,\s*\y+\s*1);
		\node[vertexnew] at (\s*\x + \s*0.5, \s*\y + \s*0.5){};
	}

\end{tikzpicture}}
		\caption{}
	\end{subfigure}\hfil
	\begin{subfigure}[b]{\wsub}
		\setlength\abovecaptionskip{0.2\baselineskip}
		\resizebox{\linewidth}{!}{\begin{tikzpicture}
	
	\foreach \x/\y/\col in {0/0/three, 1/0/three, 2/0/two, 3/0/three, 4/0/two, 5/0/three, 0/1/three, 1/1/two, 2/1/one, 3/1/three, 4/1/three, 5/1/two, 0/2/four, 1/2/four, 2/2/three, 3/2/two, 4/2/three, 5/2/two, 0/3/four, 1/3/three, 2/3/eight, 3/3/two, 4/3/three, 5/3/eight, 0/4/six, 1/4/eight, 2/4/eight, 3/4/nine, 4/4/eight, 5/4/eight, 0/5/four, 1/5/four, 2/5/seven, 3/5/seven, 4/5/seven, 5/5/three, 0/6/four, 1/6/seven, 2/6/eight, 3/6/seven, 4/6/four, 5/6/three}{%
		\draw[five, line width = 1pt, fill=\col] (\s*\x, \s*\y) rectangle (\s*\x+\s*1,\s*\y+\s*1);
	}
	
	\draw[superpix](\s*1,\s*4)--(\s*1,\s*5)--(\s*6,\s*5);
	\draw[superpix](\s*0,\s*4)--(\s*2,\s*4)--(\s*2,\s*3)--(\s*3,\s*3);
	\draw[superpix](\s*3,\s*0)--(\s*3,\s*4)--(\s*5,\s*4)--(\s*5,\s*1);	
	\draw[superpix](\s*0,\s*2)--(\s*5,\s*2);
	\draw[superpix](\s*2,\s*0)--(\s*2,\s*1)--(\s*1,\s*1)--(\s*1,\s*2);
	\draw[superpix](\s*4,\s*0)--(\s*4,\s*1)--(\s*6,\s*1);
	\draw[superpix](\s*5,\s*3)--(\s*6,\s*3);
	\draw[superpix](\s*4,\s*4)--(\s*4,\s*5);
	\draw[superpix](\s*3,\s*5)--(\s*3,\s*7);
	\draw[superpix](\s*2,\s*5)--(\s*2,\s*6)--(\s*1,\s*6)--(\s*1,\s*7);
	\draw[superpix](\s*4,\s*7)--(\s*4,\s*6)--(\s*5,\s*6)--(\s*5,\s*5);
	\draw[superpix](\s*0,\s*7) rectangle (\s*6,\s*0);
	
	\foreach \x/\y in {
		0.8/5.5,  2.3/6.3, 3.7/5.7, 5.3/6.3,
		2.5/4.2,  5.3/4.3,
		1.2/2.9, 4/3, 5.5/2, 5/0.5,
		0.7/0.7,  2.3/1.3, 3.7/1.7
	}{
		\node[vertexnew] at (\s*\x , \s*\y ){};
	}
	
\end{tikzpicture}}
		\caption{}
	\end{subfigure}\hfil
	\begin{subfigure}[b]{\wsub}
		\setlength\abovecaptionskip{0.2\baselineskip}
		\resizebox{\linewidth}{!}{\begin{tikzpicture}
	
	\foreach \x/\y/\col in {0/0/three, 1/0/three, 2/0/two, 3/0/three, 4/0/two, 5/0/three, 0/1/three, 1/1/two, 2/1/one, 3/1/three, 4/1/three, 5/1/two, 0/2/four, 1/2/four, 2/2/three, 3/2/two, 4/2/three, 5/2/two, 0/3/four, 1/3/three, 2/3/eight, 3/3/two, 4/3/three, 5/3/eight, 0/4/six, 1/4/eight, 2/4/eight, 3/4/nine, 4/4/eight, 5/4/eight, 0/5/four, 1/5/four, 2/5/seven, 3/5/seven, 4/5/seven, 5/5/three, 0/6/four, 1/6/seven, 2/6/eight, 3/6/seven, 4/6/four, 5/6/three}{%
		\draw[five, line width = 1pt, fill=\col] (\s*\x, \s*\y) rectangle (\s*\x+\s*1,\s*\y+\s*1);
	}
	
	\foreach \x/\y in {
		1.2/6.0, 2.0/5.7, 2.9/5.6, 4.2/5.3, 5.1/4.7, 5.9/4.4, 0.97/5.3, 1.8/4.9, 2.8/4.8, 3.9/4.5, 4.8/4.0, 5.6/3.6, 0.7/4.5, 1.6/4.2, 2.6/4.0, 3.6/3.8, 4.4/3.3, 5.3/2.9, 0.43/3.7, 1.5/3.4, 2.5/3.3, 3.3/3.1, 4.0/2.6, 5.0/2.2, 0.16/3.0, 1.3/2.6, 2.4/2.5, 3.0/2.4, 3.6/1.9, 4.7/1.4
	}{
		\node[vertexnew] at (\s*\x , \s*\y ){};
	}
\end{tikzpicture}}
		\caption{}
	\end{subfigure}\hfil
	\begin{subfigure}[b]{\wsub}
		\setlength\abovecaptionskip{0.2\baselineskip}
		\resizebox{\linewidth}{!}{\begin{tikzpicture}
		
	\foreach \i/\x/\y in {1/3.0/7.0, 2/5.0/6.3, 3/1.0/6.2, 4/2.2/6.2, 5/4.5/5.9, 6/6.0/4.8, 7/0.0/3.5, 8/0.9/3.9, 9/3.2/4.1, 10/5.5/3.2, 11/6.0/2.2, 12/0.5/1.0, 13/2.2/1.4, 14/4.5/1.3, 15/5.4/0.9, 16/2.0/0.0, 17/4.0/0.0}{
	\coordinate(v\i) at (\x*\s, \y*\s);}

	\foreach \i/\j/\k/\col in {1/4/3/six, 1/5/4/six, 1/2/5/four, 2/6/5/two, 
		3/8/7/six, 3/4/8/seven, 4/9/8/nine, 4/5/9/six, 5/10/9/three, 5/6/10/two, 6/11/10/one, 
		7/8/12/six, 8/13/12/seven, 8/9/13/nine, 9/14/13/four, 9/10/14/three, 10/15/14/two, 10/11/15/one,
		12/13/16/six, 13/17/16/four, 13/14/17/three, 14/15/17/two}{
		\draw[five, line width = 2.5pt, fill=\col, join=bevel] (v\i.center)--(v\j.center)--(v\k.center)--cycle;
	}
	
	\foreach \x/\y in{2.1/6.5, 3.2/6.4, 4.2/6.4, 5.2/5.7, 0.63/4.5, 1.4/5.4, 2.1/4.7, 3.3/5.4, 4.4/4.4, 5.3/4.6, 5.8/3.4, 0.47/2.8, 1.2/2.1, 2.1/3.1, 3.3/2.3, 4.4/2.9, 5.1/1.8, 5.6/2.1, 1.6/0.8, 2.7/0.47, 3.6/0.9, 4.6/0.73}{
		\node[vertexnew] at (\x*\s,\y*\s) {};
	}
	
\end{tikzpicture}}
		\caption{}
	\end{subfigure}\hfil
	\begin{subfigure}[b]{\wsub}
		\setlength\abovecaptionskip{0.2\baselineskip}
		\resizebox{\linewidth}{!}{\begin{tikzpicture}
	
	\foreach \i/\x/\y in {1/3.0/7.0, 2/5.0/6.3, 3/1.0/6.2, 4/2.2/6.2, 5/4.5/5.9, 6/6.0/4.8, 7/0.0/3.5, 8/0.9/3.9, 9/3.2/4.1, 10/5.5/3.2, 11/6.0/2.2, 12/0.5/1.0, 13/2.2/1.4, 14/4.5/1.3, 15/5.4/0.9, 16/2.0/0.0, 17/4.0/0.0}{
		\coordinate(v\i) at (\x*\s, \y*\s);}
	
	\foreach \i/\j/\k/\col in {1/4/3/six, 1/5/4/six, 1/2/5/four, 2/6/5/two, 
		3/8/7/six, 3/4/8/seven, 4/9/8/nine, 4/5/9/six, 5/10/9/three, 5/6/10/two, 6/11/10/one, 
		7/8/12/six, 8/13/12/seven, 8/9/13/nine, 9/14/13/four, 9/10/14/three, 10/15/14/two, 10/11/15/one,
		12/13/16/six, 13/17/16/four, 13/14/17/three, 14/15/17/two}{
		\draw[five, line width = 2.5pt, fill=\col, join=bevel] (v\i.center)--(v\j.center)--(v\k.center)--cycle;
	}
	
	\foreach \i in{1, ..., 17}{
		\node[vertexnew] (v\i) at (v\i) {};
	}
	
\end{tikzpicture}}
		\caption{}
	\end{subfigure}
	\caption{\textbf{Some possibilities for associating graph nodes with entities used for segmentation.} Graph nodes (\newtext{gray} dots) associated with  (a) image pixels, (b) superpixels (c) positions in the image (d) mesh faces (e) mesh vertices. \label{fig:segmentation_nodes}}
\end{figure}
\def\wsub{0.25 \linewidth} 
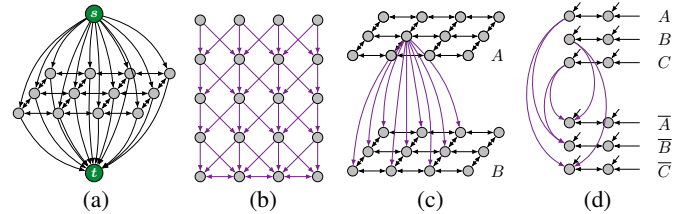
\begin{figure}
	\centering
	\begin{subfigure}[b]{\wsub}
		\setlength\abovecaptionskip{0.2\baselineskip}
		\resizebox{\linewidth}{!}{\begin{tikzpicture}
	
	\foreach \x in {1, 2, 3,4}{\foreach \y in {1,2,3}{
			\coordinate (v\x\y) at  (\s*\x+\s*0.4*\y , \s*0.5*\y ){};
	}}
	
	\node[vertexterminal] (t) at (\s*3.3, \s*-1){$t$};
	\node[vertexterminal] (s) at (\s*3.3, \s*3){$s$};

%
	
	\def\dx{15}
	\def\dy{2}
	
	\foreach \x in {1, 2, 3,4}{\foreach \y in {1,2,3}{
			\pgfmathsetmacro\bend{-\dx*(\x-2.5)-\dy*(\y-2)}
			\draw[->, edgelast] (v\x\y) to [bend right = \bend]  (t);
	}}

	\foreach \x in {1, 2, 3,4}{\foreach \y in {1,2,3}{
		\node[draw, vertexlast]  (v\x\y) at (v\x\y){};
	}}

	\foreach \x in {1, 2, 3,4}{\foreach \y in {1,2,3}{
		\pgfmathsetmacro\bend{-\dx*(\x-2.5)-\dy*(\y-2)}
		\draw [->, edgelast] (s) to[bend right=\bend]  (v\x\y);
	}}

	\foreach \x in {1,2,3,4}{\foreach \y[count=\yy from 2] in {1,2}{
		\draw[edgelastboth] (v\x\yy)--(v\x\y);	
	}}
	\foreach \x[count=\xx from 2] in {1,2,3}{\foreach \y in {1,2,3}{
		\draw[edgelastboth] (v\xx\y)--(v\x\y);	
	}}

\end{tikzpicture}}
		\caption{\label{fig:segmentation_edges:st}}
	\end{subfigure}\hfil
	\begin{subfigure}[b]{\wsub}
		\setlength\abovecaptionskip{0.2\baselineskip}
		\resizebox{\linewidth}{!}{\begin{tikzpicture}
	
	\node at (0.5,\s*2) {};
	\node at (\s*4.5,\s*2) {};
	
	\foreach \x in {1,2,3,4}{\foreach \y in {1,2,3,4,5}{
			\node[vertexlast] (v\x\y) at (\s*\x , \s*\y ){};
	}}
	
	\foreach \x in {1,2,3,4}{\foreach \y[count=\yy from 2] in {1,2,3,4}{
			\draw[->, edgelast, inf] (v\x\yy)--(v\x\y);	
	}}
	\foreach \x[count=\xx from 2] in {1,2,3}{\foreach \y[count=\yy from 2] in {1,2,3,4}{
			\draw[->, edgelast, inf] (v\xx\yy)--(v\x\y);	
			\draw[->, edgelast, inf] (v\x\yy)--(v\xx\y);	
	}}
	\foreach \x[count=\xx from 2] in {1,2,3}{
		\draw[edgelastboth, inf] (v\xx1)--(v\x1);	
	}
	
	
\end{tikzpicture}}
		\caption{\label{fig:segmentation_edges:layer}}
	\end{subfigure}\hfil
	\begin{subfigure}[b]{\wsub}
		\setlength\abovecaptionskip{0.2\baselineskip}
		\resizebox{\linewidth}{!}{\begin{tikzpicture}
	
	\foreach \x in {1, 2, 3, 4}{\foreach \y in {1,2,3}{
			\coordinate (va\x\y) at  (\s*\x+\s*0.4*\y , \s*0.5*\y );
			\coordinate (vb\x\y) at (\s*\x+\s*0.4*\y , \s*0.5*\y +\s*3);
	}}
	
	\foreach \x in {1,2,3,4}{\foreach \y in {1,2,3}{
			\node[vertexlast] (va\x\y) at (va\x\y){};
			\node[vertexlast] (vb\x\y) at (vb\x\y){};
	}}
	
	\foreach \x in {1,2,3,4}{\foreach \y[count=\yy from 2] in {1,2}{
		\draw[edgelastboth] (va\x\yy)--(va\x\y);	
		\draw[edgelastboth] (vb\x\yy)--(vb\x\y);	
	}}

	\foreach \x[count=\xx from 2] in {1,2,3}{\foreach \y in {1,2,3}{
			\draw[edgelastboth] (va\xx\y)--(va\x\y);
			\draw[edgelastboth] (vb\xx\y)--(vb\x\y);	
	}}

	\tikzstyle{lbl} = [right=10pt, text=black]
	\node[lbl] at (va41) {$B$};
	\node[lbl] at (vb41) {$A$};
	
	\def\dx{15}
	\def\dy{2}
	
	\foreach \y in {1,2,3}{
		\foreach \x in {1,2,3}{
		\pgfmathsetmacro\bend{-\dx*(\x-2)-\dy*(\y-2)}
		\draw[edgelast, inf] (vb22) to [bend right=\bend](va\x\y);	
	}}

\foreach \x in {1,2,3,4}{\foreach \y in {1,2,3}{
		\node[vertexlast] (vb\x\y) at (vb\x\y){};
}}
	
\end{tikzpicture}}
		\caption{\label{fig:segmentation_edges:inclusion}}
	\end{subfigure}\hfil
	\begin{subfigure}[b]{\wsub}
		\setlength\abovecaptionskip{0.2\baselineskip}
		\resizebox{\linewidth}{!}{\begin{tikzpicture}
	
	\foreach \x in {1, 2, 3}{\foreach \y in {1,2}{
			\coordinate (va\x\y) at (\s*\x+\s*0.4*\y , \s*0.5*\y ){};
			\coordinate (vb\x\y) at (\s*\x+\s*0.4*\y , \s*0.5*\y +\s*0.6){};
			\coordinate (vc\x\y) at (\s*\x+\s*0.4*\y , \s*0.5*\y +\s*1.2){};
			\coordinate (vaa\x\y) at (\s*\x+\s*0.4*\y , \s*0.5*\y +\s*2.75){};
			\coordinate (vbb\x\y) at (\s*\x+\s*0.4*\y , \s*0.5*\y +\s*3.35){};
			\coordinate (vcc\x\y) at (\s*\x+\s*0.4*\y , \s*0.5*\y +\s*3.95){};
	}}

	\foreach \x in {1,2}{\foreach \lag in {a,b,c,aa,bb,cc}{
		\node[vertexlast] (v\lag\x1) at (v\lag\x1){};
	}}

	\tikzstyle{lbl} = [text=black, right=3pt] %
	\node[lbl] at (va31) {$\overline{C}$};
	\node[lbl] at (vb31) {$\overline{B}$};
	\node[lbl] at (vc31) {$\overline{A}$};
	\node[lbl] at (vaa31) {$C$};
	\node[lbl] at (vbb31) {$B$};
	\node[lbl] at (vcc31) {$A$};
	
	\foreach \x in {1,2}{\foreach \lag in {a,b,c,aa,bb,cc}{
			\draw[edgelast,  shorten <=.15cm] (v\lag\x2)--(v\lag\x1);	 
	}}

	\foreach \lag in {a,b,c,aa,bb,cc}{
			\draw[edgelastboth] (v\lag21)--(v\lag11);
			\draw[edgelast, shorten <=.15cm] (v\lag31)--(v\lag21);  
	}

	\draw [edgelast, inf] (vbb11) to [bend left=60] (va11);
	\draw [edgelast, inf] (vbb11) to [bend left=60] (vc11);
	\draw [edgelast, inf] (vaa11) to [bend right=60] (vb11);
	\draw [edgelast, inf] (vaa11) to [bend right=60] (vc11);
	\draw [edgelast, inf] (vcc11) to [bend right=60] (vb11);
	\draw [edgelast, inf] (vcc11) to [bend right=60] (va11);	
	
	
\end{tikzpicture}}
		\caption{\label{fig:segmentation_edges:exclusion}}
	\end{subfigure}
	\caption{\textbf{Some typical segmentation models.}  Terminal arcs are shown only for the first example. Arcs drawn in purple have infinite capacity.
	(a)  A classical MRF segmentation with 4-connected grid graph.  
	(b) A multi-column graph used for segmenting layered structures. 
	(c) Two-object segmentation with inclusion constraint. 
	(d) Three-object segmentation with mutual exclusion using QPBO. 
	\label{fig:segmentation_edges}}
\end{figure}

When formulating segmentation as a \maxflow problem, one modeling choice involves deciding which structures to represent as graph nodes. Often, nodes of the graph represent individual image pixels, but various other entities may be associated with graph nodes, some of which are illustrated in \cref{fig:segmentation_nodes}.

The energy formulation \eqref{Eq:energy} is convenient when \maxflow algorithms are used to optimize MRFs.
Here, each unary energy term is a likelihood energy (negative log likelihood) of a pixel being labeled 0 or 1. Likelihood terms are typically computed directly from image data. The pairwise terms are defined for pairs of pixels, so-called neighbors, and for 2D images, the neighborhood structure is usually given by a 4 or 8-connectivity.

The typical pairwise energy terms used in \eqref{Eq:energy} are
\begin{equation}
	\mathcal{E}_{ij}(0,0){=}\mathcal{E}_{ij}(1,1){=}0 \text{ and }\mathcal{E}_{ij}(0,1){=}\mathcal{E}_{ij}(1,0){=}\beta_{ij}\ .
	\label{Eq:energy_with_beta}
\end{equation}
These terms penalize neighboring pixels having different labels by a fixed amount, $\beta_{ij}$, thus \emph{encouraging} smoothness of the segmentation. In this case, the construction of the $s$-$t$ graph which exactly represents the energy function is straightforward: The node set is $V = \mathcal{P} \cup \{s,t\}$. For terminal arc capacities, $c_{si}$ and $c_{it}$, we use the unary terms $\mathcal{E}_i(0)$ and $\mathcal{E}_i(1)$, respectively. Meanwhile, pairwise energy terms correspond to non-terminal arc capacities, such that $c_{ij}=c_{ji}=\beta_{ij}$. \cref{fig:segmentation_edges:st} shows this construction for a 4-connected grid graph. The binary segmentation of the image corresponds directly to the binary labeling given by the minimum cut. Put in another way, the sets $S$ and $T$ give the optimal labeling of the nodes, and because we have a 1-to-1 mapping between non-terminal nodes and pixels, the node labeling is the segmentation.
However, there are many more advanced ways to formulate image segmentation using binary energy optimization and $s$-$t$ graphs, and ways to formulate other computer vision problems as well \cite{Kolmogorov2004}.

An example closely related to image segmentation is surface fitting, where \cite{Li2005} uses arcs of infinite capacity (\ie, infinite pairwise energy terms) to \emph{impose} a structure to the optimal solution. In \cref{fig:segmentation_edges:layer}, downward-pointing arcs ensure that if a pixel is in a source set, the column of pixels below it is also in the source set --- so the optimal solution has to be a layer. The slanted arcs impose the smoothness of this layer.

It is also possible to formulate multi-label/multi-object segmentation problems that can be solved with a single $s$-$t$ cut \cite{Delong2008, Ishikawa2003, Jeppesen2020, Li2005}, or by iteratively changing and computing the cut \cite{Boykov2001b, Isack2017}. For the single-cut Ishikawa method \cite{Ishikawa2003}, it is common to duplicate the graph for each label, \ie, having a sub-graph per label. For example, in \cref{fig:segmentation_edges}, each pixel is represented by two nodes: one for object $A$ and one for object $B$, so a pixel may be segmented as belonging to $A$, $B$, both, or neither. The submodular interaction between the objects may be achieved by adding arcs between the sub-graphs. \cref{fig:segmentation_edges:inclusion} shows submodular interaction, where arcs with infinite capacity ensure that if a pixel belongs to object $A$, this pixel and all its neighbors also belong to object $B$. This is known as \emph{inclusion} or \emph{containment} with a minimum margin of one.

In the examples covered so far, the arcs between the graph nodes correspond to submodular energy terms, which means the energy is lower when the nodes belong to the same set ($S$ or $T$). Mutual exclusion, in the general case, requires non-submodular energies which are not directly translatable to arcs in the graphs shown so far. An alternative is to use QPBO \cite{Kolmogorov2007}, as illustrated in \cref{fig:segmentation_edges:exclusion}\newtext{, which can handle any energy function of the form in \eqref{Eq:energy} --- submodular or not}. When using QPBO, we construct two sub-graphs for each object: one representing the object and another representing its complement. The exclusion of two objects, say $A$ and $B$, is then achieved by adding inclusion arcs from $A$ to $\overline{B}$ and from $B$ to $\overline{A}$. However, there is no guarantee that the \maxflow solution yields a complete segmentation of the object as the object and its complement may disagree on the labeling of some nodes leaving them ``unlabeled''. The number of unlabeled nodes depends on the non-submodularity of the system. \newtext{Extensions to QPBO, such as QPBO-P and QPBO-I \cite{Rother2007}, may be used to iteratively assign labels to the nodes that QPBO failed to label.}

\section{Serial Min-Cut/Max-Flow Algorithms}

All \maxflow algorithms find the solution by iteratively updating a flow that satisfies the capacity constraints. Such a flow induces a \emph{residual graph} with the set of \emph{residual arcs}, $R$, given by
\begin{equation}
	\begin{aligned}
		R = \{ (i,j) \in V{\times}V \, \mid\ \, &(i,j) \in E, f_{ij} < c_{ij} \, \text{ or }\\
		&(j,i) \in E, f_{ji} > 0 \}.
	\label{Eq:residual_graph}
	\end{aligned}
\end{equation}
Each of the residual arcs has a \emph{residual capacity} given by $c'_{ij} = c_{ij} - f_{ij}$ if $(i,j) \in E$ or $c'_{ij} = f_{ji}$ if $(j, i) \in E$. In other words, residual arcs tell us how much flow on the original arc we can increase or decrease, see \cref{fig:graphresidual}. If the graph contains bidirectional arcs, both conditions from \eqref{Eq:residual_graph} may be met, and the residual capacity then equals the sum of two contributions.

Serial \maxflow algorithms can be divided into three families: augmenting paths, preflow push-relabel, and pseodoflow algorithms. In this section, we provide an overview of how algorithms from each family work.

\subsection{Augmenting Paths}
The augmenting paths (AP) family of \maxflow algorithms is the oldest of the three families and was introduced with the Ford-Fulkerson algorithm \cite{Ford1962}. An algorithm from the AP family always maintains a feasible flow. It works by repeatedly finding so-called \emph{augmenting paths}, which are paths from $s$ to $t$ in the residual graph.  When an augmenting path is found, a flow is pushed along the path. Pushing flow  means increasing flow for each forward arc along the path, and decreasing flow for each reverse arc. To maintain the capacity constraints, the flow that is pushed equals the minimum residual capacity along the path. Conservation constrains are maintained as the algorithm only updates complete paths from $s$ to $t$. The algorithm terminates when no augmenting paths can be found. \cref{fig:graphAP} shows an intermediate stage of an AP algorithm.

The primary difference between various AP algorithms lies in how the augmenting paths are found. For computer vision applications, the most popular AP algorithm is the Boykov-Kolmogorov (BK) algorithm \cite{Boykov2004}, which works by building search trees from both the source and sink nodes to find augmenting paths and uses a heuristic that favors shorter augmenting paths. The BK algorithm performs well on many computer vision problems, but its theoretical run time bound is worse than other algorithms \cite{Verma2012}.

In terms of performance, the BK algorithm has been surpassed by the Incremetal Breadth First Search (IBFS) algorithm by Goldberg \etal \cite{Goldberg2011}. The main difference between the two algorithms is that IBFS maintains the source and sink search trees as breadth-first search trees, which results in both better theoretical run time and better practical performance \cite{Goldberg2015, Goldberg2011}.

\subsection{Preflow Push-Relabel}\label{sec:ppr}
The second family of algorithms are the preflow push-relabel (PPR) algorithms, which were introduced by Goldberg and Tarjan \cite{Goldberg1988}. These algorithms use a so-called \emph{preflow}, which satisfies capacity constraints but allows nodes to have more incoming than outgoing flow, thus violating conservation constraints. The difference between the incoming and outgoing flows for a node, $i$, is denoted as its \emph{excess}, $e_i \geq 0$.

The PPR algorithms work by repeatedly pushing flow along the individual arcs. To determine which arcs admit flow, the algorithms maintain an integer labeling (so-called \emph{height}), $d_i$, for every node. The labeling \newtext{provides a lower bound on the distance from the node to the sink and} has a \emph{no steep drop} property, meaning $d(i)-d(j) \leq 1$ for any residual arc $(i,j)$.

An algorithm from the PPR family starts by saturating the source arcs and raising the source to $d(s)=n$.
The algorithm then works by repeatedly selecting a node with excess (after selection called a \emph{selected} node) and applying one of two actions \cite{Cherkassky1997,Goldberg1988}: \emph{push} or \emph{relabel}.  If there is an arc in the residual graph leading from the selected node to a lower-labeled node, \emph{push} is performed. This pushes excess along the arc, until all excess is pushed or the arc is saturated. If no residual arc leads to a lower node, the \emph{relabel} operation is used to lift the selected node (increase its label) by one.  \cref{fig:graphPPR} shows an intermediate step of \newtext{a PPR algorithm}.

When there are no nodes with excess left, the preflow is the maximum flow. It is possible to terminate the algorithm earlier, when no nodes with excess have a label $d_i < n$. At this point, the minimum $s$-$t$ cut can be extracted by inspecting the node labels. If $d_i \geq n$, then $i \in S$, otherwise $i \in T$. Extracting the maximum flow requires an extra step of pushing all excess back to the source. However, this work generally only represents a small part of the run time \cite{Verma2012} and, for computer vision applications, we are typically only interested in the minimum cut anyway.

The difference between various PPR algorithms lies in the order in which the push and relabel operations are performed. Early variants used simple heuristics, such as always pushing flow from the node with the highest label or using a first-in-first-out queue to keep track of nodes with positive excess \cite{Chandran2009}. More recent versions \cite{Arora2010,Goldberg2008,Goldberg2009} use sophisticated heuristics and a mix of local and global operations to obtain significant performance improvements over early PPR algorithms.

Unlike other serial algorithms, the algorithms from the PPR family operate locally on nodes and arcs. This, as we shall discuss later, has resulted in a whole family of parallel PPR algorithms.

\subsection{Pseudoflow}
The most recent family of \maxflow algorithms is the pseudoflow family, which was introduced with the Hochbaum pseudoflow (HPF) algorithm \cite{Hochbaum2008,Hochbaum2013}. These algorithms use a so-called \emph{pseudoflow}, which satisfies capacity constraints but not the conservation constraints, as it has no constraints on the difference between incoming and outgoing flow. As with preflow, we refer to the difference between incoming and outgoing flow for a node as its excess, $e_i$. A positive excess is referred to as a \emph{surplus} and a negative excess as a \emph{deficit}.

During operation, HPF algorithms maintain two auxiliary structures: the forest of trees and a node labeling function. Only one node in every tree, the root, is allowed to have an excess. The algorithm works by repeatedly pushing the flow along the paths connecting the trees, and growing the trees.

A generic algorithm from the HPF family is initialized by saturating all terminal arcs. At this point, each graph node is a singleton tree in the forest. The algorithm then selects a tree with surplus and containing at least one node with the label less than $n$ (the number of nodes in the graph). In this tree, $i$ denotes the node with the lowest label. If there are no residual arcs from $i$ to a node with a lower label, the label of $i$ is incremented. If there is a residual arc $(i,j)$ that leads to a node $j$ with a lower label, a \emph{merge} is performed. This operation involves pushing surplus along the path from the root of the tree containing $i$, over $i$, over $j$, and to the root of the tree containing $j$. If the arc capacities along this path allow it, the entire surplus will be pushed and the trees will be merged with $j$ as the root. If the flow along the path saturates an arc $(i',j')$, a surplus will be collected in $i'$, and a new tree rooted in $i'$ will be created. In contrast to the AP algorithms, the only restrictions on how much flow to push are the individual arc capacities, not the path capacity.

The algorithm terminates when no selection can be made\newtext{, at which point} nodes labeled with $n$ constitute the source set\newtext{.} Additional processing is needed to recover the maximum feasible flow. \cref{fig:graphHPS} shows an intermediate step of the \newtext{HPF} algorithm.

There are two main algorithms in this family: HPF and Excesses Incremental Breadth First Search (EIBFS) \cite{Goldberg2015}. The main differences are the order in which they scan through nodes when looking for an arc connecting two trees in the forest, and how they push flow along the paths. Both have sophisticated heuristics for these choices, which makes use of many of the same ideas developed for PPR algorithms.

\subsection{Implementation Details}
\label{sec:imple-details}
As stressed by \cite{Verma2012}, the implementation details can significantly affect the measured performance of a given \maxflow algorithm. In this section, we will highlight the trends of modern implementations and how they differ.

\subsubsection{Data Structures and Data Types}
The implementations considered in this paper all use a variant of the adjacency list structure \cite{clrs2009} to represent the underlying graph. The most common setup mimics the BK algorithm:
there is a list of nodes and a list of directed (half-)arcs. Each \texttt{Node} structure stores a pointer to its first outgoing half-arc. Each \texttt{Arc} stores a pointer to the node it points to, a pointer to the next outgoing arc for the node it points from, a pointer to its reverse arc, and a residual capacity. For algorithms implemented with computer vision applications in mind (\eg, BK, IBFS, and EIBFS), the terminal arcs are stored as a single combined terminal capacity for each \texttt{Node}, instead of using the \texttt{Arc} structures. Other implementations simply keep track of the source and sink nodes and use \texttt{Arc} structures for all arcs. The HPF implementation uses a bidirectional \texttt{Arc} structure with a capacity, a flow, and a direction. It \newtext{is} also common to store auxiliary values such as excesses, labels, or more.

As a result of these differences, the memory footprint varies between implementations, as shown in \cref{tab:mem-reqs}. The footprint also depends heavily on the data types used to store the data, in particular references to nodes and arcs, as we discuss in the next subsection. For storing arc capacities, integers are common because they are computationally efficient and may use as little as 1 byte. However, some graph constructions involve large capacities to model hard constraints, and here some care must be taken to avoid overflow issues. With floats, this can be modeled using infinite capacity. However, floats are less efficient and some algorithms are not guaranteed to terminate with floats due to numerical errors.

As the size of the data structures influences how much the CPU can store in its caches, which has a large effect on performance, it is generally beneficial to keep the data structures small. Note that some compilers do not pack data structures densely by default, which may significantly increase the size of the \texttt{Arc} and \texttt{Node} data structures.

\subsubsection{Indices vs. Pointers}
One way to reduce the size of the \texttt{Arc} and \texttt{Node} data structures on 64-bit system architectures is to use indices instead of pointers to reference nodes and arcs. As long as the indices can be stored using unsigned 32-bit integers, we can halve the size arc and node references by using unsigned 32-bit integers instead of pointers (which are 64-bit). This approach can significantly reduce the size of the \texttt{Arc} and \texttt{Node} data structures, as the majority of the structures consist of references to other arcs and nodes \cite{Jamriska2012}. As the performance of \maxflow algorithms is mainly limited by memory speed, smaller data structures can often lead to improved performance. The downside of indices is that extra computations may be needed for every look-up, although this depends on the exact assembly instructions the compiler chooses to use.

Some grid-based algorithms \cite{Jamriska2012} use 32-bit indices to reduce the size of their data structure. The generic algorithms we have investigated in this work all use pointers to store references between nodes and arcs. Some implementations avoid the extra memory requirement by compiling with 32-bit pointers. However, 32-bit pointers limit the size of the graph much more than 32-bit indices. The reason is that the 32-bit pointers only have 4 GiB of address space, and the \texttt{Node} and \texttt{Arc} structures they point to take up many bytes. For example, the smallest \texttt{Arc} structure we have tested, \cf \cref{tab:mem-reqs}, uses 24 bytes, meaning that an implementation based on 32-bit \emph{indices} could handle graphs with 24 times more arcs than an implementation based on 32-bit \emph{pointers}.


\subsubsection{Arc Packing}
The order in which the arcs are stored may significantly affect performance.
\emph{Arc packing} is used to reduce CPU cache misses by storing the arcs in the same order that the algorithm will access them.
For example, \maxflow algorithms often iterate over outgoing arcs from a node, making it beneficial to store outgoing arcs from the same node adjacent in memory. However, as arcs may be added to the graph in any order, packing the arcs usually incurs an overhead from maintaining the correct ordering or reordering all arcs as an extra step before computing the \maxflow.
Similar to arc packing, \emph{node packing} may improve performance. However, this is not done in practice as opposed to arc packing.

Of the \newtext{serial} reference implementations that we examined, only HI-PR \cite{Cherkassky1997}, IBFS, and EIBFS implement arc packing. These all implement it as an extra step, where arcs are reordered after building the graph but before the \maxflow computations start. None of the examined implementations use node packing.

\subsubsection{Arc Merging}
In practice, it is not uncommon that multiple arcs between the same pair of nodes are added to the graph. Merging these arcs into a single arc with a capacity equal to the sum of capacities of the merged arcs may reduce the graph size significantly. As this decreases both the memory footprint of the graph and the number of arcs to be processed, it can provide substantial performance benefits \cite{Jamriska2012, Shekhovtsov2013}. However, as redundant arcs can usually be avoided by careful graph construction and they should have approximately the same performance impact on all algorithms, we have not investigated the effects of this further.

\section{Parallel Min-Cut/Max-Flow}
Like serial algorithms, parallel algorithms for \maxflow problems can be split into families based on shared characteristics. A key characteristic is whether the algorithms parallelize over individual graph nodes (node-based parallelism) or split the graph into sub-graphs that are then processed in parallel (block-based parallelism). Other important algorithmic traits include whether the algorithm is distributed, which we do not consider in this paper, and the guarantees in terms of convergence, optimality, and completeness provided by the algorithm.

We should note that since many (but not all) \maxflow problems in computer vision are defined on grid graphs, several algorithms \cite{Gridcut2015,Jamriska2012,Peng2015,Vineet2008} have exploited this structure to create very efficient parallel implementations.
However, many important computer vision problems are not defined on grid graphs, so in this paper we focus on generic \maxflow algorithms.

The category of node-based parallel algorithms is generally dominated by parallel versions of PPR algorithms. In the block-based category, we have identified three main approaches: adaptive bottom-up merging, dual decomposition, and region discharge, which we investigate. In the following sections, we give an overview of each approach and briefly discuss its merits and limitations.

\subsection{Parallel Preflow Push-Relabel}
PPR algorithms have been the target of most parallelization efforts \cite{Anderson1995, Bader2006,Baumstark2015,Delong2008,Goldberg1991,Hong2010,Vineet2008}, since both push and relabel are local operations, which makes them well suited for parallelization. Because the operations are local, the algorithms generally parallelize over each node --- performing pushes and relabels concurrently. To avoid data races during these operations, PPR algorithms use either locking \cite{Anderson1995} or atomic operations \cite{Hong2010}. As new excesses are created, the corresponding nodes are added to a queue from which threads can poll them. In \cite{Baumstark2015}, a different approach is applied, where pushes are performed in parallel, but excesses and labels are updated later in a separate step, rather than immediately after the push.

Since parallel PPR algorithms parallelize over every node, they can achieve good speed-ups and scale well to modern multi-core processors \cite{Baumstark2015}, or even GPUs \cite{Vineet2008}.
However, these algorithms have not achieved dominance outside of large grid graphs for \maxflow problems \cite{Yu2015}. Since GPU hardware has advanced considerably in recent years, it is unclear whether GPU method should remain restricted to grid graphs, but this question is not within the scope of this paper.

\subsection{Adaptive Bottom-Up Merging}
\begin{figure}[ht!]
    \centering
    \begin{subfigure}[b]{0.23\linewidth}
        \setlength\abovecaptionskip{0.2\baselineskip}
        \centering\footnotesize
        Split graph\\*[2pt]
        \includegraphics[width=\linewidth]{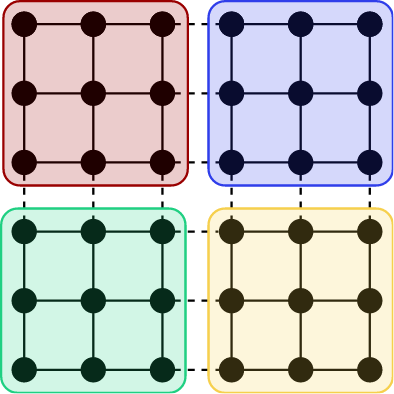}
        \caption{}
    \end{subfigure}\enspace
    \begin{subfigure}[b]{0.23\linewidth}
        \setlength\abovecaptionskip{0.2\baselineskip}
        \centering\footnotesize
        First merge\\*[2pt]
        \includegraphics[width=\linewidth]{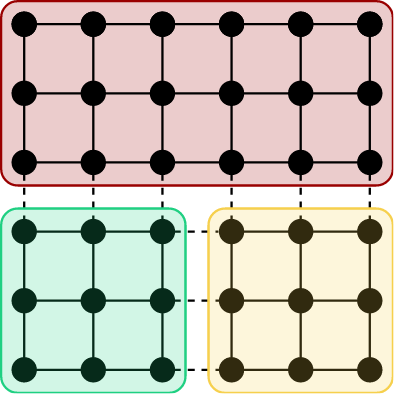}
        \caption{}
    \end{subfigure}\enspace
    \begin{subfigure}[b]{0.23\linewidth}
        \setlength\abovecaptionskip{0.2\baselineskip}
        \centering\footnotesize
        Second merge\\*[2pt]
        \includegraphics[width=\linewidth]{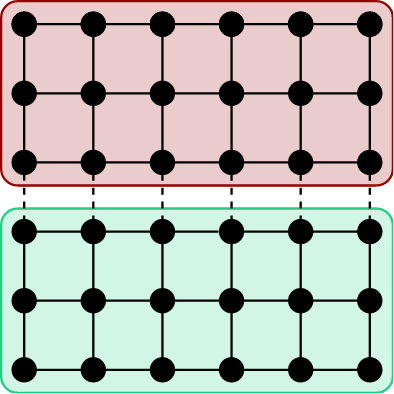}
        \caption{}
    \end{subfigure}\enspace
    \begin{subfigure}[b]{0.23\linewidth}
        \setlength\abovecaptionskip{0.2\baselineskip}
        \centering\footnotesize
        Last merge\\*[2pt]
        \includegraphics[width=\linewidth]{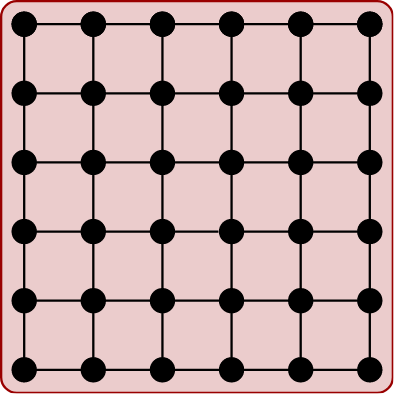}
        \caption{}
    \end{subfigure}
    \caption{\textbf{Illustration of the adaptive bottom-up merging approach for parallel \maxflow.} Terminal nodes and arcs are not shown. Note that the underlying graph does \emph{not} have to be a grid graph. Phase one: (a) The graph is split into blocks and the \maxflow is computed for each block in parallel. Phase two: (b) The topmost blocks are locked, merged, and the \maxflow recomputed. (c) As the topmost block is locked, the next thread works on the bottom-most blocks (in parallel). (d) Last two blocks are merged and \maxflow recomputed to achieve the globally optimal solution.}
    \label{fig:liu-sun-split}
\end{figure}

The adaptive bottom-up merging approach introduced by Liu and Sun \cite{Liu2010} uses block-based parallelism and has two phases, which are summarized in \cref{fig:liu-sun-split}. In phase one, the graph is partitioned into a number of disjoint sets (blocks), and arcs between blocks have their capacities set to 0 --- effectively removing them from the graph. For each pair of blocks connected by arcs, we store a list of the connecting arcs (with capacities now set to 0) along with their original capacities. Disregarding $s$ and $t$, the nodes in each block now belong to disjoint sub-graphs and we can compute the \maxflow solution for each sub-graph in parallel. The \maxflow computations are done with the BK algorithm --- although one could in theory use any \maxflow algorithm.

In phase two, we merge the blocks to obtain the complete globally optimal \maxflow. To merge two blocks, we restore the arc capacities for the connecting arcs and then recompute the \maxflow for the combined graph. This step makes use of the fact that the BK algorithm can
efficiently recompute the \maxflow when small changes are made to the residual graph for a \maxflow solution \cite{Kohli2007}.

For merges in phase two to be performed in parallel, the method marks the blocks being merged as locked. The computational threads then scan the list of block pairs, which were originally connected by arcs, until they find a pair of unlocked blocks. The thread then locks both blocks, performs the merge, and unlocks the new combined block. To avoid two threads trying to lock the same block, a global lock prevents more than one thread from scanning the list of block pairs at a time.

As the degree of parallelism decreases towards the end of phase two --- since there are few blocks left to merge --- performance increases when computationally expensive merges are performed early in phase two. To estimate the cost of merging two blocks, \cite{Liu2010} uses a heuristic based on the potential for new augmenting paths to be formed by merging two blocks. This heuristic determines the merging order of the blocks.

By using block-based, rather than node-based parallelism, adaptive bottom-up merging avoids much of the synchronization overhead that the parallel PPR algorithms suffer from. However, its performance depends on the majority of the work being performed in phase one and in the beginning of phase two, where the degree of parallelism is high.

\subsection{Dual Decomposition}
\begin{figure}[ht!]
    \centering
    \begin{subfigure}[b]{0.215\linewidth}
        \setlength\abovecaptionskip{0.2\baselineskip}
        \centering\footnotesize
        Split graph\\*[8pt]
        \includegraphics[width=\linewidth]{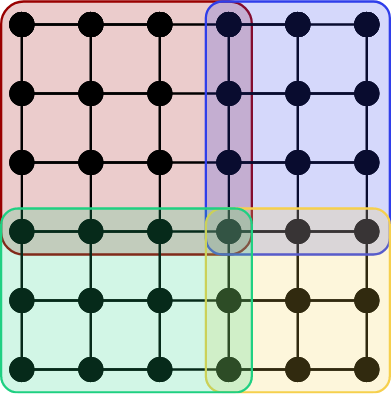}
        \caption{}
        \label{fig:sk-split-a}
    \end{subfigure}\enspace
    \begin{subfigure}[b]{0.24\linewidth}
        \setlength\abovecaptionskip{0.2\baselineskip}
        \centering\footnotesize
        Solve blocks\\*[3.5pt]
        \includegraphics[width=\linewidth]{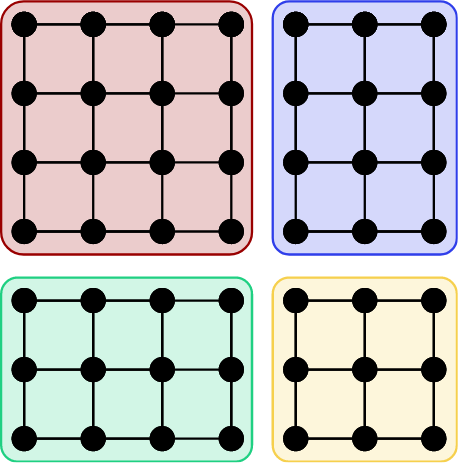}
        \caption{}
        \label{fig:sk-split-b}
    \end{subfigure}\enspace
    \begin{subfigure}[b]{0.24\linewidth}
        \setlength\abovecaptionskip{0.2\baselineskip}
        \centering\footnotesize
        Overlap disagrees\\*[2pt]
        \includegraphics[width=\linewidth]{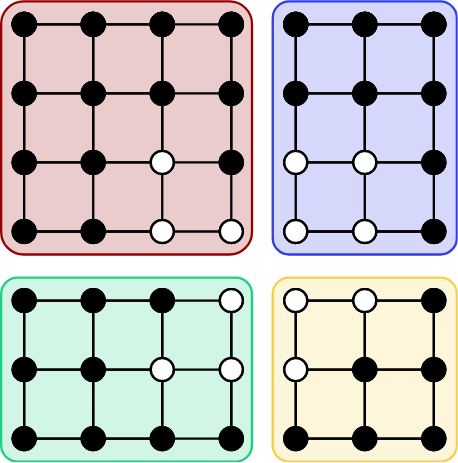}
        \caption{}
        \label{fig:sk-split-c}
    \end{subfigure}\enspace
    \begin{subfigure}[b]{0.24\linewidth}
        \setlength\abovecaptionskip{0.2\baselineskip}
        \centering\footnotesize
        Overlap agrees\\*[2pt]
        \includegraphics[width=\linewidth]{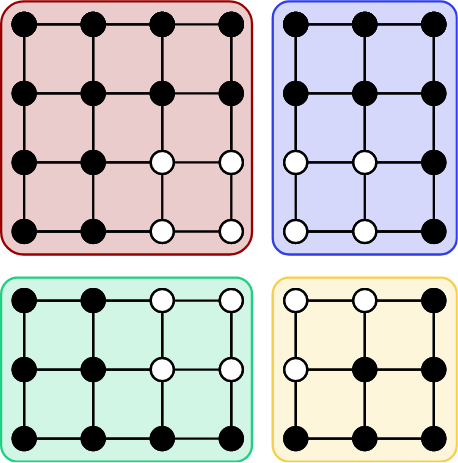}
        \caption{}
        \label{fig:sk-split-d}
    \end{subfigure}
    \caption{\textbf{Illustration of the dual decomposition approach.} Terminal nodes and arcs are not shown. Note that the underlying graph does \emph{not} have to be a grid graph. (a) Graph nodes are divided into a set of overlapping blocks. (b) The graph is split into disjoint sub-graphs and nodes in overlapping regions are duplicated into each blocks. (c) The \maxflow for each block is computed in parallel which gives an assignment to source set (black) or sink set (white). The source/sink capacities are then adjusted for disagreeing duplicated nodes. (d) The \maxflow is recomputed and capacities are adjusted until all duplicated nodes agree.}
    \label{fig:sk-split}
\end{figure}

The dual decomposition (DD) approach was introduced by Strandmark and Kahl \cite{Strandmark2010} and later refined by Yu et al. \cite{Yu2015}. The approach was originally designed to allow for distributed computing, such that it is never necessary to keep the full graph in memory. Their algorithm works as follows: first, the nodes of the graph are divided into a set of overlapping blocks (see \cref{fig:sk-split-a}). The graph is then split into disjoint blocks, where the nodes in the overlapping regions are duplicated in each block (see \cref{fig:sk-split-b}). It is important that the blocks overlap such that if node $i$ is connected to node $j$ in block $b_j$ and node $k$ in block $b_k$, then $i$ is also in both blocks $b_j$ and $b_k$.

Once the graph has been partitioned into overlapping blocks, the algorithm proceeds iteratively. First, the \maxflow for each disjoint block is computed in parallel using the BK algorithm. Next, for each duplicated node, it is checked if all duplicates of that node are in the same \emph{s-t} partitioned set, $S$ or $T$. In that case, we say that the node duplicates agree on their assignment. If all duplicated nodes agree on their assignment, the computed solution is globally optimal and the algorithm terminates. If not, the terminal arc capacities for the disagreeing duplicated nodes are updated according to a supergradient\footnote{Analogous to subgradients for convex functions \cite{Boyd2004}.} ascent scheme and the \maxflow is recomputed. This process of updating terminal capacities and recomputing the \maxflow is repeated until all duplicated nodes agree on their assignment.

A limitation of the original dual decomposition approach is that convergence is not guaranteed. Furthermore, \cite{Yu2015} and \cite{Shekhovtsov2013} have demonstrated that the risk of nonconvergence increases as the graph is split into more blocks. To overcome this, Yu et al. \cite{Yu2015} introduced a new version with a simple strategy that guarantees convergence: if the duplicated nodes in two blocks do not belong to the same set, $S$ or $T$, after a fixed number of iterations, the blocks are merged and the algorithm continues. This trivially guarantees convergence since, in the worst case, all blocks will be merged, at which point the global solution will be computed serially. However, performance significantly drops when merging is needed for the algorithm to converge, as merging only happens after a fixed number of iterations and all blocks may (in the worst case) have to be merged for convergence.

\subsection{Region Discharge}
The region discharge (RD) approach was introduced by \newtext{Delong and Boykov \cite{Delong2008} and later generalized by Shekhovtsov and Hlav{\'a}{\v{c}} \cite{Shekhovtsov2013}. The idea builds on the vertex discharge operation introduced for PPR in \cite{Goldberg1988}.}
Similarly to DD by Strandmark and Kahl, RD was designed to allow for distributed computing. The method first partitions the graph into a set of blocks (called regions in \cite{Shekhovtsov2013} following the terminology of \cite{Delong2008}). Each block $R$ has an associated \emph{boundary} defined as the set of nodes
\begin{equation}
    B^R = \{ v \in V \mid v \notin R, (u, v) \in E, u \in R, v \neq s,t \}.
\end{equation}
Capacities for arcs going from a boundary node to a block node are set to zero. This means that flow can be pushed \emph{out} of the \newtext{block} into the boundary, but not vice versa. Furthermore, each node is allowed to have an excess.

\begin{figure}[ht!]
    \centering
    \begin{subfigure}[b]{0.3\linewidth}
        \setlength\abovecaptionskip{0.2\baselineskip}
        \centering\footnotesize
        Split graph\\*[2pt]
        \includegraphics[width=0.825\linewidth]{figs/split-liusun-1.png}
        \caption{}
    \end{subfigure}\enspace
    \begin{subfigure}[b]{0.3\linewidth}
        \setlength\abovecaptionskip{0.2\baselineskip}
        \centering\footnotesize
        Sync. borders\\*[2pt]
        \includegraphics[width=0.825\linewidth]{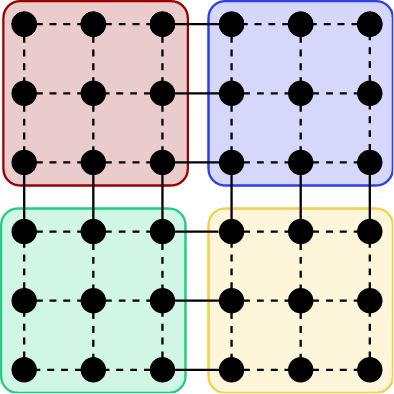}
        \caption{}
    \end{subfigure}
    \begin{subfigure}[b]{0.3\linewidth}
        \setlength\abovecaptionskip{0.2\baselineskip}
        \centering\footnotesize
        Re-solve blocks\\*[3.5pt]
        \includegraphics[width=0.825\linewidth]{figs/split-liusun-1.png}
        \caption{}
    \end{subfigure}\enspace
    \caption{\textbf{Illustration of the region discharge approach.} Terminal nodes and arcs are not shown. Note that the underlying graph does \emph{not} have to be a grid graph. (a) Graph nodes are divided into a set of blocks \newtext{and} the region discharge operation is run on each block, which pushes flow to the sink or boundary. (b) Flow is synchronized between \newtext{boundaries}. (c) Region discharge is run again. The process repeats until no flow crosses the block boundaries.}
    \label{fig:reg-discharge-split}
\end{figure}

The method then performs the \emph{region discharge} operation, which aims to push as much excess flow to the sink \newtext{and/}or the boundary nodes as possible (the source, $s$, is assumed to have infinite excess).
\newtext{This has been done with a PPR \cite{Delong2008, Shekhovtsov2013} or an AP algorithm (specifically BK) \cite{Shekhovtsov2013}.}
When using a PPR algorithm, the discharge of a block is done by performing only push and relabel operations between nodes in the same block.

When using the BK algorithm, \newtext{a distance labeling is maintained for the boundary nodes which gives an estimate of how many boundaries must be crossed to reach the sink. Initially, in each block, flow is pushed exclusively to the sink. Then, flow is pushed to the boundary nodes with distance labels less than 1, then less than 2, etc., until no more flow can be pushed. The BK implementation used by Shekhovtsov and Hlav{\'a}{\v{c}} has been slightly modified to allow excess in the boundary nodes and for flow to be pushed from the boundary nodes out of the block (but not back).}

The discharge operation is performed on all blocks in parallel. Afterward, flow along boundary arcs is synchronized between neighboring blocks. This may create additional excesses in some blocks, since boundary nodes overlap with another block.
The discharge and synchronization process is repeated until no new excesses are created, at which point the algorithm terminates. It is proved in \cite{Shekhovtsov2013} that this process \newtext{terminates} in at most \newtext{$2n^2$ iterations of discharge and synchronization when using PPR and $2n_B^2 + 1$ when using AP}, where $n_B$ is the total number of boundary nodes.

The guarantee of convergence, without having to merge blocks, is beneficial, as it means that the algorithm can maintain a high degree of parallelism while computing the \maxflow solution. However, because flow must be synchronized between blocks, the practical performance of the method still depends on well-chosen blocks and may be limited by synchronization overhead. For details on the heuristics used in the algorithm, which are also important for its practical performance, see \cite{Shekhovtsov2013}.

\section{Performance Comparison}
\label{sec:performance-comparison}
We now compare the performance of the algorithms discussed in the previous sections. For all experiments, the source code was compiled with the GCC C++ compiler version 9.2.0 with \texttt{-O3} optimizations on a 64-bit Linux-based operating system with kernel release 3.10. Experiments were run on a dual socket NUMA (Non-Uniform Memory Access) system with two Intel Xeon Gold 6226R processors with 16 cores each and HTT (Hyper-Threading Technology) disabled, for a total of 32 parallel CPU threads. The system has 756 GB of RAM, and for all experiments all data were kept in memory. All resources were provided by the DTU Computing Center \cite{DTU_DCC_resource}.

For all parallel benchmarks, we prefer local CPU core and memory allocation. This means that for all parallel benchmarks with up to 16 threads, all cores are allocated on the same CPU/NUMA node. If the data fits in the local memory of the active node, we use this memory exclusively. If the data cannot fit in the local memory of one node, memory of both NUMA nodes is used. For benchmarks with more than 16 threads, both CPUs and their memory pools are used.

Run time was measured as the minimum time over three runs and no other processes (apart from the OS) were running during the benchmarks. We split our measured run time into two distinct phases: build time and solve time. Build time refers to the construction of the graph and any additional data structures used by an algorithm. If the algorithm performs arc packing or similar steps, this is included in the build time. To ensure that the build time is a fair representation of the time used by a given algorithm, we precompute a list of nodes and arcs and load these lists fully into memory before starting the timer. Solve time refers to the time required to compute the \maxflow.
For the pseudoflow\newtext{, PPR, and region discharge} algorithms (\cf \cref{tab:mem-reqs}), that only compute a minimum cut,
we do not include the time to extract the full feasible maximum flow solution. The reason for this is that for most computer vision applications the minimum cut is of principal interest. Furthermore, converting to a maximum flow solution usually only adds a small overhead \cite{Verma2012}.

\setlist[enumerate,1]{leftmargin=0.6cm}

\subsection{Datasets}
We test the algorithms on the following benchmark datasets:
\begin{enumerate}[(1)]
    \item The commonly used University of Waterloo \cite{Waterloo} benchmarks problems. Specifically, we use 6 stereo \cite{Boykov1998, Kolmogorov2001}, 36 3D voxel segmentation \cite{Boykov2001, Boykov2003, Boykov2006}, 2 multi-view reconstruction \cite{Lempitsky2006, Boykov2006b}, and 1 surface fitting \cite{Lempitsky2007} problems.

    \item The 4 super resolution \cite{Freeman2000, Rother2007}, 4 texture restoration \cite{Rother2007}, 2 deconvolution \cite{Rother2007}, 78 decision tree field (DTF) \cite{Nowozin2011}, and 3 automatic labelling environment (ALE) \cite{PascalVoc2007, ALE, Ladicky2009, Ladicky2010} datasets from Verma's and Batra's survey \cite{Verma2012}.

    \item New problems that use anisotropic MRFs \cite{Grau2006} to segment blood vessels in large voxel volumes from \cite{Reichardt2021}. We include 3 problems where the segmentation is applied directly to the image data and 3 to the output of a trained V-Net \cite{Milletari2016}.

    \item New problems that use MRFs to clean 3D U-Net \cite{Cciccek2016} segmentations of prostate images from \cite{Simpson2019}. We contribute 4 benchmark problems.

    \item New problems on mesh segmentation based on \cite{Liu2015}. We contribute 8 benchmark problems. The original paper uses $\alpha$-expansion and $\alpha\beta$-swaps \cite{Boykov2001, Boykov2004} to handle the multi-class segmentation problem. For our benchmarks, we instead use QPBO \newtext{to obtain the segmentation with a single min-cut, which may lead to different results compared with the referenced method.}

    \item New problems using the recent Deep LOGISMOS \cite{Guo2018} to segment prostate images from \cite{Simpson2019}. We contribute 8 problems.

    \item New problems performing multi-object image segmentation via surface fitting from two recent papers \cite{Jensen2020, Jeppesen2020}. We contribute 9 problems using \cite{Jensen2020} and 8 using \cite{Jeppesen2020}.

    \item New problems performing graph matching from the recent paper \cite{Hutschenreiter2021}. The original matching problems can be found at \url{https://vislearn.github.io/libmpopt/iccv2021}. For each matching several QPBO sub-problems are solved. We contribute the QPBO subproblems (300 per matching problem) for each of the 316 matching problems.
\end{enumerate}

In total, our benchmark includes 495 problems covering a variety of different computer vision applications.
Note that some datasets consist of many small sub-problems that must be run in sequence. Here, we report the accumulated times. All the benchmark problems are available at: \href{https://doi.org/10.11583/DTU.17091101}{DOI:10.11583/DTU.17091101} \cite{Data}.

For the parallel algorithm benchmarks, we only include a subset of all datasets. This is because parallelization is mainly of interest for large problems with long solve times.
For the block-based algorithms, we split the graph into blocks in one of the following ways:
For graphs based on an underlying image grid, we define blocks by recursively splitting the image grid along its longest axis.
For the surface-based segmentation methods \cite{Jensen2020,Jeppesen2020}, we define blocks such that nodes associated with a surface are in their own block.
For mesh segmentation, we compute the geodesic distance between face centers and then use agglomerative clustering to divide the nodes associated with each face into blocks.
For bottom-up merging, we use 64 blocks for the following dataset: the grid graphs, the mesh segmentation, and the cells, foam, and simcells. For the NT32\_tomo data we use two blocks per object. For 4Dpipe we use a block per 2D slice. For P-GridCut we use the same blocks as for bottom-up merging. For dual decomposition and region discharge, we use one and two blocks per thread, respectively.

\subsection{Tested Implementations}
\label{sec:tested-impl}
All tested implementations (except GridCut \cite{Gridcut2015}) are available at \url{https://github.com/patmjen/maxflow_algorithms} and are archived at \href{https://doi.org/10.5281/zenodo.4903945}{DOI:10.5281/zenodo.4903945} \cite{Code}. Beware that the implementations are published under different licenses --- some open and some restrictive. See the links above for more information.

In the following, \texttt{typewriter} font refers to a specific implementation of a given algorithm. We use this for BK and EIBFS, where we test more that one implementation of each algorithm, \eg, BK refers to the algorithm, \texttt{BK} is the reference implementation, and \texttt{MBK} is one of our implementations.

\textbf{BK} \cite{Boykov2004} \enskip We test the reference implementation (\texttt{BK}) of the Boykov-Kolmogorov algorithm from \url{http://pub.ist.ac.at/~vnk/software.html}. Furthermore, we test our own implementation of BK (\texttt{MBK}), which contains several optimizations. Most notably, our version uses indices instead of pointers to reduce the memory footprint of the \texttt{Node} and \texttt{Arc} data structures. Finally, we test a second version (\texttt{MBK-R}), which reorders arcs so that all outgoing arcs from a node are adjacent in memory. This increases cache efficiency, but uses more memory (see \cref{tab:mem-reqs}) and requires an extra initialization step. The memory overhead from reordering could be reduced by ordering the arcs in-place; however, this may negatively impact performance. Therefore, we opt for the same sorting strategy as \texttt{EIBFS}, where arcs are copied during reordering.

\textbf{EIBFS} \cite{Goldberg2015} \enskip  We test a slightly modified version \cite{Isack2017} (\texttt{EIBFS}) of the excesses incremental breadth first search algorithm originally implemented by \cite{Goldberg2015} available from \url{https://github.com/sydbarrett/AlphaPathMoves}. This version uses slightly larger data structures to support non-integer arc capacities and larger graphs, compared to the implementation tested in \cite{Goldberg2015}. Although these changes may slightly decrease performance, we think it is reasonable to use the modified version, as several of the other algorithms have made similar sacrifices in terms of performance. Additionally, we test our own modified version of \texttt{EIBFS} (\texttt{EIBFS-I}), which replaces pointers with indices to reduce the memory footprint. Finally, since both \texttt{EIBFS} and \texttt{EIBFS-I} perform arc reordering during initialization, we also test a version without arc reordering (\texttt{EIBFS-I-NR}) to better compare with other algorithms.

\textbf{HPF} \cite{Hochbaum2008} \enskip We test the reference implementation of Hochbaum pseudoflow (HPF) from \url{https://riot.ieor.berkeley.edu/Applications/Pseudoflow/maxflow.html}. This implementation has four different configurations that we test:
\begin{enumerate}
    \item Highest label with FIFO buckets (\texttt{HPF-H-F}).
    \item Highest label with LIFO buckets (\texttt{HPF-H-L}).
    \item Lowest label with FIFO buckets (\texttt{HPF-L-F}).
    \item Lowest label with LIFO buckets (\texttt{HPF-L-L}).
\end{enumerate}

\textbf{HI-PR} \cite{Cherkassky1997} \enskip We test the implementation of the preflow push-relabel algorithm from \url{https://cmp.felk.cvut.cz/~shekhovt/d_maxflow/index.html}\footnote{Orignally from \url{http://www.avglab.com/andrew/soft.html}, but the link is no longer available.}.

\textbf{P-ARD} \cite{Shekhovtsov2013} \enskip We test the implementation of parallel augmenting paths region discharge (P-ARD) from \url{https://cmp.felk.cvut.cz/~shekhovt/d_maxflow/index.html}. P-ARD is an example of the region discharge approach. It uses \texttt{BK} as the base solver. Note that, as the implementation is designed for distributed computing, it makes use of disk storage during initialization, which increases the build time.

\textbf{Liu-Sun} \cite{Liu2010} \enskip Since no public reference implementation is available, we test our own implementation of the adaptive bottom-up merging approach based on the paper by Liu and Sun \cite{Liu2010}. Our implementation uses \texttt{MBK} as the base solver.

\textbf{P-PPR} \cite{Baumstark2015} \enskip We test the implementation of a recent parallel preflow push-relabel algorithm from \url{https://github.com/niklasb/pbbs-maxflow}.

\textbf{Strandmark-Kahl} \cite{Strandmark2010} \enskip We test our own implementation of the Strandmark-Kahl dual decomposition algorithm based on the implementation at \url{https://cmp.felk.cvut.cz/~shekhovt/d_maxflow/index.html}\footnote{Originally from \url{https://www1.maths.lth.se/matematiklth/personal/petter/cppmaxflow.php} but the link is no longer available.}. The original implementation can only handle grid graphs with rectangular blocks, while our implementation can handle arbitrary graphs and arbitrary blocks at the cost of some additional overhead during graph construction. Our implementation uses \texttt{MBK} as the base solver. Note that our implementation does not implement the merging strategy proposed by \cite{Yu2015} and, therefore, is not guaranteed to converge. We only include results for cases where the algorithm does converge.

\textbf{GridCut} \cite{Gridcut2015, Jamriska2012} \enskip We test both the serial and parallel versions of the highly optimized commercial GridCut implementation from \url{https://gridcut.com}. The primary goal is to show how much performance can be gained by using an implementation optimized for grid graphs. GridCut is only tested on problems with graph structures that are supported by the reference implementation, \ie, 4- and 8-connected neighbor grids in 2D, and 6- and 26-connected (serial only) neighbor grids in 3D.

\Cref{tab:mem-reqs} lists the tested implementations along with their type and memory footprint. Their memory footprint can be calculated based on the number of nodes and arcs in the graph and will be discussed further in \cref{sec:discussion}.

\newcommand{\myarrow}[1][]{%
  \begin{tikzpicture}[#1]%
    \draw (0,1ex) -- (0,0) -- (0.75em,0);
    \draw (0.55em,0.2em) -- (0.75em,0) -- (0.55em,-0.2em);
  \end{tikzpicture}%
}

\begin{table}[ht!]
    \centering
    \caption{\textbf{Summary of the tested implementations including their memory footprint}. The table shows the bytes required as a function of the number of nodes, $n$, number of terminal arcs, $m_T$, and number of neighbor arcs, $m_N$. We assume the common case of 32-bit capacities and 32-bit indices, which is also what we use for all of our experiments. Since HPF stores undirected arcs, we give all sizes as undirected arcs, \ie, for implementations using directed arcs the size per arc reported here is doubled.
    Note that the numbers depend on, but are \emph{not} the same as, the \texttt{Node} and \texttt{Arc} structure sizes, as the footprint reported includes all stored data (connectivity, capacity, and any auxiliary data).}
    \label{tab:mem-reqs}
    \setlength{\tabcolsep}{2pt}
    \begin{tabularx}{\linewidth}{lXlr}
        \toprule
        \textbf{Serial algorithms} & Algorithm type & \multicolumn{2}{r}{Memory footprint} \\
        \midrule
        HI-PR$^a$ \cite{Cherkassky1997} & Preflow push-relabel & $40 n \: +\: 40 m_T$ & $+\: 40 m_N$ \\
        HPF$^b$ \cite{Hochbaum2008} & Pseudoflow & $104n \:+\: 48 m_T$ & $+\: 48 m_N$ \\ 
        EIBFS \cite{Goldberg2015} & Pseudoflow & & \\ 
        \quad\myarrow \texttt{EIBFS}$^{c}$ & Pseudoflow & $72 n$ & $+\: 72m_N$ \\ 
        \quad\myarrow \texttt{EIBFS-I}$^{*i}$ & Pseudoflow & $29 n$ & $+\: 50m_N$ \\ 
        \quad\myarrow \texttt{EIBFS-I-NR}$^{*i}$ & Pseudoflow & $49n$ & $+\: 24m_N$ \\
        BK \cite{Boykov2004} & Augmenting path & & \\ 
        \quad\myarrow \texttt{BK}$^d$ & Augmenting path & $48n$ & $+\: 64m_N$  \\ 
        \quad\myarrow \texttt{MBK}$^{*i}$ & Augmenting path & $23n$ & $+\: 24m_N$ \\ 
        \quad\myarrow \texttt{MBK-R}$^{*i}$ & Augmenting path & $23n$ & $+\: 48m_N$ \\
        \midrule
        \multicolumn{4}{l}{\textbf{Parallel algorithms}} \\
        \midrule
        P-PPR$^{ei}$ \cite{Baumstark2015} & Parallel PPR & $48n\: +\: 68m_T$ & $+\:68m_N$ \\
        Liu-Sun$^{*i}$ \cite{Liu2010} & Ada. bot.-up merging$^\dagger$ & $25n$ & $+\: 24m_N$ \\ 
        Strandmark-Kahl$^{*i}$ \cite{Strandmark2010} & Dual decomposition$^\dagger$ & $29n$ & $+\: 24m_N$ \\
        P-ARD$^a$ \cite{Shekhovtsov2013} & Region discharge$^\dagger$ & $40n$ & $+\: 32m_N$ \\
        \bottomrule
     \end{tabularx}

    \begin{tabularx}{\linewidth}{X}
        $^\dagger$Uses BK (augmenting path) \\
        $^*$Implemented or updated by us:\\ 
        \quad\quad\url{https://github.com/patmjen/maxflow_algorithms} \\
        $^i$Assuming 32-bit indices \\
        $^a$\url{https://cmp.felk.cvut.cz/~shekhovt/d_maxflow/index.html} \\
        $^b$\url{https://riot.ieor.berkeley.edu/Applications/Pseudoflow/maxflow.html} \\
        $^c$\url{https://github.com/sydbarrett/AlphaPathMoves} \\
        $^d$\url{http://pub.ist.ac.at/~vnk/software.html} \\
        $^e$\url{https://github.com/niklasb/pbbs-maxflow} \\
    \end{tabularx}
\end{table}

\subsection{Serial Algorithms}
\begin{figure}
    \centering
    \begin{subfigure}{\linewidth}
        \setlength\abovecaptionskip{0\baselineskip}
        \centering
        \includegraphics[trim=0pt 6pt 0pt 6pt, clip, width=\linewidth]{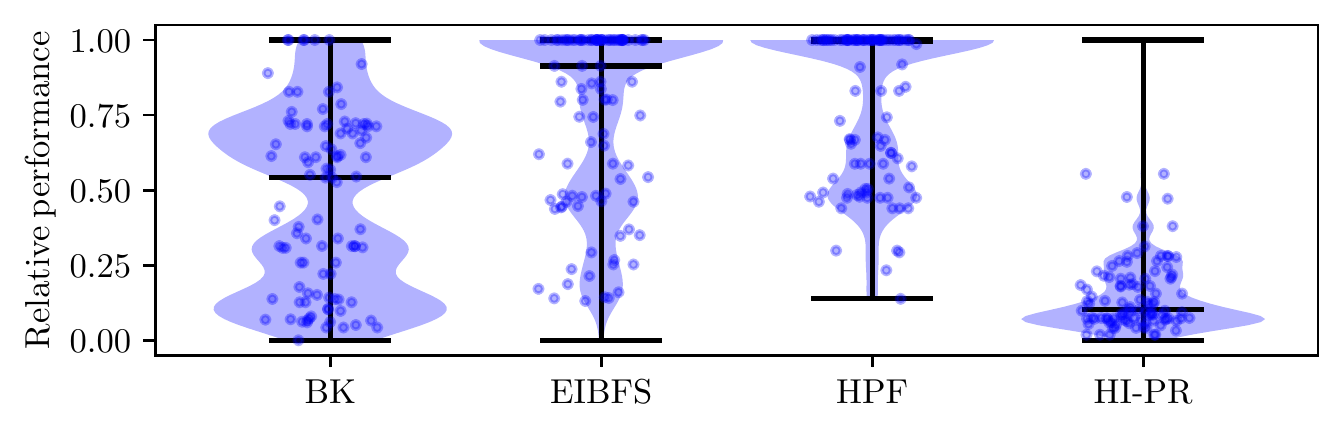}
        \caption{Relative total times.}
        \label{fig:serial_rel_perf_total}
    \end{subfigure}

    \begin{subfigure}{\linewidth}
        \setlength\abovecaptionskip{0\baselineskip}
        \centering
        \includegraphics[trim=0pt 6pt 0pt 6pt, clip, width=\linewidth]{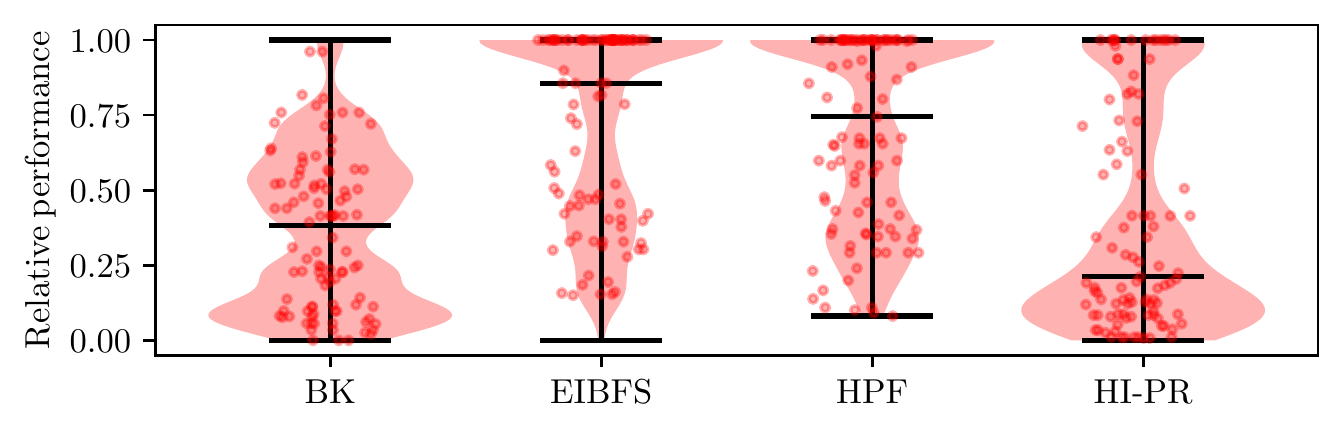}
        \caption{Relative solve times.}
        \label{fig:serial_rel_perf_solve}
    \end{subfigure}
    \caption{\textbf{Relative performance for the serial algorithms.} For each dataset, the solve and total times for each algorithm were compared to those of the fastest algorithm for that dataset and a relative time was computed. This shows how often an algorithm was fastest and, if it was not fastest, how much slower than the fastest it was.  We oversample speed-ups from each problem family (\cf \cref{tab:serial-perf}) so all groups have the same number of entries. This is to avoid bias due to some problem groups having more entries than others. Finally, we overlay a random sample of the (oversampled) speed-ups as jittered points.
    }
    \label{fig:serial_rel_perf}
\end{figure}

The primary experimental results for the serial algorithms are listed in \cref{tab:serial-perf} and \cref{fig:serial_rel_perf}. \cref{tab:serial-perf} shows a representative subset of the results, grouped by problem family, while \cref{fig:serial_rel_perf} shows the distribution of the solve time and the total time for each algorithm on each dataset relative to the fastest algorithm on the dataset. Thus, for a given dataset, a relative performance score of 0.5 means that the algorithm used double the amount of time as the fastest algorithm on that dataset. The distribution of these scores indicates how well the different algorithms perform relative to each other.

\begin{figure*}
    \centering
    \begin{subfigure}{0.33\linewidth}
        \setlength\abovecaptionskip{0\baselineskip}
        \centering
        \includegraphics[trim=0pt 6pt 0pt 6pt, clip, width=\linewidth]{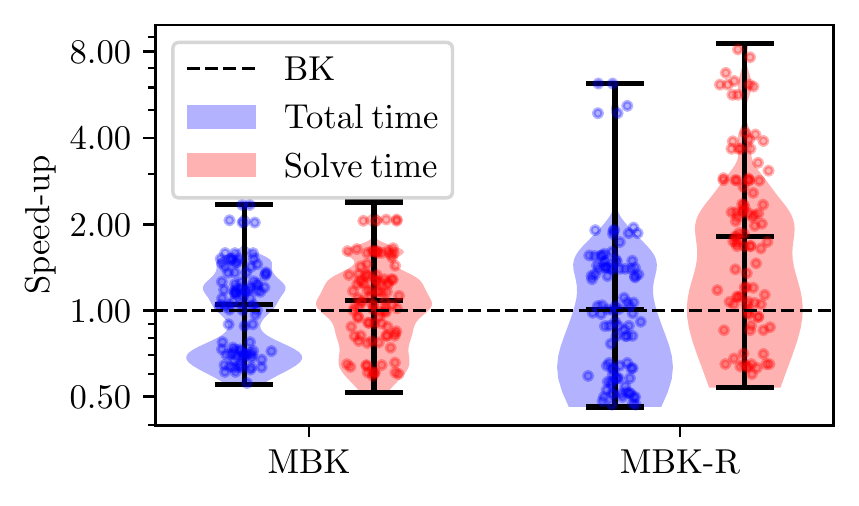}
        \caption{BK}
    \end{subfigure}
    \begin{subfigure}{0.33\linewidth}
        \setlength\abovecaptionskip{0\baselineskip}
        \centering
        \includegraphics[trim=0pt 6pt 0pt 6pt, clip, width=\linewidth]{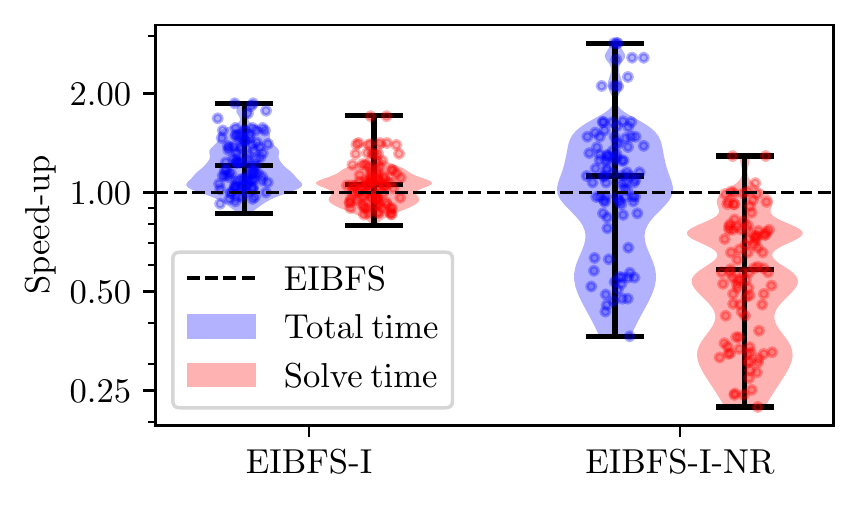}
        \caption{EIBFS}
    \end{subfigure}
    \begin{subfigure}{0.33\linewidth}
        \setlength\abovecaptionskip{0\baselineskip}
        \centering
        \includegraphics[trim=0pt 6pt 0pt 6pt, clip, width=\linewidth]{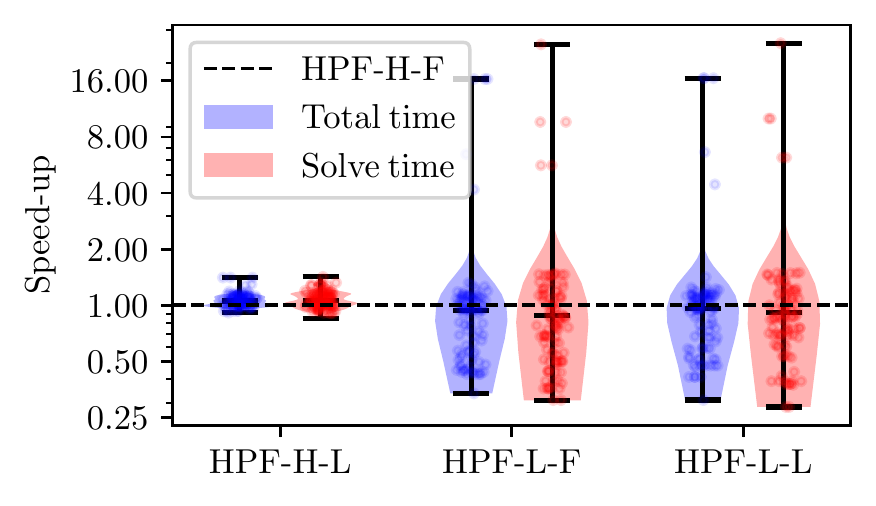}
        \caption{HPF}
    \end{subfigure}
    \caption{\textbf{Performance comparison of serial algorithm variants.} The solve time and total time is compared against the times for the chosen reference algorithm for each dataset. The violin plots show a Gaussian kernel density estimate of the data and the horizontal bars indicate --- from top to bottom --- the maximum, median, and minimum. The values were re-sampled as described in \cref{fig:serial_rel_perf}.}
    \label{fig:serial_variations_comp}
\end{figure*}

From \cref{fig:serial_rel_perf_solve}, we see that EIBFS and HPF outperform the other two algorithms on the majority of the datasets in terms of solve time and total time, as the algorithms have most of their relative times close to 1. Looking at the median, EIBFS has a slightly better relative solve time than HPF, while HPF is faster \wrt total time. Furthermore, HPF has the best worst-case performance for both solve and total time. However, despite its overall good performance, HPF performs significantly worse on the oriented MRF and U-Net cleaning datasets. The performance of BK varies significantly depending on the benchmark problem. Although it has a median relative total time of just over 0.5, its relative performance is considerably more inconsistent than that of the three other algorithms. It performs particularly poorly on the 4Dpipe datasets, using over 6 hours on 4Dpipe\_small, which both HPF and EIBFS completed in less than 30 seconds. For 4Dpipe\_large, BK was not able to find the solution within 45 hours. HI-PR generally has the worst performance but does have the fastest solve time for a few datasets. However, measured on total time, it almost never manages a relative score of more than 0.5. It is worth noting that the distribution of relative times for all algorithms exhibits a bimodality. This indicates that all algorithms have datasets where they are poorly suited compared to the others. We further investigate this in \cref{sec:algo-selection}.

\begin{table*}[ht!]
\centering

\caption{\textbf{Performance comparison of serial algorithms} based on both their solve and total (build + solve) times. We show a representative subset of the datasets, which have been grouped according to their problem family. For each problem family we only show the fastest variant of each algorithm measured in total time. The fastest solve time for each dataset has been \underline{underlined} and the fastest total time has been marked with \textbf{bold face}. 
\newtext{Datasets which contain} many sub-problems are marked with (s).
}
\label{tab:serial-perf}

\setlength\tabcolsep{3.5pt} 

\begin{tabularx}{\linewidth}{Xrrrrrrrrrrrr}
\toprule
Dataset & Nodes & Arcs & Solve & Total & Solve & Total & Solve & Total & Solve & Total & Solve & Total \\
\midrule
\multicolumn{3}{l}{3D segmentation: voxel-based} & \multicolumn{2}{c}{\texttt{MBK-R} \cite{Boykov2004}} & \multicolumn{2}{c}{\texttt{EIBFS-I} \cite{Goldberg2015}} & \multicolumn{2}{c}{\texttt{HPF-H-L} \cite{Hochbaum2008}} & \multicolumn{2}{c}{HI-PR \cite{Cherkassky1997}} & \multicolumn{2}{c}{GridCut \cite{Jamriska2012, Gridcut2015}} \\
\midrule
 adhead.n26c100 \cite{Boykov2001, Boykov2003, Boykov2006} & 12 M & 327 M & 65.81 s & 92.57 s & \underline{22.60 s} & 33.93 s & 24.29 s & 29.03 s & 225.38 s & 424.67 s & 25.19 s & \textbf{27.79 s} \\
 adhead.n6c100 \cite{Boykov2001, Boykov2003, Boykov2006} & 12 M & 75 M & 23.88 s & 28.03 s & 13.23 s & 15.85 s & 14.13 s & 15.87 s & 59.65 s & 102.84 s & \underline{6.98 s} & \textbf{7.31 s} \\
 babyface.n26c100 \cite{Boykov2001, Boykov2003, Boykov2006} & 5 M & 131 M & 82.29 s & 92.87 s & \underline{30.13 s} & \textbf{34.74 s} & 54.47 s & 56.71 s & 183.60 s & 228.09 s & 53.21 s & 54.11 s \\
 babyface.n6c100 \cite{Boykov2001, Boykov2003, Boykov2006} & 5 M & 30 M & 7.78 s & 9.44 s & 5.56 s & 6.61 s & 11.56 s & 12.24 s & 57.28 s & 69.66 s & \underline{2.88 s} & \textbf{3.00 s} \\
 bone.n26c100 \cite{Boykov2001, Boykov2003, Boykov2006} & 7 M & 202 M & 9.01 s & 25.62 s & 9.18 s & 16.28 s & \underline{4.24 s} & 7.16 s & 68.39 s & 173.75 s & 4.52 s & \textbf{5.88 s} \\
 bone.n6c100 \cite{Boykov2001, Boykov2003, Boykov2006} & 7 M & 46 M & 4.09 s & 6.65 s & 2.74 s & 4.35 s & 2.30 s & 3.36 s & 23.66 s & 46.71 s & \underline{0.91 s} & \textbf{1.12 s} \\
 bone\_subx.n6c100 \cite{Boykov2001, Boykov2003, Boykov2006} & 3 M & 23 M & 4.10 s & 5.36 s & 2.38 s & 3.11 s & \underline{1.28 s} & 1.81 s & 10.34 s & 21.49 s & 1.34 s & \textbf{1.44 s} \\
 bone\_subx.n26c100 \cite{Boykov2001, Boykov2003, Boykov2006} & 3 M & 101 M & 7.70 s & 15.78 s & 4.74 s & 8.23 s & \underline{2.14 s} & \textbf{3.61 s} & 25.51 s & 75.15 s & 3.69 s & 4.45 s \\
 liver.n26c100 \cite{Boykov2001, Boykov2003, Boykov2006} & 4 M & 108 M & 11.78 s & 20.41 s & 10.49 s & 14.20 s & 5.72 s & 6.50 s & 71.88 s & 131.00 s & \underline{5.62 s} & \textbf{6.21 s} \\
 liver.n6c100 \cite{Boykov2001, Boykov2003, Boykov2006} & 4 M & 25 M & 10.08 s & 11.40 s & 5.82 s & 6.57 s & 5.70 s & 6.24 s & 30.49 s & 42.71 s & \underline{3.87 s} & \textbf{3.99 s} \\
\midrule
\multicolumn{3}{l}{3D segmentation: oriented MRF} & \multicolumn{2}{c}{\texttt{MBK} \cite{Boykov2004}} & \multicolumn{2}{c}{\texttt{EIBFS-I-NR} \cite{Goldberg2015}} & \multicolumn{2}{c}{\texttt{HPF-H-L} \cite{Hochbaum2008}} & \multicolumn{2}{c}{HI-PR \cite{Cherkassky1997}} & \multicolumn{2}{c}{GridCut \cite{Jamriska2012, Gridcut2015}} \\
\midrule
 vessel.orimrf.256 \cite{Boykov2004, Grau2006, Reichardt2021} & 16 M & 66 M & 1.84 s & 2.95 s & 1.13 s & 2.03 s & 3.19 s & 6.80 s & 4.11 s & 30.99 s & \underline{0.40 s} & \textbf{1.04 s} \\
 vessel.orimrf.512 \cite{Boykov2004, Grau2006, Reichardt2021} & 134 M & 536 M & 12.44 s & 21.40 s & 7.95 s & 15.39 s & 25.29 s & 55.32 s & 32.16 s & 321.75 s & \underline{2.43 s} & \textbf{7.73 s} \\
 vessel.orimrf.900 \cite{Boykov2004, Grau2006, Reichardt2021} & 688 M & 2 B & 75.23 s & 121.82 s & 48.13 s & 88.09 s & 147.22 s & 300.79 s & 177.38 s & 1774.65 s & \underline{15.97 s} & \textbf{44.70 s} \\
\midrule
\multicolumn{3}{l}{3D U-Net segmentation cleaning} & \multicolumn{2}{c}{\texttt{MBK} \cite{Boykov2004}} & \multicolumn{2}{c}{\texttt{EIBFS-I-NR} \cite{Goldberg2015}} & \multicolumn{2}{c}{\texttt{HPF-H-L} \cite{Hochbaum2008}} & \multicolumn{2}{c}{HI-PR \cite{Cherkassky1997}} & \multicolumn{2}{c}{GridCut \cite{Jamriska2012, Gridcut2015}} \\
\midrule
 clean.orimrf.256 \cite{Boykov2004, Grau2006, Reichardt2021} & 16 M & 66 M & 0.97 s & 2.09 s & 0.69 s & 1.61 s & 3.21 s & 6.89 s & 3.93 s & 30.83 s & \underline{0.13 s} & \textbf{0.77 s} \\
 clean.orimrf.512 \cite{Boykov2004, Grau2006, Reichardt2021} & 134 M & 536 M & 7.87 s & 17.03 s & 5.51 s & 13.51 s & 27.10 s & 58.22 s & 31.40 s & 320.87 s & \underline{0.91 s} & \textbf{6.27 s} \\
 clean.orimrf.900 \cite{Boykov2004, Grau2006, Reichardt2021} & 688 M & 2 B & 35.83 s & 81.92 s & 25.96 s & 64.22 s & 130.22 s & 280.73 s & 163.88 s & 1755.87 s & \underline{3.90 s} & \textbf{31.43 s} \\
 unet\_mrfclean\_2 \cite{Boykov2004} & 8 M & 32 M & 0.47 s & 1.01 s & 0.29 s & 0.74 s & 3.55 s & 5.36 s & 9.80 s & 22.82 s & \underline{62 ms} & \textbf{0.36 s} \\
 unet\_mrfclean\_3 \cite{Boykov2004} & 15 M & 63 M & 0.82 s & 1.88 s & 0.52 s & 1.37 s & 5.68 s & 9.14 s & 20.59 s & 46.69 s & \underline{0.11 s} & \textbf{0.68 s} \\
 unet\_mrfclean\_8 \cite{Boykov2004} & 4 M & 19 M & 0.48 s & 0.81 s & 0.24 s & 0.50 s & 2.39 s & 3.46 s & 6.55 s & 13.89 s & \underline{0.11 s} & \textbf{0.28 s} \\
\midrule
\multicolumn{3}{l}{Surface fitting} & \multicolumn{2}{c}{\texttt{MBK} \cite{Boykov2004}} & \multicolumn{2}{c}{\texttt{EIBFS-I} \cite{Goldberg2015}} & \multicolumn{2}{c}{\texttt{HPF-H-L} \cite{Hochbaum2008}} & \multicolumn{2}{c}{HI-PR \cite{Cherkassky1997}} & \multicolumn{2}{c}{GridCut \cite{Jamriska2012, Gridcut2015}} \\
\midrule
 LB07-bunny-lrg \cite{Lempitsky2007} & 49 M & 300 M & 15.40 s & 21.17 s & 6.38 s & 15.25 s & 21.87 s & 32.13 s & 610.24 s & 820.64 s & \underline{2.36 s} & \textbf{3.75 s} \\
\midrule
\multicolumn{3}{l}{3D segmentation: sparse layered graphs (SLG)} & \multicolumn{2}{c}{\texttt{MBK-R} \cite{Boykov2004}} & \multicolumn{2}{c}{\texttt{EIBFS-I} \cite{Goldberg2015}} & \multicolumn{2}{c}{\texttt{HPF-H-L} \cite{Hochbaum2008}} & \multicolumn{2}{c}{HI-PR \cite{Cherkassky1997}} & \multicolumn{2}{c}{GridCut \cite{Jamriska2012, Gridcut2015}} \\
\midrule
 4Dpipe\_small \cite{Jeppesen2020} & 14 M & 124 M & 6.03 h & 6.03 h & \underline{2.06 s} & \textbf{15.55 s} & 17.91 s & 28.85 s & 202.49 s & 266.01 s & - & - \\
 4Dpipe\_big \cite{Jeppesen2020} & 143 M & 1 B & - & - & \underline{20.59 s} & \textbf{195.41 s} & 222.09 s & 332.06 s & 2611.65 s & 3436.43 s & - & - \\
 NT32\_tomo3\_.raw\_3 \cite{Jeppesen2020} & 7 M & 49 M & \underline{15.42 s} & 18.69 s & 24.22 s & 27.19 s & 15.87 s & \textbf{18.33 s} & 176.11 s & 200.29 s & - & - \\
 NT32\_tomo3\_.raw\_10 \cite{Jeppesen2020} & 22 M & 154 M & 52.86 s & 63.15 s & 50.82 s & 60.01 s & \underline{36.46 s} & \textbf{44.14 s} & 645.33 s & 741.98 s & - & - \\
 NT32\_tomo3\_.raw\_30 \cite{Jeppesen2020} & 67 M & 462 M & \underline{145.23 s} & \textbf{176.37 s} & 194.79 s & 221.90 s & 179.82 s & 202.73 s & 2939.04 s & 3260.63 s & - & - \\
 NT32\_tomo3\_.raw\_100 \cite{Jeppesen2020} & 183 M & 1 B & 778.39 s & 860.71 s & 553.50 s & 627.08 s & \underline{520.26 s} & \textbf{583.76 s} & 9732.34 s & 2.95 h & - & - \\
\midrule
\multicolumn{3}{l}{3D segmentation: seperating surfaces} & \multicolumn{2}{c}{\texttt{MBK-R} \cite{Boykov2004}} & \multicolumn{2}{c}{\texttt{EIBFS-I} \cite{Goldberg2015}} & \multicolumn{2}{c}{\texttt{HPF-H-L} \cite{Hochbaum2008}} & \multicolumn{2}{c}{HI-PR \cite{Cherkassky1997}} & \multicolumn{2}{c}{GridCut \cite{Jamriska2012, Gridcut2015}} \\
\midrule
 cells.sd3 \cite{Jensen2020} & 13 M & 126 M & 48.23 s & 59.03 s & 35.24 s & 40.66 s & \underline{15.52 s} & \textbf{21.84 s} & 98.25 s & 167.56 s & - & - \\
 foam.subset.r160.h210 \cite{Jensen2020} & 15 M & 205 M & 6.05 s & 22.02 s & \underline{3.21 s} & \textbf{12.52 s} & 17.14 s & 26.18 s & 15.16 s & 145.58 s & - & - \\
 foam.subset.r60.h210 \cite{Jensen2020} & 1 M & 24 M & 0.62 s & 2.58 s & \underline{0.39 s} & \textbf{1.49 s} & 1.98 s & 3.01 s & 1.85 s & 12.82 s & - & - \\
 simcells.sd3 \cite{Jensen2020} & 3 M & 27 M & 9.93 s & 12.10 s & \underline{2.94 s} & \textbf{4.12 s} & 3.23 s & 4.60 s & 21.57 s & 33.89 s & - & - \\
\midrule
\multicolumn{3}{l}{Deep LOGISMOS} & \multicolumn{2}{c}{\texttt{MBK} \cite{Boykov2004}} & \multicolumn{2}{c}{\texttt{EIBFS-I} \cite{Goldberg2015}} & \multicolumn{2}{c}{\texttt{HPF-H-F} \cite{Hochbaum2008}} & \multicolumn{2}{c}{HI-PR \cite{Cherkassky1997}} & \multicolumn{2}{c}{GridCut \cite{Jamriska2012, Gridcut2015}} \\
\midrule
 deeplogismos.2 \cite{Guo2018} & 511 K & 4 M & 0.15 s & 0.25 s & \underline{28 ms} & \textbf{0.21 s} & 0.12 s & 0.31 s & 0.16 s & 1.29 s & - & - \\
 deeplogismos.3 \cite{Guo2018} & 707 K & 5 M & 0.18 s & 0.31 s & \underline{41 ms} & \textbf{0.30 s} & 0.18 s & 0.45 s & 0.24 s & 1.90 s & - & - \\
 deeplogismos.7 \cite{Guo2018} & 989 K & 7 M & 0.34 s & \textbf{0.54 s} & \underline{0.26 s} & 0.66 s & 0.29 s & 0.69 s & 0.36 s & 2.86 s & - & - \\
\midrule
\multicolumn{3}{l}{Super resolution} & \multicolumn{2}{c}{\texttt{BK} \cite{Boykov2004}} & \multicolumn{2}{c}{\texttt{EIBFS-I} \cite{Goldberg2015}} & \multicolumn{2}{c}{\texttt{HPF-H-L} \cite{Hochbaum2008}} & \multicolumn{2}{c}{HI-PR \cite{Cherkassky1997}} & \multicolumn{2}{c}{GridCut \cite{Jamriska2012, Gridcut2015}} \\
\midrule
 super\_res-E1 \cite{Freeman2000, Rother2007} & 10 K & 62 K & 2 ms & 2 ms & \underline{1 ms} & \textbf{2 ms} & 2 ms & 3 ms & 1 ms & 7 ms & - & - \\
 super\_res-E2 \cite{Freeman2000, Rother2007} & 10 K & 103 K & 4 ms & 5 ms & \underline{2 ms} & 3 ms & 2 ms & \textbf{3 ms} & 2 ms & 12 ms & - & - \\
 super\_res-Paper1 \cite{Freeman2000, Rother2007} & 10 K & 62 K & 2 ms & 3 ms & \underline{1 ms} & \textbf{2 ms} & 2 ms & 3 ms & 1 ms & 7 ms & - & - \\
 superres\_graph \cite{Freeman2000, Rother2007} & 43 K & 742 K & 62 ms & 78 ms & 10 ms & 26 ms & \underline{7 ms} & \textbf{12 ms} & 19 ms & 0.16 s & - & - \\
\midrule
\multicolumn{3}{l}{Texture} & \multicolumn{2}{c}{\texttt{MBK-R} \cite{Boykov2004}} & \multicolumn{2}{c}{\texttt{EIBFS-I} \cite{Goldberg2015}} & \multicolumn{2}{c}{\texttt{HPF-H-L} \cite{Hochbaum2008}} & \multicolumn{2}{c}{HI-PR \cite{Cherkassky1997}} & \multicolumn{2}{c}{GridCut \cite{Jamriska2012, Gridcut2015}} \\
\midrule
 texture-Cremer \cite{Rother2007} & 44 K & 783 K & 1.54 s & 1.58 s & 0.35 s & 0.37 s & 0.17 s & 0.19 s & \underline{42 ms} & \textbf{0.19 s} & - & - \\
 texture-OLD-D103 \cite{Rother2007} & 43 K & 742 K & 0.60 s & 0.65 s & 0.19 s & 0.21 s & 73 ms & \textbf{92 ms} & \underline{41 ms} & 0.19 s & - & - \\
 texture-Paper1 \cite{Rother2007} & 43 K & 742 K & 0.65 s & 0.69 s & 0.19 s & 0.21 s & 76 ms & \textbf{95 ms} & \underline{36 ms} & 0.17 s & - & - \\
 texture-Temp \cite{Rother2007} & 14 K & 239 K & 0.22 s & 0.23 s & 30 ms & 34 ms & 9 ms & \textbf{15 ms} & \underline{6 ms} & 32 ms & - & - \\
\midrule
\multicolumn{3}{l}{Automatic labelling envrionment (ALE)} & \multicolumn{2}{c}{\texttt{MBK-R} \cite{Boykov2004}} & \multicolumn{2}{c}{\texttt{EIBFS-I-NR} \cite{Goldberg2015}} & \multicolumn{2}{c}{\texttt{HPF-L-L} \cite{Hochbaum2008}} & \multicolumn{2}{c}{HI-PR \cite{Cherkassky1997}} & \multicolumn{2}{c}{GridCut \cite{Jamriska2012, Gridcut2015}} \\
\midrule
 graph\_1 (s) \cite{Ladicky2009, Ladicky2010, PascalVoc2010, ALE} & 185 K & 5 M & 16.80 s & 18.52 s & \underline{0.35 s} & \textbf{0.79 s} & 1.00 s & 1.60 s & 1.58 s & 10.60 s & - & - \\
 graph\_2 (s) \cite{Ladicky2009, Ladicky2010, PascalVoc2010, ALE} & 175 K & 3 M & 7.38 s & 10.47 s & \underline{0.83 s} & \textbf{1.64 s} & 2.25 s & 3.55 s & 2.91 s & 20.87 s & - & - \\
 graph\_3 (s) \cite{Ladicky2009, Ladicky2010, PascalVoc2010, ALE} & 179 K & 7 M & 27.68 s & 35.55 s & \underline{2.69 s} & \textbf{4.51 s} & 4.63 s & 6.96 s & 6.49 s & 43.73 s & - & - \\
 \midrule
 \multicolumn{3}{l}{Multi-view} & \multicolumn{2}{c}{\texttt{MBK-R} \cite{Boykov2004}} & \multicolumn{2}{c}{\texttt{EIBFS-I} \cite{Goldberg2015}} & \multicolumn{2}{c}{\texttt{HPF-H-L} \cite{Hochbaum2008}} & \multicolumn{2}{c}{HI-PR \cite{Cherkassky1997}} & \multicolumn{2}{c}{GridCut \cite{Jamriska2012, Gridcut2015}} \\
 \midrule
 BL06-camel-lrg \cite{Boykov2006b} & 18 M & 93 M & 107.53 s & 111.42 s & 28.55 s & 31.54 s & \underline{24.44 s} & \textbf{28.82 s} & 291.71 s & 337.91 s & - & - \\
 BL06-gargoyle-lrg \cite{Boykov2006b} & 17 M & 86 M & 238.08 s & 241.65 s & 33.76 s & 36.57 s & \underline{26.51 s} & \textbf{30.61 s} & 208.27 s & 251.10 s & - & - \\
\bottomrule
\end{tabularx}
\end{table*}

\begin{table*}
\setlength\tabcolsep{5pt} 
\caption*{TABLE \ref{tab:serial-perf}: Continued}
\centering
\begin{tabularx}{\linewidth}{Xrrrrrrrrrrrr}
\toprule
Dataset & Nodes & Arcs & Solve & Total & Solve & Total & Solve & Total & Solve & Total & Solve & Total \\
\midrule
\multicolumn{3}{l}{Deconvolution} & \multicolumn{2}{c}{\texttt{MBK-R} \cite{Boykov2004}} & \multicolumn{2}{c}{\texttt{EIBFS-I} \cite{Goldberg2015}} & \multicolumn{2}{c}{\texttt{HPF-H-L} \cite{Hochbaum2008}} & \multicolumn{2}{c}{HI-PR \cite{Cherkassky1997}} & \multicolumn{2}{c}{GridCut \cite{Jamriska2012, Gridcut2015}} \\
\midrule
 graph3x3 \cite{Rother2007} & 2 K & 47 K & 9 ms & 11 ms & 3 ms & 3 ms & 1 ms & \textbf{1 ms} & \underline{1 ms} & 5 ms & - & - \\
 graph5x5 \cite{Rother2007} & 2 K & 139 K & 62 ms & 67 ms & 6 ms & 9 ms & 3 ms & \textbf{4 ms} & \underline{2 ms} & 15 ms & - & - \\
\midrule
\multicolumn{3}{l}{Stereo 1} & \multicolumn{2}{c}{\texttt{BK} \cite{Boykov2004}} & \multicolumn{2}{c}{\texttt{EIBFS-I-NR} \cite{Goldberg2015}} & \multicolumn{2}{c}{\texttt{HPF-H-L} \cite{Hochbaum2008}} & \multicolumn{2}{c}{HI-PR \cite{Cherkassky1997}} & \multicolumn{2}{c}{GridCut \cite{Jamriska2012, Gridcut2015}} \\
\midrule
 BVZ-sawtooth (s) \cite{Boykov1998} & 164 K & 796 K & 0.91 s & 1.16 s & \underline{0.58 s} & \textbf{0.69 s} & 1.39 s & 1.85 s & 7.89 s & 12.27 s & - & - \\
 BVZ-tsukuba (s) \cite{Boykov1998} & 110 K & 513 K & 0.49 s & 0.58 s & \underline{0.35 s} & \textbf{0.41 s} & 0.66 s & 0.84 s & 4.69 s & 6.64 s & - & - \\
 BVZ-venus (s) \cite{Boykov1998} & 166 K & 795 K & 1.72 s & 2.03 s & \underline{1.30 s} & \textbf{1.44 s} & 1.94 s & 2.46 s & 15.00 s & 20.11 s & - & - \\
\midrule
\multicolumn{3}{l}{Stereo 2} & \multicolumn{2}{c}{\texttt{BK} \cite{Boykov2004}} & \multicolumn{2}{c}{\texttt{EIBFS-I} \cite{Goldberg2015}} & \multicolumn{2}{c}{\texttt{HPF-H-L} \cite{Hochbaum2008}} & \multicolumn{2}{c}{HI-PR \cite{Cherkassky1997}} & \multicolumn{2}{c}{GridCut \cite{Jamriska2012, Gridcut2015}} \\
\midrule
 KZ2-sawtooth (s) \cite{Kolmogorov2001} & 294 K & 1 M & 2.59 s & 3.40 s & \underline{1.14 s} & \textbf{2.02 s} & 3.30 s & 4.66 s & 23.79 s & 36.55 s & - & - \\
 KZ2-tsukuba (s) \cite{Kolmogorov2001} & 199 K & 1 M & 1.41 s & 1.84 s & \underline{0.71 s} & \textbf{1.12 s} & 1.92 s & 2.55 s & 20.95 s & 27.14 s & - & - \\
 KZ2-venus (s) \cite{Kolmogorov2001} & 301 K & 2 M & 3.98 s & 4.89 s & \underline{2.18 s} & \textbf{3.16 s} & 4.70 s & 6.21 s & 41.63 s & 55.60 s & - & - \\
\midrule
\multicolumn{3}{l}{Decision tree field (DTF)} & \multicolumn{2}{c}{\texttt{MBK-R} \cite{Boykov2004}} & \multicolumn{2}{c}{\texttt{EIBFS-I} \cite{Goldberg2015}} & \multicolumn{2}{c}{\texttt{HPF-H-L} \cite{Hochbaum2008}} & \multicolumn{2}{c}{HI-PR \cite{Cherkassky1997}} & \multicolumn{2}{c}{GridCut \cite{Jamriska2012, Gridcut2015}} \\
\midrule
 printed\_graph1 \cite{Nowozin2011} & 20 K & 1 M & 0.63 s & 0.73 s & 0.13 s & 0.17 s & \underline{40 ms} & \textbf{51 ms} & 51 ms & 0.25 s & - & - \\
 printed\_graph16 \cite{Nowozin2011} & 11 K & 683 K & 0.24 s & 0.29 s & 44 ms & 62 ms & \underline{16 ms} & \textbf{22 ms} & 25 ms & 0.12 s & - & - \\
\midrule
\multicolumn{3}{l}{Graph matching: small} & \multicolumn{2}{c}{\texttt{BK} \cite{Boykov2004}} & \multicolumn{2}{c}{\texttt{EIBFS-I-NR} \cite{Goldberg2015}} & \multicolumn{2}{c}{\texttt{HPF-L-L} \cite{Hochbaum2008}} & \multicolumn{2}{c}{HI-PR \cite{Cherkassky1997}} & \multicolumn{2}{c}{GridCut \cite{Jamriska2012, Gridcut2015}} \\
\midrule
 atlas1.dd (s) \cite{Kainmueller2014, Hutschenreiter2021} & 1 K & 5 K & 37 ms & 59 ms & 34 ms & 52 ms & \underline{21 ms} & \textbf{39 ms} & 23 ms & 0.18 s & - & - \\
 car1.dd (s) \cite{PascalVoc2007, Leordeanu2012, Hutschenreiter2021} & 38 & 131 & 1 ms & 1 ms & 0 ms & \textbf{1 ms} & 1 ms & 1 ms & \underline{0 ms} & 3 ms & - & - \\
 hassan1.dd (s) \cite{Alhaija2015, Swoboda2017, Hutschenreiter2021} & 120 & 2 K & 17 ms & 30 ms & 5 ms & 25 ms & \underline{2 ms} & \textbf{6 ms} & 4 ms & 65 ms & - & - \\
 matching1.dd (s) \cite{Komodakis2008, Kappes2015, Hutschenreiter2021} & 38 & 380 & 10 ms & 12 ms & 5 ms & 8 ms & \underline{2 ms} & \textbf{4 ms} & 6 ms & 14 ms & - & - \\
\midrule
\multicolumn{3}{l}{Graph matching: big} & \multicolumn{2}{c}{\texttt{MBK-R} \cite{Boykov2004}} & \multicolumn{2}{c}{\texttt{EIBFS-I} \cite{Goldberg2015}} & \multicolumn{2}{c}{\texttt{HPF-H-L} \cite{Hochbaum2008}} & \multicolumn{2}{c}{HI-PR \cite{Cherkassky1997}} & \multicolumn{2}{c}{GridCut \cite{Jamriska2012, Gridcut2015}} \\
\midrule
 pair1.dd (s) \cite{Hutschenreiter2021} & 1 K & 58 K & 1.42 s & 1.97 s & 0.70 s & 0.96 s & \underline{92 ms} & \textbf{0.13 s} & 0.82 s & 1.68 s & - & - \\
\midrule
\multicolumn{3}{l}{Mesh segmentation} & \multicolumn{2}{c}{\texttt{MBK-R} \cite{Boykov2004}} & \multicolumn{2}{c}{\texttt{EIBFS-I} \cite{Goldberg2015}} & \multicolumn{2}{c}{\texttt{HPF-H-F} \cite{Hochbaum2008}} & \multicolumn{2}{c}{HI-PR \cite{Cherkassky1997}} & \multicolumn{2}{c}{GridCut \cite{Jamriska2012, Gridcut2015}} \\
\midrule
 bunny.segment \cite{Liu2015} & 97 K & 536 K & 0.12 s & 0.14 s & \underline{63 ms} & \textbf{75 ms} & 68 ms & 91 ms & 0.20 s & 0.30 s & - & - \\
 bunnybig.segment \cite{Liu2015} & 2 M & 13 M & 1.01 s & 1.59 s & \underline{0.62 s} & \textbf{1.23 s} & 1.43 s & 2.12 s & 4.99 s & 9.91 s & - & - \\
 candle.segment \cite{Liu2015} & 159 K & 959 K & 87 ms & 0.13 s & \underline{49 ms} & \textbf{83 ms} & 0.11 s & 0.15 s & 0.29 s & 0.53 s & - & - \\
 candlebig.segment \cite{Liu2015} & 1 M & 5 M & 0.51 s & 0.72 s & \underline{0.26 s} & \textbf{0.44 s} & 0.60 s & 0.91 s & 2.03 s & 3.70 s & - & - \\
 chair.segment \cite{Liu2015} & 305 K & 1 M & 0.76 s & 0.88 s & 0.31 s & 0.39 s & \underline{0.27 s} & \textbf{0.37 s} & 0.86 s & 1.37 s & - & - \\
 chairbig.segment \cite{Liu2015} & 3 M & 26 M & 1.62 s & 2.89 s & \underline{1.02 s} & \textbf{2.45 s} & 3.23 s & 4.59 s & 9.98 s & 20.92 s & - & - \\
 handbig.segment \cite{Liu2015} & 248 K & 1 M & 0.15 s & 0.19 s & \underline{71 ms} & \textbf{0.11 s} & 0.13 s & 0.18 s & 0.35 s & 0.63 s & - & - \\
 handsmall.segment \cite{Liu2015} & 15 K & 69 K & 4 ms & 5 ms & \underline{2 ms} & \textbf{3 ms} & 4 ms & 6 ms & 10 ms & 16 ms & - & - \\
\bottomrule
\end{tabularx}

\end{table*}

\subsubsection{Algorithm Variants}

The different variants of each algorithm are compared in \cref{fig:serial_variations_comp}, which shows the relative performance of each implementation compared to a chosen ``reference'' implementation. For the BK algorithm, the \texttt{BK} implementation is used for reference, for the EIBFS algorithm, the \texttt{EIBFS} implementation is used as a reference, and for HPF the \texttt{HPF-H-F} configuration is used as reference, since it is the one recommended by the authors. As we are now measuring relative to a specific implementation, rather than the fastest implementation as in \cref{fig:serial_rel_perf}, it is possible to obtain a relative performance score of more than one.

For the BK algorithm, both \texttt{MBK} and \texttt{MBK-R} overall perform similarly or slightly better than \texttt{BK}, when measured on total time. Looking at solve time, \texttt{MBK-R} shows a large speed-up over the other variants.
This clearly reflects the effect of arc packing (reordering the arcs), in that it typically decreases solve at the cost of increased build time.
From \cref{tab:serial-perf}, we see that \texttt{BK} is generally best for smaller problems where the smaller memory footprint of the index-based variants is less of an advantage. However, the very small difference in absolute time for these small problems will in many cases render the choice of algorithm irrelevant.

For the EIBFS variants, the index-based version (\texttt{EIBFS-I}) consistently outperforms the reference implementation with a median improvement of more than 20\%.
Meanwhile, \texttt{EIBFS-I-NR} performs worse than \texttt{EIBFS} on almost all datasets \wrt to solve time, but better \wrt total time for the majority of the problems. In some cases, it also outperforms \texttt{EIBFS-I}, again showing that while arc packing generally significantly reduces the solve time, the additional overhead is not always worth it.

For the HPF algorithm, \texttt{HPF-H-L} consistently performs the best, while \texttt{HPF-L-F} and\texttt{HPF-L-L} perform worse than the reference \texttt{HPF-H-F} for the majority of datasets. However, for some datasets \texttt{HPF-L-F} and \texttt{HPF-L-L} show large speed-ups over the other variants. \cref{tab:serial-perf} reveals that the \texttt{HPF-L} variants seem to be better for graph matching and ALE datasets.

\subsection{Parallel Algorithms}
Our benchmark results for the parallel algorithms are shown in \cref{tab:parallel-perf}, where we compare the build and solve time for each algorithm on each dataset.
The table includes the number of CPU threads used by each algorithm for the listed solve times. Furthermore, it includes the solve time of the best serial algorithm for each dataset for comparison.
We focus on the solve time, as that is what reveals how successfully the algorithms distribute the work as more threads are added.
Additionally, a lack of optimization leads to very long build times for some of the parallel implementations, especially P-PPR and P-ARD.
\newtext{Finally, some datasets are omitted for P-PPR due to run-time errors and for Strandmark-Kahl due to excessive run time.}

From \cref{tab:parallel-perf}, it is clear that no algorithm is dominant, except P-GridCut for 6-connected grid graphs. Every algorithm has datasets where it is the fastest and
\newtext{a serial algorithm often} gives the best or close to the best performance.
The parallel algorithms show their strength for the large datasets with more than 1~M nodes where significant performance improvements are found.
Curiously, only P-ARD shows a significant speed-up for smaller problems.

\begin{figure}
    \centering
    \includegraphics[trim=0pt 6pt 0pt 6pt, clip, width=\linewidth]{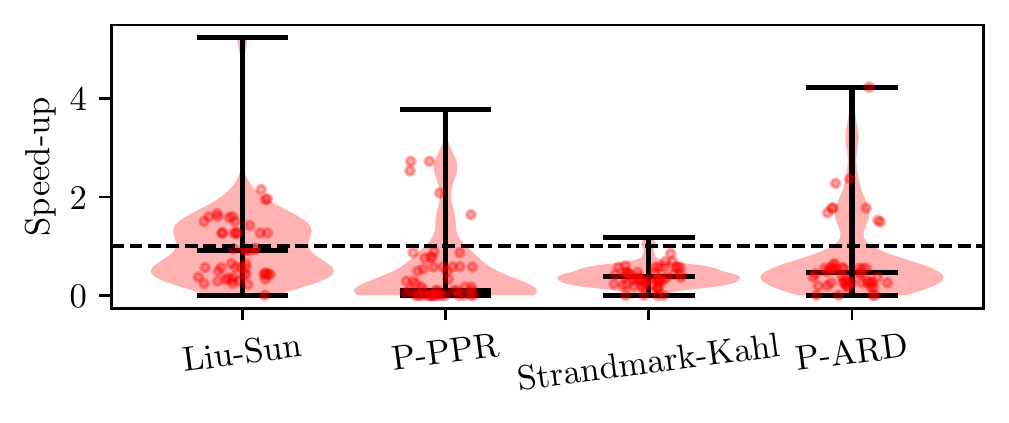}
    \caption{Speed-up of the parallel algorithms relative to the best serial solve time for each dataset. The values were re-sampled as described in \cref{fig:serial_rel_perf}.}
    \label{fig:parallel_rel_perf_solve}
\end{figure}

The parallel benchmarks are also summarized in \cref{fig:parallel_rel_perf_solve}. All algorithms have median speed-ups less than one. Liu-Sun generally performs best, giving a speed-up for almost half of the dataset and having the largest maximum speed-up. P-PPR and P-ARD still provide good speed-ups for some datasets. Strandmark-Kahl comes off the worst, as it rarely beats the best serial algorithm.

\begin{figure}
    \centering
    \setlength{\tabcolsep}{0pt}
    \begin{tabular}{rr}
        \includegraphics[trim=0pt 6pt 0pt 6pt, clip, height=0.36\linewidth]{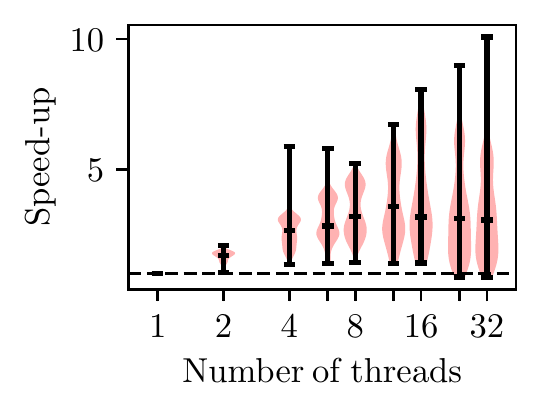} &
        \includegraphics[trim=0pt 6pt 0pt 6pt, clip, height=0.36\linewidth]{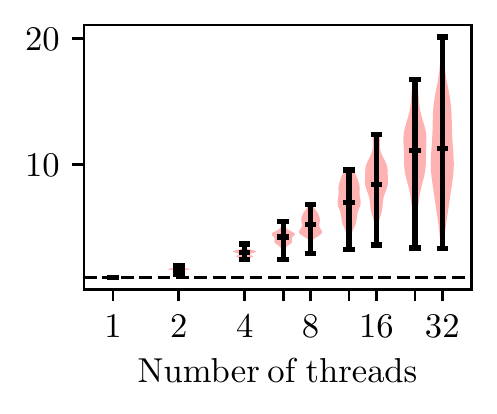}\\
        \multicolumn{1}{c}{(a) Liu-Sun} & \multicolumn{1}{c}{(b) P-PPR} \\
        \includegraphics[trim=0pt 6pt 0pt 6pt, clip, height=0.36\linewidth]{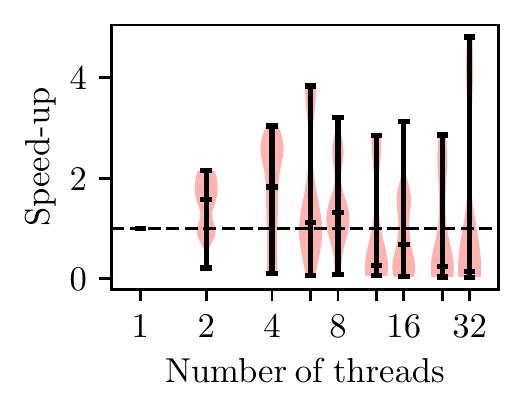} &
        \includegraphics[trim=0pt 6pt 0pt 6pt, clip, height=0.36\linewidth]{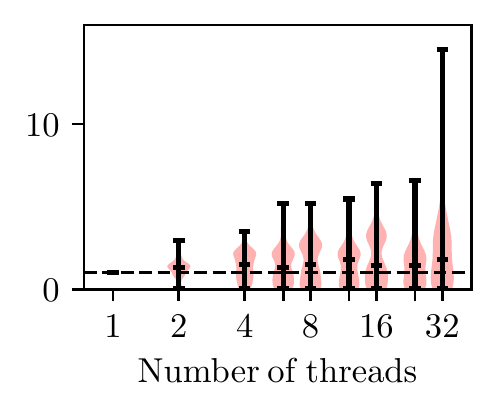} \\
        \multicolumn{1}{c}{(c) Strandmark-Kahl} & \multicolumn{1}{c}{(d) P-ARD} \\
    \end{tabular}
    \caption{Speed-up of the the parallel algorithms compared to their single-threaded performance. For each number of threads, the distribution of the speed-ups over all datasets is shown. The values were re-sampled as described in \cref{fig:serial_rel_perf}.}
    \label{fig:parallel-speedup}
\end{figure}

\begin{table*}

\centering
\caption{\textbf{Performance of parallel algorithms} based on build and solve times. We show a representative subset of the datasets grouped according to their problem family. See \cref{tab:serial-perf} for the number of nodes and arcs. The algorithms were run with 1, 2, 4, 6, 8, 12, 16, 24, and 32 threads. Only the best time is shown along with the thread count for that run. For comparison, the solve time for the fastest serial algorithm is also included. All times are in seconds. The fastest solve time for each dataset has been marked with \textbf{bold face}}
\label{tab:parallel-perf}
\setlength\tabcolsep{2.45pt} 
\def\arraystretch{1.05}

\begin{tabularx}{\linewidth}{Xrrrrrrrrrrrrrrrrrlr}
\toprule
 & 
\multicolumn{3}{c}{Liu-Sun \cite{Liu2010}} &
\multicolumn{3}{c}{P-PPR \cite{Baumstark2015}} &
\multicolumn{3}{c}{Strandmark-Kahl \cite{Strandmark2010}} &
\multicolumn{3}{c}{P-ARD \cite{Shekhovtsov2013}} &
\multicolumn{3}{c}{P-GridCut \cite{Jamriska2012, Gridcut2015}} &
\multicolumn{2}{c}{Best serial} \\
Dataset & Build & \multicolumn{2}{c}{Best solve} & Build & \multicolumn{2}{c}{Best solve} & Build & \multicolumn{2}{c}{Best solve} & Build & \multicolumn{2}{c}{Best solve} & Build & \multicolumn{2}{c}{Best solve} & Algo. & Solve \\
\midrule
\multicolumn{18}{l}{3D segmentation: voxel-based} \\
\midrule
 adhead.n26c10 \cite{Boykov2001, Boykov2003, Boykov2006} & 6.90 & 17.41 & 8T & 42.71 & 14.37 & 12T & 26.14 & 25.40 & 2T & 78.20 & 35.17 & 4T & - & - & - & GridCut & \textbf{13.21} \\
 adhead.n26c100 \cite{Boykov2001, Boykov2003, Boykov2006} & 6.86 & 20.87 & 8T & 41.79 & \textbf{8.41} & 32T & 27.65 & 19.30 & 6T & 75.37 & 42.91 & 4T & - & - & - & EIBFS & 22.60 \\
 babyface.n26c100 \cite{Boykov2001, Boykov2003, Boykov2006} & 2.89 & 72.26 & 32T & 15.99 & \textbf{7.93} & 32T & 9.15 & 40.86 & 4T & 32.01 & 61.20 & 32T & - & - & - & EIBFS & 30.13 \\
 bone.n26c100 \cite{Boykov2001, Boykov2003, Boykov2006} & 4.36 & 3.68 & 32T & 25.63 & 4.01 & 32T & 16.38 & 5.04 & 8T & 48.90 & 11.21 & 4T & - & - & - & HPF & \textbf{3.48} \\
 bone\_subx.n26c100 \cite{Boykov2001, Boykov2003, Boykov2006} & 2.30 & 4.04 & 16T & 12.48 & 2.34 & 32T & 8.03 & 4.43 & 8T & 24.07 & 11.71 & 16T & - & - & - & HPF & \textbf{2.14} \\
 liver.n26c10 \cite{Boykov2001, Boykov2003, Boykov2006} & 2.37 & 10.79 & 6T & 14.42 & 7.91 & 32T & 0.89 & 3.49 & 1T & 21.93 & 14.95 & 1T & - & - & - & GridCut & \textbf{2.95} \\
 liver.n26c100 \cite{Boykov2001, Boykov2003, Boykov2006} & 2.36 & 18.24 & 6T & 13.94 & \textbf{5.45} & 32T & 0.86 & 6.69 & 1T & 27.07 & 25.74 & 24T & - & - & - & GridCut & 5.62 \\
 liver.n6c100 \cite{Boykov2001, Boykov2003, Boykov2006} & 0.53 & 7.62 & 6T & 4.78 & 3.68 & 16T & 0.51 & 7.24 & 1T & 6.18 & 7.74 & 32T & 0.11 & \textbf{2.70} & 6T & GridCut & 3.87 \\
 adhead.n6c100 \cite{Boykov2001, Boykov2003, Boykov2006} & 1.59 & 11.17 & 8T & 14.49 & 4.72 & 32T & 2.52 & 7.17 & 4T & 15.46 & 14.41 & 2T & 0.31 & \textbf{3.83} & 4T & GridCut & 6.98 \\
 babyface.n6c10 \cite{Boykov2001, Boykov2003, Boykov2006} & 0.66 & 2.70 & 32T & 5.30 & 3.09 & 24T & 0.73 & 3.55 & 1T & 8.87 & 5.35 & 1T & 0.12 & \textbf{0.88} & 32T & GridCut & 1.52 \\
 babyface.n6c100 \cite{Boykov2001, Boykov2003, Boykov2006} & 0.66 & 5.34 & 32T & 6.53 & 3.63 & 24T & 0.98 & 5.24 & 4T & 7.29 & 7.28 & 32T & 0.12 & \textbf{1.66} & 16T & GridCut & 2.88 \\
 bone.n6c100 \cite{Boykov2001, Boykov2003, Boykov2006} & 0.99 & 0.79 & 24T & 8.30 & 2.66 & 32T & 1.23 & 2.01 & 2T & 11.18 & 2.08 & 4T & 0.20 & \textbf{0.17} & 12T & GridCut & 0.91 \\
\midrule
\multicolumn{18}{l}{3D segmentation: oriented MRF} \\
\midrule
 vessel.orimrf.256 \cite{Boykov2004} & 1.22 & 0.69 & 6T & 14.67 & 1.69 & 32T & 1.89 & 1.04 & 2T & 16.24 & 0.82 & 32T & 0.52 & \textbf{0.14} & 12T & GridCut & 0.40 \\
 vessel.orimrf.512 \cite{Boykov2004} & 9.72 & 4.92 & 8T & - & - & - & 15.08 & 6.49 & 2T & 100.95 & 5.54 & 16T & 4.26 & \textbf{0.43} & 32T & GridCut & 2.43 \\
 vessel.orimrf.900 \cite{Boykov2004} & 49.62 & 28.66 & 8T & - & - & - & \newtext{79.82} & \newtext{38.45} & \newtext{2T} & 599.09 & 24.02 & 32T & 21.98 & \textbf{2.49} & 32T & GridCut & 15.97 \\
\midrule
\multicolumn{18}{l}{3D U-Net segmentation cleaning} \\
\midrule
 clean.orimrf.256 \cite{Boykov2004} & 1.23 & 0.39 & 12T & 16.50 & 2.48 & 24T & 2.39 & 0.48 & 4T & 15.96 & 0.84 & 8T & 0.52 & \textbf{0.06} & 32T & GridCut & 0.13 \\
 clean.orimrf.512 \cite{Boykov2004} & 9.73 & 2.45 & 16T & - & - & - & 18.26 & 3.81 & 4T & 131.62 & 4.81 & 8T & 4.37 & \textbf{0.27} & 32T & GridCut & 0.91 \\
 clean.orimrf.900 \cite{Boykov2004} & 50.52 & 12.36 & 16T & - & - & - & \newtext{85.77} & \newtext{18.66} & \newtext{4T} & 578.09 & 20.17 & 16T & 23.64 & \textbf{0.82} & 32T & GridCut & 3.90 \\
 unet\_mrfclean\_3 \cite{Boykov2004} & 1.15 & 0.27 & 32T & - & - & - & 2.15 & 0.45 & 4T & 12.66 & 0.53 & 4T & 0.46 & \textbf{0.04} & 32T & GridCut & 0.11 \\
 unet\_mrfclean\_8 \cite{Boykov2004} & 0.37 & 0.21 & 8T & - & - & - & 0.64 & 0.23 & 4T & 4.01 & 0.28 & 2T & 0.14 & \textbf{0.04} & 16T & GridCut & 0.11 \\
\midrule
\multicolumn{18}{l}{Surface fitting} \\
\midrule
 LB07-bunny-lrg \cite{Lempitsky2007} & 6.14 & 1.86 & 16T & 55.88 & 24.27 & 32T & 7.73 & 4.14 & 4T & 72.02 & 4.31 & 16T & 1.24 & \textbf{0.32} & 24T & GridCut & 2.36 \\
\midrule
\multicolumn{18}{l}{3D segmentation: sparse layered graphs (SLG)} \\
\midrule
 4Dpipe\_small \cite{Jeppesen2020} & 9.93 & 9.39 & 12T & - & - & - & - & - & - & 47.53 & 421.60 & 24T & - & - & - & EIBFS & \textbf{2.06} \\
 4Dpipe\_big \cite{Jeppesen2020} & 122.43 & 86.18 & 16T & - & - & - & - & - & - & 570.44 & 7485.11 & 4T & - & - & - & EIBFS & \textbf{20.59} \\
 NT32\_tomo3\_.raw\_10 \cite{Jeppesen2020} & 4.99 & 18.58 & 12T & 34.90 & \textbf{14.39} & 32T & 22.02 & 85.49 & 1T & 43.95 & 15.40 & 12T & - & - & - & HPF & 36.46 \\
 NT32\_tomo3\_.raw\_30 \cite{Jeppesen2020} & 14.93 & 45.70 & 16T & 111.70 & 59.81 & 32T & 66.56 & 363.06 & 1T & 132.41 & \textbf{36.13} & 32T & - & - & - & BK & 145.23 \\
 NT32\_tomo3\_.raw\_100 \cite{Jeppesen2020} & 38.93 & \textbf{95.24} & 32T & - & - & 1T & 170.38 & 1189.78 & 1T & 365.60 & 158.92 & 24T & - & - & - & HPF & 498.94 \\
\midrule
\multicolumn{18}{l}{3D segmentation: seperating surfaces} \\
\midrule
 cells.sd3 \cite{Jensen2020} & 4.18 & 10.33 & 16T & 25.93 & \textbf{9.47} & 32T & 23.90 & 76.40 & 1T & 21.84 & 44.98 & 1T & - & - & - & HPF & 15.52 \\
 foam.subset.r160.h210 \cite{Jensen2020} & 5.91 & 8.59 & 32T & 37.08 & 3.61 & 32T & 52.11 & 17.24 & 1T & 33.72 & 7.32 & 1T & - & - & - & EIBFS & \textbf{3.21} \\
 simcells.sd3 \cite{Jensen2020} & 0.69 & 2.28 & 16T & 5.60 & 1.99 & 32T & 1.49 & 2.82 & 32T & 4.96 & \textbf{0.89} & 32T & - & - & - & EIBFS & 2.94 \\
\midrule
\multicolumn{18}{l}{Multi-view} \\
\midrule
 BL06-camel-lrg \cite{Boykov2006b} & 3.99 & 57.41 & 8T & - & - & - & 1.87 & 75.56 & 1T & 13.76 & 95.40 & 1T & - & - & - & HPF & \textbf{24.44} \\
 BL06-gargoyle-lrg \cite{Boykov2006b} & 3.70 & 29.28 & 16T & - & - & - & 1.70 & 190.07 & 1T & 12.20 & 102.31 & 2T & - & - & - & HPF & \textbf{26.51} \\
\midrule
\multicolumn{18}{l}{Mesh segmentation} \\
\midrule
 bunnybig.segment \cite{Liu2015} & 0.30 & \textbf{0.37} & 12T & 2.81 & 1.15 & 32T & 1.12 & 1.89 & 1T & 3.66 & 0.41 & 32T & - & - & - & EIBFS & 0.62 \\
 chairbig.segment \cite{Liu2015} & 0.60 & 0.64 & 24T & 6.10 & 1.76 & 32T & 2.62 & 3.29 & 1T & 6.89 & \textbf{0.57} & 32T & - & - & - & EIBFS & 1.02 \\
 handbig.segment \cite{Liu2015} & 0.02 & 0.11 & 8T & 0.25 & 0.25 & 16T & 0.04 & 0.16 & 1T & 0.40 & 0.11 & 32T & - & - & - & EIBFS & \textbf{0.07} \\
\bottomrule
\end{tabularx}

\end{table*}

Finally, \cref{fig:parallel-speedup} shows the speed-up distribution of the parallel algorithms compared to their single-threaded performance. Only P-PPR improves consistently as more threads are added. Liu-Sun and P-ARD only show consistent improvements when looking at the maximum speed-up, and for over half of the datasets they have issues scaling beyond 12 threads.

\section{Algorithm Selection}\label{sec:algo-selection}
As the previous section shows, the performance of the individual \maxflow algorithms depends on the problem to be solved, \ie, the structure of the graph. Choosing the wrong algorithm may significantly increase the run time. In this section, we investigate strategies for selecting a \maxflow algorithm that maximize the expected performance given different levels of knowledge about the graph. To quantify the expected performance of a strategy, we will use the relative performance (RP), which we compute as follows: 1. Use the strategy to select an algorithm for each dataset. 2. For each dataset, compute the relative performance of the selected algorithm. For serial algorithms, this is the total time of the selected algorithm divided by the total time of the fastest algorithm for that dataset. For parallel algorithms, we use the solve time. This score shows the expected performance of a given strategy compared to choosing the fastest algorithm.

\textbf{\newtext{Scenario} 1: No Graph Knowledge} \enskip  If one has no knowledge of the graph to be solved, the best strategy is to choose the overall best algorithm. \cref{tab:global-rel-perf-summary} shows summary statistics for the performance scores of each algorithm. To avoid bias in \cref{fig:serial_variations_comp}, we oversample scores from each problem family so that they all have the same number of samples.

\begin{table}
    \centering
    \setlength\tabcolsep{8pt} 
    \caption{\textbf{Summary of relative performance (RP) scores for each of the \maxflow algorithm variants.} The best score (higher is better) in each column has been marked with \textbf{bold face}. Results were oversampled as described in \cref{fig:serial_rel_perf}. We only include results where the algorithm ran to completion.}
    \label{tab:global-rel-perf-summary}
    \begin{tabularx}{\linewidth}{Xrrr}
        \toprule
        Serial algorithms & Mean RP $\pm$ Std. RP & Min RP & Max RP \\
        \midrule
        \texttt{EIBFS-I} & 0.59 $\pm$ 0.28 & \textbf{0.1309} & \textbf{1.00} \\
        \texttt{EIBFS-I-NR} & 0.56 $\pm$ 0.32 & 0.0535 & \textbf{1.00} \\
        \texttt{EIBFS} & 0.47 $\pm$ 0.23 & 0.1288 & 0.94 \\
        HI-PR & 0.16 $\pm$ 0.17 & 0.0046 & \textbf{1.00} \\
        \texttt{HPF-H-F} & 0.59 $\pm$ 0.33 & 0.0279 & \textbf{1.00} \\
        \texttt{HPF-H-L} & \textbf{0.64 $\pm$ 0.36} & 0.0393 & \textbf{1.00} \\
        \texttt{HPF-L-F} & 0.49 $\pm$ 0.29 & 0.0313 & \textbf{1.00} \\
        \texttt{HPF-L-L} & 0.53 $\pm$ 0.31 & 0.0312 & \textbf{1.00} \\
        \texttt{MBK-R} & 0.27 $\pm$ 0.20 & 0.0006 & \textbf{1.00} \\
        \texttt{BK} & 0.27 $\pm$ 0.24 & 0.0005 & \textbf{1.00} \\
        \texttt{MBK} & 0.28 $\pm$ 0.22 & 0.0005 & \textbf{1.00} \\
        GridCut$^*$ & \textbf{0.99 $\pm$ 0.03} & \textbf{0.6419} & \textbf{1.00} \\
        \midrule
        Parallel algorithms & & & \\
        \midrule
        Liu-Sun& \textbf{0.48 $\pm$ 0.30} & 0.0667 & \textbf{1.00} \\
        P-PPR & 0.46 $\pm$ 0.38 & 0.0133 & \textbf{1.00} \\
        Strandmark-Kahl & 0.23 $\pm$ 0.16 & 0.0667 & 0.85 \\
        P-ARD & 0.35 $\pm 0.32$ & 0.0028 & \textbf{1.00} \\
        P-GridCut$^{*}$ & \textbf{1.00 $\pm$ 0.00} & \textbf{1.0000} & \textbf{1.00} \\
        Best serial & \textbf{0.59 $\pm$ 0.33} & \textbf{0.1365} & \textbf{1.00} \\
        \bottomrule
        \multicolumn{4}{l}{$^*$ Only grid graphs included (6- and 26-conn. for serial, 6-conn for parallel).}
    \end{tabularx}
\end{table}

For the serial algorithms, the best choice is by far GridCut if it is applicable. It is almost always the fastest option and never more than 36\% slower than the best option. Otherwise, the best option is \texttt{HPF-H-L} in which case the expected performance 64\% of the optimal. Another good option is \texttt{EIBFS-I} due to its high mean and high minimum RP scores.
\newtext{All implementations, except \texttt{EIBFS}, have a maximum RP of 1, meaning that they outperformed all other implementations on at least one problem instance.}

For the parallel algorithms, GridCut again dominates when applicable. Otherwise, the best parallel option is Liu-Sun which is slightly better than P-PPR. \newtext{Surprisingly}, using the best serial algorithm for a dataset is the overall best option, \newtext{although we should note that comparing to the \textit{best} serial algorithm gives some advantage to the serial algorithms. If one compares to a single serial algorithm, the parallel algorithms do give an improvement --- although the mean RP is only 1.8x higher in the best case.}

\textbf{\newtext{Scenario} 2: Known Problem Family} \enskip If one knows from which problem family the graph to be solved comes, a good strategy is to select the algorithm that performs well on that problem family. This could, for example, be established beforehand by running a set of benchmarks on example graphs.

\cref{tab:serial-family-rel-perf-summary} shows the best performing serial algorithm for each problem family. Note that, as opposed to \cref{tab:serial-perf}, we split graph matching into sub-groups as papers use different energy functions for the matching. For all but four problem families, the best algorithm achieves a mean relative performance of 95\% or higher. Furthermore, for most problem families, one algorithm is always the best. This indicates that the problem family is a strong predictor of algorithm performance. The problem family where this strategy performs the worst is 3D segmentation with \newtext{sparse layered graphs (SLG)}. Here, the mean RP is only 81\%, which is likely due to the large variation in graph size in this problem family.
\begin{table}
    \centering
    \caption{\textbf{Relative performance (RP) scores for the best serial algorithm variant for each problem family.} Almost all problem families have one dominant algorithm.}
    \label{tab:serial-family-rel-perf-summary}
    \setlength\tabcolsep{4pt} 
    \begin{tabularx}{\linewidth}{Xlr}
        \toprule
        Problem family & Algorithm & Mean RP \\
        \midrule
        3D segmentation: SLG \cite{Jeppesen2020} & \texttt{HPF-H-L} & 0.81 \\
        Multi-view \cite{Boykov2006b} & \texttt{HPF-H-L} & 1.00 \\
        Surface fitting \cite{Lempitsky2007} & GridCut & 1.00 \\
        3D segmentation: voxel-based \cite{Boykov2001, Boykov2003, Boykov2006} & GridCut & 0.98 \\
        Mesh segmentation \cite{Liu2015} & \texttt{EIBFS-I} & 0.95 \\
        3D segmentation: sep. surfaces \cite{Jensen2020} & \texttt{EIBFS-I} & 0.92 \\
        3D MRF \cite{Boykov2004} & GridCut & 1.00 \\
        Deep LOGISMOS \cite{Guo2018} & \texttt{EIBFS-I-NR} & 0.96 \\
        Deconvolution \cite{Rother2007} & \texttt{HPF-H-L} & 0.96 \\
        DTF \cite{Nowozin2011} & \texttt{HPF-H-L} & 1.00 \\
        Super resolution \cite{Freeman2000, Rother2007} & \texttt{EIBFS-I} & 0.87 \\
        Stereo 1 \cite{Boykov1998} & \texttt{EIBFS-I} & 0.99 \\
        Stereo 2 \cite{Kolmogorov2001} & \texttt{EIBFS-I} & 1.00 \\
        ALE \cite{Ladicky2009, Ladicky2010, PascalVoc2010} & \texttt{EIBFS-I-NR} & 1.00 \\
        Graph matching: small \cite{Kainmueller2014, Hutschenreiter2021} & \texttt{HPF-L-L} & 1.00 \\
        Graph matching: small \cite{PascalVoc2007, Leordeanu2012, Hutschenreiter2021} & \texttt{EIBFS-I-NR} & 0.91 \\
        Graph matching: small \cite{Alhaija2015, Swoboda2017, Hutschenreiter2021} & \texttt{HPF-L-F} & 1.00 \\
        Graph matching: small \cite{Torresani2008, Caetano2009, Hutschenreiter2021} & \texttt{HPF-L-L} & 1.00 \\
        Graph matching: small \cite{Komodakis2008, Kappes2015, Hutschenreiter2021} & \texttt{HPF-L-F} & 1.00 \\
        Graph matching: big \cite{Hutschenreiter2021} & \texttt{HPF-H-L} & 1.00 \\
        \midrule
        Mean $\pm$ std. & & 0.97 $\pm$ 0.05 \\
        \bottomrule
    \end{tabularx}
\end{table}

\cref{tab:parallel-family-rel-perf-summary} shows the best performing parallel algorithm for each problem family. For the 6-connected graphs, the parallel GridCut algorithm is clearly superior, but otherwise, the different families appear to favor different algorithms.
\begin{table}
    \centering
    \caption{\textbf{Relative performance (RP) scores for the best parallel algorithm for each problem family.} Since the parallel GridCut implementation can only handle 6-connected graphs `3D segmentation: voxel based' has been split into two subgroups: 6-connected graphs and 26-connected graphs. If an algorithm did not run to completion on a dataset we count the RP as 0.}
    \label{tab:parallel-family-rel-perf-summary}
    \setlength\tabcolsep{4pt} 
    \begin{tabularx}{\linewidth}{Xlr}
        \toprule
        Problem family & Algorithm & Mean RP \\
        \midrule
        3D segmentation: SLG \cite{Jeppesen2020} & Liu-Sun & 0.63 \\
        Multi-view \cite{Boykov2006b} & Serial & 1.00 \\
        Surface fitting \cite{Lempitsky2007} & P-GridCut & 1.00 \\
        3D seg.: voxel-based \cite{Boykov2001, Boykov2003, Boykov2006} (26-conn.) & Serial & 0.86 \\
        3D seg.: voxel-based \cite{Boykov2001, Boykov2003, Boykov2006} (6-conn.) & P-GridCut & 1.00 \\
        Mesh segmentation \cite{Liu2015} & P-ARD & 0.88 \\
        3D segmentation: sep. surfaces \cite{Jensen2020} & P-PPR & 0.74 \\
        3D MRF \cite{Boykov2004} & P-GridCut & 1.00 \\
        \midrule
        Mean $\pm$ std. & & 0.89 $\pm$ 0.14 \\
        \bottomrule
    \end{tabularx}
\end{table}

\textbf{\newtext{Scenario} 3: Known Graph} \enskip Finally, we consider a strategy where the graph is known, but the problem family is not. Here, our strategy is to train a simple decision tree to predict the best algorithm given a feature vector that describes the graph to be solved. Although a single decision tree is not the strongest classifier, it has the benefit of being easily interpretable.

\begin{figure}
    \centering
    \includegraphics[width=\linewidth]{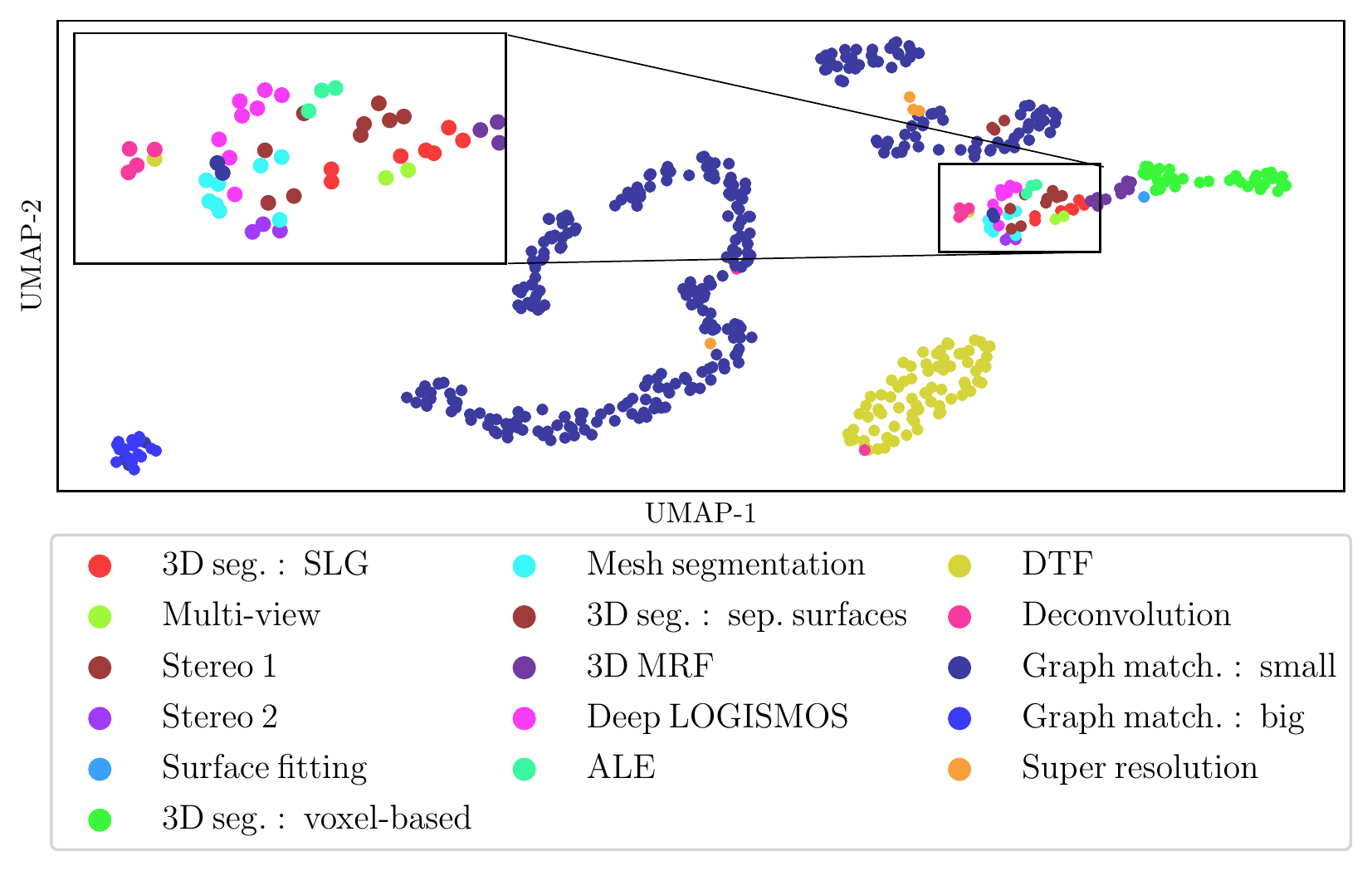}
    \caption{\textbf{UMAP embedding \cite{Mcinnes2018} of the extracted graph features.} Each point correspond to a benchmark dataset and is colored according to its problem family. When a benchmark consists of multiple sub-problem we use the mean feature vector. Notice that points from the same problem family tend to cluster together.}
    \label{fig:umap}
\end{figure}

The first components of our feature vector consist of the number of nodes, the number of terminal arcs, the number of neighbor arcs, and whether the graph is a grid graph.
Then we include mean, standard deviation, and standard deviation of non-zero values for a number of arc and node properties. For arc properties, we use: source, sink, terminal (source and sink combined), and neighbor capacities. Finally, for node properties we use: sum of in-going neighbor capacities, sum of out-going neighbor capacities, sum of neighbor capacities, degrees, out degrees, and in degrees counts \newtext{only non-zero arcs}. Note that these statistics can be computed efficiently during graph construction. \newtext{We normalize all capacity statistics by the mean over all arc capacities.}
In total, our feature vector has 31 entries per graph.
\cref{fig:umap} shows a UMAP embedding \cite{Mcinnes2018} of the feature vectors for all benchmark datasets. Similar problem families cluster together, despite UMAP receiving no information on this. \newtext{This} suggests the feature vectors provide a good description of the graphs.

We train the decision tree using Scikit-learn \cite{scikit-learn} \newtext{version 0.23.1}. We use Gini impurity as the split criterion and reduce the tree using minimal cost-complexity pruning \cite{Breiman2017}. The optimal amount of pruning is determined with 5-fold cross validation. We split each problem family evenly into the folds (if it contains at least 5 datasets). When fitting, each dataset is weighed by one over the number of datasets in its problem family. When evaluating, we oversample the validation data, so that each problem family has the same number of entries. This indicates how well the decision tree will perform with representative training data. We also perform an additional evaluation where we hold out one problem family, fit on the rest, and then evaluate on the held out family. This indicates how well the decision tree will perform for a problem family that it has not yet encountered. We use the mean RP as validation metric.

\begin{figure}[t!]
    \centering
    \includegraphics[width=0.925\linewidth]{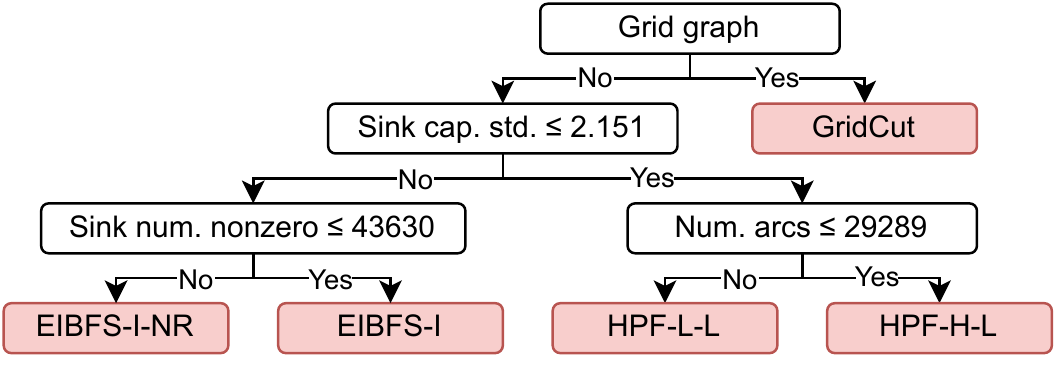}
    \caption{Decision tree trained to select the best serial algorithm. \newtext{Note that capacity statistics are normalized, \cf \cref{sec:algo-selection}.}}
    \label{fig:serial-tree}
\end{figure}

We first train a decision tree for the serial algorithms; the result is shown in \cref{fig:serial-tree}. It achieves a mean RP of \newtext{0.82 and 0.82} in the two evaluations, respectively. This means that the tree is significantly better than naively choosing the overall best algorithm but not as good as knowing the best algorithm for a problem family.

\begin{figure}
    \centering
    \includegraphics[width=0.925\linewidth]{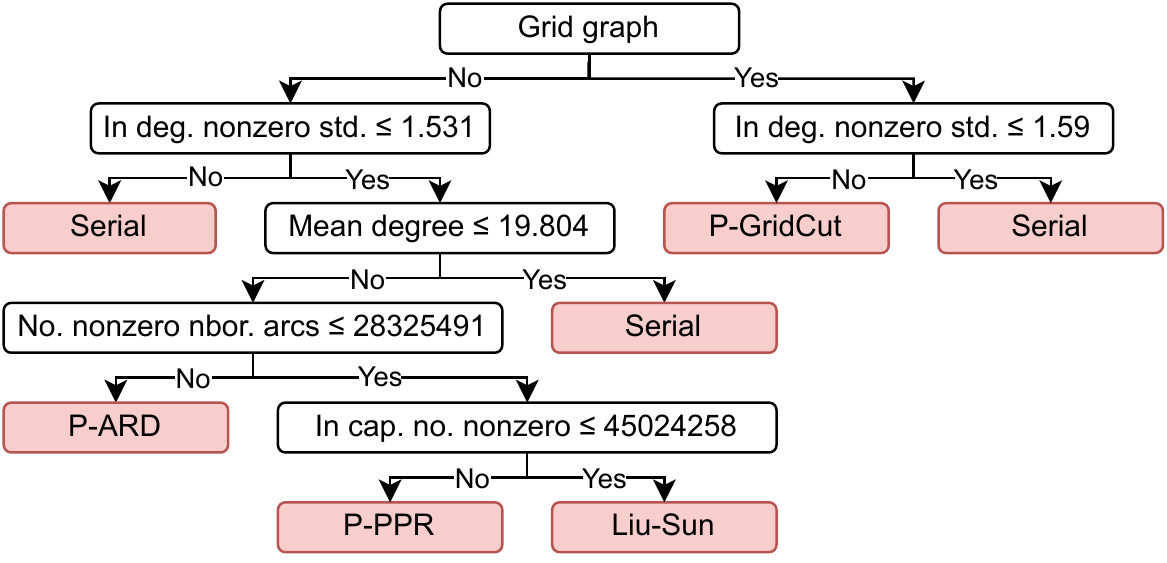}
    \caption{Decision tree trained to select the best parallel algorithm. `Serial' means a serial algorithm would be the best option. \newtext{Note that capacity statistics are normalized, \cf \cref{sec:algo-selection}.}}
    \label{fig:parallel-tree}
\end{figure}

Next, we train a decision \newtext{tree} for the parallel algorithms. We include a category `Serial', which means that choosing a serial algorithm would be faster. For simplicity, we do not specify which serial algorithm to choose in this scenario. The result is shown in \cref{fig:parallel-tree}. The decision tree achieves a mean RP of \newtext{0.56 and 0.57} in the two evaluations, respectively. Thus, the tree is slightly better than simply choosing the overall best algorithm.
However, the best option is to choose the best algorithm for a given category.


\section{Discussion}
\label{sec:discussion}
In this section, we discuss the most interesting findings from our experiments.

\subsection{Serial Algorithms}

Our results clearly show that GridCut is superior to the other tested algorithms for \maxflow problems with fixed neighborhood grids. This is not surprising since GridCut has been designed and optimized specifically for this type of graph. However, as shown in \cref{tab:serial-perf}, the performance benefit of GridCut decreases significantly when moving from 6-connected to 26-connected graphs. Actually, both EIBFS and HPF manage to beat GridCut on a couple of the 26-connected problems. This indicates that the benefit of using GridCut significantly decreases for graphs with high connectivity, perhaps because it is an AP algorithm.

In general, the pseudoflow algorithms have the best overall performance. Measured on solve time, EIBFS performs the best, which aligns with existing literature \cite{Goldberg2015}. However, looking at the total time, HPF performs better overall, slightly contradicting previous benchmarks \cite{Goldberg2015}. The reason for the difference between the results may be that \cite{Goldberg2015} compared EIBFS with \texttt{HPF-H-F}, which we show to be inferior to \texttt{HPF-H-L}. Also, in \cite{Goldberg2015} they use 32-bit pointer for most datasets, which may also provide slightly improved performance. Finally, the hardware used in \cite{Goldberg2015} may have different performance characteristics than ours. We observed that EIBFS actually performed better on an older system than on the one we used for our experiments. We speculate if this could be due to the lower cache and memory latency (estimated using Intel$^{\tiny{\textregistered}}$ Memory Latency Checker v3.9a) on the older system compared to the one used for our benchmarks. In any case, this raises the question whether HPF implemented with arc packing could outperform EIBFS on even more problems.

Our results also show that the performance of the different algorithm variants varies, and the choice of variant can significantly affect the run time. Optimizing for cache efficiency seems to be of particular importance, since optimizations such as arc packing and smaller data structures have large effects on the solve times for both BK and EIBFS.

As shown in \cref{sec:algo-selection}, for non-grid problems, the best algorithm most often comes down to a choice between EIBFS or HPF. From \cref{fig:serial-tree} it seems that HPF is faster when \newtext{the sink (or, more likely, terminal) arc capacities vary a lot. As expected, \texttt{EIBFS-I-NR} is preferred for small graphs, while the preferred HPF variant for small graphs appears to be \texttt{HPF-L-L}, which aligns with the results in \cref{tab:serial-family-rel-perf-summary}.}
However, the best strategy \newtext{is} to test several algorithms on a set of problems from the family at hand.

\subsection{Parallel Algorithms}

P-GridCut provides the best performance of the parallel algorithms for 6-connected grid graph problems and scales well with many threads. Of the other parallel algorithms, Liu-Sun is overall the best, closely followed by P-PPR, which aligns with previous results \cite{Liu2010} and expectations \cite{Shekhovtsov2013,Strandmark2010}. However, all the block-based algorithms only scale well for large graphs. For small to medium problems, they do not scale to many threads, but seem to peak at 8-12 threads, \cf \cref{fig:parallel-speedup}. This also means that choosing an optimal thread count may be difficult. Only P-PPR scaled consistently with up to 32 threads.
In addition, all parallel algorithms were often outperformed by a serial algorithm except on large graphs. In fact, as \cref{tab:global-rel-perf-summary} shows, selecting a good serial algorithm has better expected performance than selecting any of the parallel algorithms.

For practical use, only the Liu-Sun, P-PPR, and P-ARD algorithms seem to be relevant as is. However, the block-based algorithms have the additional challenge of dividing the graph into blocks --- the result of which significantly affects the run time of the algorithms. \newtext{This was also shown in \cite{Shekhovtsov2013}, where it was noticed that the multiview problems would scale better with more processors when partitioned on vertex numbers vs.\ the grid.} While the graphs tested in this work have a natural way to be split, this may not always be the case. Meanwhile, even though this problem is avoided with P-PPR, it does not perform as well as the block-based algorithms overall, as shown in \cref{fig:parallel_rel_perf_solve}.

Finally, while all parallel algorithms had datasets where they were best, selecting the best parallel algorithm is difficult (except for 6-connected grid graphs). \newtext{No algorithm} showed dominant performance --- neither globally nor per problem family. Furthermore, using the decision tree only gives a small improvement over selecting the best overall algorithm. \cref{fig:parallel-tree} indicates that
\newtext{for grid and low-degree graphs, a serial algorithm or GridCut performs best. Otherwise, the choice comes down to graph size, with P-ARD doing better for the smaller graphs, P-PPR being faster for the medium-sized ones, and Liu-Sun performing the best for the largest graphs.}
However, as \cref{tab:parallel-family-rel-perf-summary} shows, the best strategy is again to test on a number of graphs from the problem family at hand.

\section{Conclusions and Perspectives}
We now summarize our findings for the serial and parallel algorithms tested in this work. We also provide perspectives on possible future developments of \maxflow methods, as well as how these may fit into the future of computer vision.

\subsection{Serial Algorithms}
For the serial \maxflow algorithms, we have tested a total of 12 different variants across five of the fastest and most popular algorithms: PPR, BK, EIBFS, HPF, and GridCut. These include representatives for the three families of \maxflow algorithms: augmenting paths, push-relabel, and pseudoflow.

Our results clearly show that, for simple grid graphs, GridCut has the best performance. In most other cases, the two pseudoflow algorithms, EIBFS and HPF, are significantly faster than the other algorithms and thus should be the first choice for anyone looking for a fast serial \maxflow algorithm for static computer vision problems. For dynamic problems, we refer to \cite{Goldberg2015}.

Contrary to existing literature, we recommend the HPF algorithm in the H-LIFO configuration as the default, since it has the best overall performance. However, the EIBFS algorithm (\texttt{EIBFS-I} implementation) is a very close contender and can easily replace HPF with little impact on performance \newtext{--- and indeed may perform better on some problem families.}
If memory usage is of chief concern, the \newtext{\texttt{MBK} and \texttt{EIBFS-I-NR} implementations} are both good options, as they use significantly less memory than the reference EIBFS and HPF implementations.

Furthermore, we think significant performance improvements may be gained from further improving the algorithm implementations --- especially with a focus on memory use and cache efficiency. In particular,
faster and more memory efficient methods for arc (and node) packing
could result in significant benefits, since the extra initialization step incurs a large memory and run time overhead. We would like to see a reimplementation of HPF with a half-arc data structure and arc packing.

Finally, we found significant gains through automatic algorithm selection. Based on our results, it seems likely that one could train a robust classifier for selecting the appropriate algorithm based on the \maxflow problem to be solved. By selecting the right algorithm for the job, run time could in many cases be significantly reduced without the need for new algorithms or implementations. In general, we find it unlikely that a single algorithm will ever be dominant for all types of graphs.

\subsection{Parallel Algorithms}
We tested five different parallel algorithms for \maxflow problems: parallel PPR (P-PPR), adaptive bottom-up merging (Liu-Sun), dual decomposition (Strandmark-Kahl), region discharge (P-ARD), and parallel GridCut (P-GridCut).

If the graph is a simple grid, P-GridCut significantly outperforms all other algorithms. For other graphs, we found adaptive bottom-up merging, as proposed by Liu and Sun \cite{Liu2010}, to be the best overall parallel approach. However, each parallel algorithm had an area in which it was the best, and it is difficult to predict the best parallel algorithm for a graph (except for 6-connected grid graphs).

Of the parallel algorithms, only P-GridCut and P-PPR improved consistently with more threads. All block-based algorithms failed to scale beyond 12 threads, except on large graphs. Furthermore, except for P-GridCut, all parallel algorithms were often outperformed by a serial algorithm, and consistent improvements over serial algorithms were obtained only for large graphs. These issues reveal a major deficiency in the state of current parallel \maxflow algorithms and deserve further study. \newtext{While providing good scaling on any type of graph may be unreachable as \maxflow is P-complete and therefore hard to parallelize \cite{Greenlaw1995}, computer vision graphs often come with additional structure. Therefore, it seems highly likely that further improvements in practical performance can be achieved. However, at this time}, we only recommend using a parallel algorithm for graphs with more than 5 M nodes or where a serial algorithm uses at least 5 seconds.

To improve the parallel \maxflow algorithms, one could try to replace BK, which is currently used in all the tested block-based parallel algorithms, with a pseudoflow algorithm. However, this may not be trivial. In \cite{Jeppesen2021}, results for a Liu-Sun implementation using EIBFS instead of BK showed a significant performance decrease compared to serial EIBFS. Still, given the superior performance of pseudoflow algorithms, this is an important area to investigate. Furthermore, parallelized graph construction is currently only available for P-GridCut. As the build time is a significant part of the total time, reducing build time will be important --- especially as solve time decreases.

Finally, choosing an optimal blocking strategy remains an open problem. Generally, when nodes correspond to spatial positions (\eg, pixels or mesh vertices), we find that grouping based on spatial distance works well. However, \newtext{we recommend that practitioners experiment with different blocking strategies since it can significantly affect the performance. Furthermore,} a general method that only considers the graph structure would be of \newtext{high} interest, as this would also make the algorithms more accessible to the average user. An alternative would be to focus on P-PPR algorithms that do not rely on blocking. Further improvements in these areas could also open the door to GPU-based implementations for solving general \maxflow problems.

\subsection{Min-Cut/Max-Flow in Modern Computer Vision}
It is no secret that the field of computer vision is currently dominated by deep learning. In this context, it is highly relevant to consider the future role of traditional computer vision tools, such as \maxflow algorithms.

For 3D images used in medical imaging and materials science research \cite{Maire2014}, it is common to have images where no relevant training data are available. Here, segmentation methods based on \maxflow continue to play an important role, as they work without training data and allow geometric prior knowledge to be incorporated. Furthermore, while modern 3D images can already be very large (many GB per image), dynamic imaging (3D + time) with high acquisition rates is now also possible \cite{Garcia2019, Mokso2017}. Computational efficiency is paramount to be able to process this ever increasing amount of data, and for this, parallel \maxflow algorithms could prove particularly useful.

Finally, as mentioned in \cite{Minaee2021}, there is agreement that the performance of deep learning-based segmentation methods has started to plateau, and investigating how to integrate CNNs with `classical' approaches should be pursued. Already, combinations with active contours have shown promising results \cite{Guo2021,Peng2020,Wickramasinghe2020} and a combination of CNNs and \maxflow methods could lead to new advances. \newtext{As} deep learning involves repeated forward and backward passes through a model, it is crucial that the \maxflow algorithms are fast and efficient. \newtext{While not the focus of this work, this is also an area where dynamic \maxflow algorithms can be of great importance, as they are effective at handling repeated solves of graphs where capacities do not change drastically between successive solves.}

{
\printbibliography

@String { BMVC         = {{British Machine Vision Conference}} }

@String { CVPR         = {{{IEEE} Conference on Computer Vision and Pattern Recognition}} }

@String { CVPRW        = {{{IEEE} Conference on Computer Vision and Pattern Recognition Workshops}} }

@String { ECCV         = {{European Conference on Computer Vision}} }

@String { ESA          = {{European Symposium on Algorithms}} }

@String { ICALP        = {{International Colloquium on Automata, Languages, and Programming}} }

@String { ICCV         = {{International Conference on Computer Vision}} }

@String { ICDCS        = {{International Conference on Distributed Computing Systems}} }

@String { IJCV         = {{International Journal of Computer Vision}} }

@String  { IPL         = {{Information Processing Letters}} }

@String { ISBI         = {{{IEEE} International Symposium on Biomedical Imaging}} }

@String { ISVC         = {{International Symposium on Visual Computing}}}

@String { JACM         = {{Journal of the ACM}} }

@String { JMLR         = {{Journal of Machine Learning and Research}} }

@String { JPDC         = {{Journal of Parallel and Distributed Computing}} }

@String { JPRS         = {{Joint Pattern Recognition Symposium}} }

@String { JRTIP        = {{Journal of Real-Time Image Processing}} }

@String { MIA          = {{Medical Image Analysis}} }

@String { MICCAI       = {{International Conference on Medical Image Computing and Computer Assisted Intervention}} }

@String { MP           = {{Mathematical Programming}} }

@String { NETWORKS     = {{Networks}} }

@String { OR           = {{Operations Research}} }

@String { PAMI         = {{{IEEE} Transactions on Pattern Analysis and Machine Intelligence}} }

@String { PR           = {{Pattern Recognition}} }

@String { RBME         = {{{IEEE} Reviews in Biomedical Engineering ({RBME})}} }

@String { SP           = {{Signal Processing}} }

@String { THREEDV      = {{International Conference on 3D Vision}} }

@String { TIP          = {{{IEEE} Transactions on Image Processing}} }

@String { TMI          = {{{IEEE} Transactions on Medical Imaging}} }

@String { TPDS         = {{{IEEE} Transactions on Parallel and Distributed Systems}} }

@book{Greenlaw1995,
    author={Greenlaw, Raymond and H. James Hoover and Walter L. Ruzzo},
    title={Limits to parallel computation: P-completeness theory},
    publisher={Oxford University Press on Demand},
    year={1995}
}

@ARTICLE{Boykov2001b,
  author={Boykov, Y. and Veksler, O. and Zabih, R.},
  journal={IEEE Transactions on Pattern Analysis and Machine Intelligence},
  title={Fast approximate energy minimization via graph cuts},
  year={2001},
  volume={23},
  number={11},
  pages={1222-1239}
}

@misc{Code,
    title = {Max-Flow/Min-Cut Algorithms},
    author = {Jensen, Patrick M. and Jeppesen, Niels},
    howpublished = {\url{https://doi.org/10.5281/zenodo.4903945}},
    doi = {10.5281/zenodo.4903945},
    note = {Accessed 2021-06-08}
}

@misc{Data,
    title = {Min-Cut/Max-Flow Problem Instances for Benchmarking},
    author = {Jensen, Patrick M. and Jeppesen, Niels and Dahl, Anders B. and Dahl, Vedrana A.},
    howpublished = {\url{https://doi.org/10.11583/DTU.17091101}},
    doi = {10.11583/DTU.17091101},
    note = {Accessed 2021-11-29}
}

@misc{DTU_DCC_resource,
    author    = {{DTU Computing Center}},
    title     = {{DTU Computing Center resources}},
    year      = {2021},
    publisher = {Technical University of Denmark},
    doi       = {10.48714/DTU.HPC.0001},
    url       = {https://doi.org/10.48714/DTU.HPC.0001},
}

@article{scikit-learn,
 title={Scikit-learn: Machine Learning in {P}ython},
 author={Pedregosa, F. and Varoquaux, G. and Gramfort, A. and Michel, V.
         and Thirion, B. and Grisel, O. and Blondel, M. and Prettenhofer, P.
         and Weiss, R. and Dubourg, V. and Vanderplas, J. and Passos, A. and
         Cournapeau, D. and Brucher, M. and Perrot, M. and Duchesnay, E.},
 journal=JMLR,
 volume={12},
 pages={2825--2830},
 year={2011}
}

@inproceedings{Halim2011,
  title={A MapReduce-based maximum-flow algorithm for large small-world network graphs},
  author={Halim, Felix and Yap, Roland HC and Wu, Yongzheng},
  booktitle=ICDCS,
  pages={192--202},
  year={2011}
}

@book{Breiman2017,
  title={Classification and regression trees},
  author={Breiman, Leo and Friedman, Jerome H and Olshen, Richard A and Stone, Charles J},
  year={2017},
  publisher={Routledge}
}

@article{Mcinnes2018,
  title={Umap: Uniform manifold approximation and projection for dimension reduction},
  author={McInnes, Leland and Healy, John and Melville, James},
  journal={arXiv:1802.03426},
  year={2018}
}

@InProceedings{Jeppesen2021,
    author    = {Jeppesen, Niels and Jensen, Patrick M. and Christensen, Anders N. and Dahl, Anders B. and Dahl, Vedrana A.},
    title     = {Faster Multi-Object Segmentation Using Parallel Quadratic Pseudo-Boolean Optimization},
    booktitle = ICCV,
    month     = {October},
    year      = {2021},
    pages     = {6260-6269}
}

@article{Grau2006,
  title={Segmentation of trabeculated structures using an anisotropic Markov random field: application to the study of the optic nerve head in glaucoma},
  author={Grau, Vicente and Downs, J Crawford and Burgoyne, Claude F},
  journal= TMI,
  volume={25},
  pages={245--255},
  year={2006}
}

@article{Simpson2019,
  title={A large annotated medical image dataset for the development and evaluation of segmentation algorithms},
  author={Simpson, Amber L and Antonelli, Michela and Bakas, Spyridon and Bilello, Michel and Farahani, Keyvan and Van Ginneken, Bram and Kopp-Schneider, Annette and Landman, Bennett A and Litjens, Geert and Menze, Bjoern and others},
  journal={arXiv:1902.09063},
  year={2019}
}

@inproceedings{Milletari2016,
  title={V-net: Fully convolutional neural networks for volumetric medical image segmentation},
  author={Milletari, Fausto and Navab, Nassir and Ahmadi, Seyed-Ahmad},
  booktitle=THREEDV,
  pages={565--571},
  year={2016}
}

@article{Reichardt2021,
  title={3D virtual Histopathology of Cardiac Tissue from Covid-19 Patients based on Phase-Contrast X-ray Tomography},
  author={Reichardt, Marius and Jensen, Patrick M. and Dahl, Vedrana Andersen and Dahl, Anders Bjorholm and Ackermann, Maximilian and Shah, Harshit and L{\"a}nger, Florian and Werlein, Christopher and Kuehnel, Mark P and Jonigk, Danny and others},
  journal={eLife.},
  year={2021}
}

@article{Greig1989,
 author = {D. M. Greig and B. T. Porteous and A. H. Seheult},
 journal = {Journal of the Royal Statistical Society. Series B (Methodological)},
 number = {2},
 pages = {271--279},
 title = {Exact Maximum A Posteriori Estimation for Binary Images},
 volume = {51},
 year = {1989}
}

@inproceedings{Boykov2001,
  title={Interactive graph cuts for optimal boundary \& region segmentation of objects in ND images},
  author={Boykov, Yuri Y and Jolly, M-P},
  booktitle=ICCV,
  volume={1},
  pages={105--112},
  year={2001}
}

@inproceedings{Boykov2003,
  title={Computing geodesics and minimal surfaces via graph cuts.},
  author={Boykov, Yuri and Kolmogorov, Vladimir},
  booktitle=ICCV,
  volume={3},
  pages={26--33},
  year={2003}
}

@article{Boykov2006,
  title={Graph cuts and efficient ND image segmentation},
  author={Boykov, Yuri and Funka-Lea, Gareth},
  journal=IJCV,
  volume={70},
  pages={109--131},
  year={2006}
}

@inproceedings{Boykov2006b,
  title={From Photohulls to Photoflux Optimization.},
  author={Boykov, Yuri and Lempitsky, Victor S},
  booktitle=BMVC,
  volume={3},
  pages={27},
  year={2006}
}

@inproceedings{Lempitsky2006,
  title={Oriented visibility for multiview reconstruction},
  author={Lempitsky, Victor and Boykov, Yuri and Ivanov, Denis},
  booktitle=ECCV,
  pages={226--238},
  year={2006}
}

@inproceedings{Hutschenreiter2021,
    author    = {Hutschenreiter, Lisa and Haller, Stefan and Feineis, Lorenz and Rother, Carsten and Kainm\"uller, Dagmar and Savchynskyy, Bogdan},
    title     = {Fusion Moves for Graph Matching},
    booktitle = ICCV,
    year      = {2021},
    pages     = {6270-6279}
}

@inproceedings{Torresani2008,
  title={Feature correspondence via graph matching: Models and global optimization},
  author={Torresani, Lorenzo and Kolmogorov, Vladimir and Rother, Carsten},
  booktitle= ECCV,
  pages={596--609},
  year={2008}
}

@article{Caetano2009,
  title={Learning graph matching},
  author={Caetano, Tib{\'e}rio S and McAuley, Julian J and Cheng, Li and Le, Quoc V and Smola, Alex J},
  journal=PAMI,
  volume={31},
  pages={1048--1058},
  year={2009}
}

@misc{PascalVoc2007,
	author = "Everingham, M. and Van~Gool, L. and Williams, C. K. I. and Winn, J. and Zisserman, A.",
	title = "The {PASCAL} {V}isual {O}bject {C}lasses {C}hallenge 2007 {(VOC2007)} {R}esults",
	howpublished = {\url{http://www.pascal-network.org/challenges/VOC/voc2007/workshop/index.html}}
}

@article{Leordeanu2012,
  title={Unsupervised learning for graph matching},
  author={Leordeanu, Marius and Sukthankar, Rahul and Hebert, Martial},
  journal= IJCV,
  volume={96},
  pages={28--45},
  year={2012},
}

@inproceedings{Cciccek2016,
  title={3D U-Net: learning dense volumetric segmentation from sparse annotation},
  author={{\c{C}}i{\c{c}}ek, {\"O}zg{\"u}n and Abdulkadir, Ahmed and Lienkamp, Soeren S and Brox, Thomas and Ronneberger, Olaf},
  booktitle= MICCAI,
  pages={424--432},
  year={2016}
}

@article{Guo2021,
  title={Learned snakes for 3D image segmentation},
  author={Guo, Lihong and Liu, Yueyun and Wang, Yu and Duan, Yuping and Tai, Xue-Cheng},
  journal= SP,
  volume={183},
  pages={108013},
  year={2021}
}

@inproceedings{Alhaija2015,
  title={Graphflow--6D large displacement scene flow via graph matching},
  author={Alhaija, Hassan Abu and Sellent, Anita and Kondermann, Daniel and Rother, Carsten},
  booktitle={German Conference on Pattern Recognition},
  pages={285--296},
  year={2015},
  organization={Springer}
}

@inproceedings{Swoboda2017,
  title={A study of lagrangean decompositions and dual ascent solvers for graph matching},
  author={Swoboda, Paul and Rother, Carsten and Abu Alhaija, Hassan and Kainmuller, Dagmar and Savchynskyy, Bogdan},
  booktitle=CVPR,
  pages={1607--1616},
  year={2017}
}

@inproceedings{Komodakis2008,
  title={Beyond loose LP-relaxations: Optimizing MRFs by repairing cycles},
  author={Komodakis, Nikos and Paragios, Nikos},
  booktitle= ECCV,
  pages={806--820},
  year={2008},
}

@article{Kappes2015,
  title={A comparative study of modern inference techniques for structured discrete energy minimization problems},
  author={Kappes, J{\"o}rg H and Andres, Bjoern and Hamprecht, Fred A and Schn{\"o}rr, Christoph and Nowozin, Sebastian and Batra, Dhruv and Kim, Sungwoong and Kausler, Bernhard X and Kr{\"o}ger, Thorben and Lellmann, Jan and others},
  journal= IJCV,
  volume={115},
  pages={155--184},
  year={2015},
}

@inproceedings{Kainmueller2014,
  title={Active graph matching for automatic joint segmentation and annotation of C. elegans},
  author={Kainmueller, Dagmar and Jug, Florian and Rother, Carsten and Myers, Gene},
  booktitle=MICCAI,
  pages={81--88},
  year={2014},
}

@inproceedings{Liu2015,
  title={Graph cut based mesh segmentation using feature points and geodesic distance},
  author={Liu, Lei and Sheng, Yun and Zhang, Guixu and Ugail, Hassan},
  booktitle={Proceedings of the International Conference on Cyberworlds ({CW})},
  pages={115--120},
  year={2015},
}

@misc{PascalVoc2010,
	author = "Everingham, M. and Van~Gool, L. and Williams, C. K. I. and Winn, J. and Zisserman, A.",
	title = "The {PASCAL} {V}isual {O}bject {C}lasses {C}hallenge 2010 {(VOC2010)} {R}esults",
	howpublished = {\url{http://www.pascal-network.org/challenges/VOC/voc2010/workshop/index.html}}
}

@inproceedings{Ladicky2009,
  title={Associative hierarchical crfs for object class image segmentation},
  author={Ladický, L'ubor and Russell, Chris and Kohli, Pushmeet and Torr, Philip HS},
  booktitle=ICCV,
  pages={739--746},
  year={2009}
}

@misc{ALE,
    author = {Ladický, L'ubor and Torr, Philip HS},
    title = {{The automatic labelling environment}},
    howpublished = {\url{https://www.robots.ox.ac.uk/~phst/ale.htm}},
    note = {Accessed 2021-11-24}
}

@inproceedings{Ladicky2010,
  title={Graph cut based inference with co-occurrence statistics},
  author={Ladický, L'ubor and Russell, Chris and Kohli, Pushmeet and Torr, Philip HS},
  booktitle=ECCV,
  pages={239--253},
  year={2010}
}

@article{Freeman2000,
  title={Learning low-level vision},
  author={Freeman, William T and Pasztor, Egon C and Carmichael, Owen T},
  journal=IJCV,
  volume={40},
  pages={25--47},
  year={2000}
}

@inproceedings{Nowozin2011,
  title={Decision tree fields},
  author={Nowozin, Sebastian and Rother, Carsten and Bagon, Shai and Sharp, Toby and Yao, Bangpeng and Kohli, Pushmeet},
  booktitle=ICCV,
  pages={1668--1675},
  year={2011}
}

@inproceedings{Wickramasinghe2020,
  title={Voxel2mesh: 3d mesh model generation from volumetric data},
  author={Wickramasinghe, Udaranga and Remelli, Edoardo and Knott, Graham and Fua, Pascal},
  booktitle= MICCAI,
  pages={299--308},
  year={2020},
  organization={Springer}
}

@inproceedings{Peng2020,
  title={Deep snake for real-time instance segmentation},
  author={Peng, Sida and Jiang, Wen and Pi, Huaijin and Li, Xiuli and Bao, Hujun and Zhou, Xiaowei},
  booktitle= CVPR,
  pages={8533--8542},
  year={2020}
}

@article{Mokso2017,
  title={GigaFRoST: the gigabit fast readout system for tomography},
  author={Mokso, Rajmund and Schlep{\"u}tz, Christian M and Theidel, Gerd and Billich, Heiner and Schmid, Elmar and Celcer, Tine and Mikuljan, Gordan and Sala, Leonardo and Marone, Federica and Schlumpf, Nick and others},
  journal={Journal of synchrotron radiation},
  volume={24},
  number={6},
  pages={1250--1259},
  year={2017}
}

@article{Garcia2019,
  title={Using X-ray tomoscopy to explore the dynamics of foaming metal},
  author={García-Moreno, Francisco and Kamm, Paul Hans and Neu, Tillmann Robert and B{\"u}lk, Felix and Mokso, Rajmund and Schlep{\"u}tz, Christian Matthias and Stampanoni, Marco and Banhart, John},
  journal={Nature communications},
  volume={10},
  number={1},
  pages={1--9},
  year={2019}
}

@article{Maire2014,
  title={Quantitative X-ray tomography},
  author={Maire, Eric and Withers, Philip John},
  journal={International materials reviews},
  volume={59},
  number={1},
  pages={1--43},
  year={2014}
}

@article{Minaee2021,
  title={Image segmentation using deep learning: A survey},
  author={Minaee, Shervin and Boykov, Yuri Y and Porikli, Fatih and Plaza, Antonio J and Kehtarnavaz, Nasser and Terzopoulos, Demetri},
  journal= PAMI,
  year = {2021}
}

@article{Tajbakhsh2020,
  title = {Embracing imperfect datasets: A review of deep learning solutions for medical image segmentation},
  author = {Tajbakhsh, Nima and Jeyaseelan, Laura and Li, Qian and Chiang, Jeffrey N and Wu, Zhihao and Ding, Xiaowei},
  journal = MIA,
  volume = {63},
  pages = {101693},
  year = {2020}
}

@inproceedings{Lempitsky2007,
  title={Global optimization for shape fitting},
  author={Lempitsky, Victor and Boykov, Yuri},
  booktitle=CVPR,
  pages={1--8},
  year={2007}
}

@inproceedings{Boykov1998,
  title={Markov random fields with efficient approximations},
  author={Boykov, Yuri and Veksler, Olga and Zabih, Ramin},
  booktitle=CVPR,
  pages={648--655},
  year={1998}
}

@inproceedings{Kolmogorov2001,
  title={Computing visual correspondence with occlusions using graph cuts},
  author={Kolmogorov, Vladimir and Zabih, Ramin},
  booktitle=ICCV,
  volume={2},
  pages={508--515},
  year={2001}
}

@techreport{Boros1991,
  title = {Network flows and minimization of quadratic pseudo-Boolean functions},
  author = {Boros, Endre and Hammer, Peter L and Sun, Xiaorong},
  year = {1991},
  institution = {17-1991, RUTCOR}
}

@misc{Waterloo,
    author = {University of Waterloo},
    title = {Max-flow problem instances in vision},
    howpublished = {\url{https://vision.cs.uwaterloo.ca/data/maxflow}},
    note = {Accessed 2021-02-05}
}

@inproceedings{Wang2010,
  author = {Wang, Yao and Beichel, Reinhard},
  title = {Graph-based segmentation of lymph nodes in CT data},
  booktitle = ISVC,
  pages = {312--321},
  year = {2010}
}

@inproceedings{Egger2010,
  author = {Egger, Jan and Bauer, Miriam HA and Kuhnt, Daniela and Carl, Barbara and Kappus, Christoph and Freisleben, Bernd and Nimsky, Christopher},
  title = {Nugget-cut: a segmentation scheme for spherically-and elliptically-shaped 3D objects},
  booktitle = JPRS,
  pages = {373--382},
  year = {2010}
}

@article{Hammer1984,
  author = {Hammer, Peter L and Hansen, Pierre and Simeone, Bruno},
  title = {Roof duality, complementation and persistency in quadratic 0--1 optimization},
  journal = MP,
  volume = {28},
  number = {2},
  pages = {121--155},
  year = {1984}
}

@article{Kolmogorov2004,
  title={What energy functions can be minimized via graph cuts?},
  author={Kolmogorov, Vladimir and Zabin, Ramin},
  journal=PAMI,
  volume={26},
  number={2},
  pages={147--159},
  year={2004}
}

@inproceedings{Freedman2005,
  title={Energy minimization via graph cuts: Settling what is possible},
  author={Freedman, Daniel and Drineas, Petros},
  booktitle= CVPR,
  pages={939--946},
  year={2005}
}

@inproceedings{Khoreva2017,
  author = {Khoreva, Anna and Benenson, Rodrigo and Hosang, Jan and Hein, Matthias and Schiele, Bernt},
  title = {Simple does it: Weakly supervised instance and semantic segmentation},
  booktitle = CVPR,
  pages = {876--885},
  year = {2017}
}

@article{Yu2017,
    author = {Yu, Miao and Shen, Shuhan and Hu, Zhanyi},
    journal = TIP,
    number = {8},
    publisher = {Institute of Electrical and Electronics Engineers Inc.},
    title = {{Dynamic Graph Cuts in Parallel}},
    volume = {26},
    year = {2017}
}

@article{Shekhovtsov2013,
    author = {Shekhovtsov, Alexander and Hlaváč, Václav},
    journal = IJCV,
    number = {3},
    pages = {315--342},
    title = {{A distributed mincut/maxflow algorithm combining path augmentation and push-relabel}},
    volume = {104},
    year = {2013}
}

@article{Hochbaum2008,
    author = {Hochbaum, Dorit S.},
    journal = OR,
    number = {4},
    pages = {992--1009},
    title = {{The pseudoflow algorithm: A new algorithm for the maximum-flow problem}},
    volume = {56},
    year = {2008}
}

@article{Hochbaum2013,
    author = {Hochbaum, Dorit S. and Orlin, James B.},
    journal = NETWORKS,
    number = {1},
    pages = {40--57},
    title = {{Simplifications and speedups of the pseudoflow algorithm}},
    volume = {61},
    year = {2013}
}

@article{Fishbain2016,
    author = {Fishbain, B. and Hochbaum, Dorit S. and Mueller, Stefan},
    journal = JRTIP,
    number = {3},
    pages = {589--609},
    publisher = {Springer Verlag},
    title = {{A competitive study of the pseudoflow algorithm for the minimum s–t cut problem in vision     applications}},
    volume = {11},
    year = {2016}
}

@article{Peng2015,
    author = {Peng, Yi and Chen, Li and Ou-Yang, Fang Xin and Chen, Wei and Yong, Jun Hai},
    journal = TIP,
    number = {2},
    pages = {655--666},
    title = {{JF-Cut: A parallel graph cut approach for large-scale image and video}},
    volume = {24},
    year = {2015}
}

@article{Boykov2004,
    author = {Boykov, Yuri and Kolmogorov, Vladimir},
    journal = PAMI,
    number = {9},
    pages = {1124--1137},
    title = {{An Experimental Comparison of Min-Cut/Max-Flow Algorithms for Energy Minimization in Vision}},
    volume = {26},
    year = {2004}
}

@inproceedings{Verma2012,
    author = {Verma, Tanmay and Batra, Dhruv},
    title = {{MaxFlow Revisited: An Empirical Comparison of Maxflow Algorithms for Dense Vision Problems}},
    booktitle = BMVC,
    pages = {1--12},
    year = {2012}
}

@inproceedings{Strandmark2010,
    author = {Strandmark, Petter and Kahl, Fredrik},
    title = {{Parallel and Distributed Graph Cuts by Dual Decomposition}},
    booktitle = CVPR,
    pages = {2085--2092},
    year = {2010}
}

@article{Kohli2007,
    author = {Kohli, Pushmeet and Torr, Philip H.S.},
    journal = PAMI,
    number = {12},
    pages = {2079--2088},
    title = {{Dynamic graph cuts for efficient inference in Markov random fields}},
    volume = {29},
    year = {2007}
}

@article{Yu2015,
    author = {Yu, Miao and Shen, Shuhan and Hu, Zhanyi},
    journal = TIP,
    number = {12},
    pages = {5511--5525},
    title = {{Dynamic Parallel and Distributed Graph Cuts}},
    volume = {25},
    year = {2015}
}

@inproceedings{Liu2010,
    author = {Liu, Jiangyu and Sun, Jian},
    title = {{Parallel Graph-cuts by Adaptive Bottom-up Merging}},
    booktitle = CVPR,
    pages = {2181--2188},
    year = {2010}
}

@inproceedings{Goldberg2015,
    author = {Goldberg, Andrew V and Hed, Sagi and Kaplan, Haim and Kohli, Pushmeet and Tarjan, Robert E and Werneck, Renato F},
    title = {{Faster and More Dynamic Maximum Flow by Incremental Breadth-First Search}},
    booktitle = ESA,
    pages = {619--630},
    year = {2015}
}

@inproceedings{Goldberg2011,
    author = {Goldberg, Andrew V and Hed, Sagi and Kaplan, Haim and Tarjan, Robert E and Werneck, Renato F},
    title = {{Maximum Flows by Incremental Breadth-First Search}},
    booktitle = ESA,
    pages = {457--468},
    year = {2011}
}

@inproceedings{Jamriska2012,
    author = {Jamriška, Ondřej and Sýkora, Daniel and Hornung, Alexander},
    title = {{Cache-efficient Graph Cuts on Structured Grids}},
    booktitle = CVPR,
    pages = {3673--3680},
    year = {2012}
}

@misc{Gridcut2015,
    author = {Jamriška, Ondřej and Sýkora, Daniel},
    title = {{GridCut. Version 1.3}},
    howpublished = {\url{https://gridcut.com}},
    note = {Accessed 2020-06-12},
    year = {2015},
}

@inproceedings{Vineet2008,
    author = {Vineet, Vibhav and Narayanan, P J},
    title = {{CUDA cuts: Fast graph cuts on the GPU}},
    booktitle = CVPRW,
    pages = {1--8},
    year = {2008},
}

@inproceedings{Delong2008,
    author = {Delong, Andrew and Boykov, Yuri},
    title = {A scalable graph-cut algorithm for ND grids},
    booktitle = CVPR,
    pages = {1--8},
    year = {2008},
}

@inproceedings{Wu2002,
    author= {Wu, Xiaodong and Chen, Danny Z},
    title= {Optimal net surface problems with applications},
    booktitle= ICALP,
    pages= {1029--1042},
    year= {2002},
}

@article{Li2005,
    author = {Li, Kang and Wu, Xiaodong and Chen, Danny Z and Sonka, Milan},
    title = {Optimal surface segmentation in volumetric images-a graph-theoretic approach},
    journal = PAMI,
    volume = {28},
    number = {1},
    pages = {119--134},
    year = {2005},
}

@book{Ford1962,
  author={Ford Jr, Lester Randolph and Fulkerson, Delbert Ray},
  title={Flows in networks},
  year={1962},
  publisher={Princeton university press}
}

@article{Goldberg1988,
  author = {Goldberg, Andrew V and Tarjan, Robert E},
  title = {A new approach to the maximum-flow problem},
  journal = JACM,
  volume = {35},
  number = {4},
  pages = {921--940},
  year = {1988},
}

@article{Chen2018,
  author = {Chen, Xinjian and Pan, Lingjiao},
  title = {A survey of graph cuts/graph search based medical image segmentation},
  journal = RBME,
  volume = {11},
  pages = {112--124},
  year = {2018},
}

@article{Chandran2009,
  author = {Chandran, Bala G and Hochbaum, Dorit S},
  title = {A computational study of the pseudoflow and push-relabel algorithms for the maximum flow problem},
  journal = OR,
  volume = {57},
  number = {2},
  pages = {358--376},
  year = {2009},
}

@article{Peng2013,
  author = {Peng, Bo and Zhang, Lei and Zhang, David},
  title = {A survey of graph theoretical approaches to image segmentation},
  journal = PR,
  volume = {46},
  number = {3},
  pages = {1020--1038},
  year = {2013},
}

@article{Kashyap2017,
  title = {Accurate Fully Automated 4D Segmentation of Osteoarthritic Knee MRI},
  author = {Kashyap, S and Zhang, H and Sonka, M},
  journal = {Osteoarthritis and Cartilage},
  volume = {25},
  pages = {S227--S228},
  year = {2017},
}

@article{Yin2010,
  author = {Yin, Yin and Zhang, Xiangmin and Williams, Rachel and Wu, Xiaodong and Anderson, Donald D and Sonka, Milan},
  title = {LOGISMOS—layered optimal graph image segmentation of multiple objects and surfaces: cartilage segmentation in the knee joint},
  journal = TMI,
  volume = {29},
  number = {12},
  pages = {2023--2037},
  year = {2010},
}

@inproceedings{Lee2020,
  author={Lee, Kyungmoo and Sohn, Elliott H and Zhang, Honghai and Warren, Alexis K and Wahle, Andreas and Whitmore, S Scott and Sonka, Milan and Abr{\`a}moff, Michael D},
  title={Multiresolution LOGISMOS graph search for automated choroidal layer segmentation of 3D macular OCT scans},
  booktitle={Medical Imaging 2020: Image Processing},
  volume={11313},
  pages={113130B},
  year={2020},
  organization={International Society for Optics and Photonics}
}

@article{Ishikawa2003,
  title = {Exact optimization for Markov random fields with convex priors},
  author = {Ishikawa, Hiroshi},
  journal= PAMI,
  volume = {25},
  number = {10},
  pages = {1333--1336},
  year = {2003},
}

@inproceedings{Isack2017,
    author = {Isack, Hossam and Veksler, Olga and Oguz, Ipek and Sonka, Milan and Boykov, Yuri},
    title = {Efficient optimization for hierarchically-structured interacting segments (HINTS)},
    booktitle = CVPR,
    pages = {1445--1453},
    year = {2017}
}

@article{Kolmogorov2007,
  author = {Kolmogorov, Vladimir and Rother, Carsten},
  title = {Minimizing nonsubmodular functions with graph cuts-a review},
  journal = PAMI,
  volume = {29},
  number = {7},
  pages = {1274--1279},
  year = {2007},
}

@inproceedings{Rother2007,
  author = {Rother, Carsten and Kolmogorov, Vladimir and Lempitsky, Victor and Szummer, Martin},
  title = {Optimizing binary MRFs via extended roof duality},
  booktitle = CVPR,
  pages = {1--8},
  year = {2007},
}

@inproceedings{Guo2018,
  author = {Guo, Zhihui and Zhang, Ling and Lu, Le and Bagheri, Mohammadhadi and Summers, Ronald M and Sonka, Milan and Yao, Jianhua},
  title = {Deep LOGISMOS: deep learning graph-based 3D segmentation of pancreatic tumors on CT scans},
  booktitle = ISBI,
  pages = {1230--1233},
  year = {2018},
}

@inproceedings{Jeppesen2020,
  author = {Jeppesen, Niels and Christensen, Anders N and Dahl, Vedrana A and Dahl, Anders B},
  title = {Sparse Layered Graphs for Multi-Object Segmentation},
  booktitle = CVPR,
  pages = {12777--12785},
  year = {2020}
}

@inproceedings{Goldberg2008,
  author = {Goldberg, Andrew V},
  title = {The partial augment--relabel algorithm for the maximum flow problem},
  booktitle = ESA,
  pages = {466--477},
  year = {2008}
}

@inproceedings{Goldberg2009,
  author = {Goldberg, Andrew V},
  title = {Two-level push-relabel algorithm for the maximum flow problem},
  booktitle = {International Conference on Algorithmic Applications in Management},
  pages = {212--225},
  year = {2009}
}

@article{Cherkassky1997,
  title = {On implementing the push—relabel method for the maximum flow problem},
  author = {Cherkassky, Boris V and Goldberg, Andrew V},
  journal = {Algorithmica},
  volume = {19},
  number = {4},
  pages = {390--410},
  year = {1997}
}

@inproceedings{Baumstark2015,
  author = {Baumstark, Niklas and Blelloch, Guy and Shun, Julian},
  title = {Efficient implementation of a synchronous parallel push-relabel algorithm},
  booktitle = ESA,
  pages = {106--117},
  year = {2015}
}

@inproceedings{Arora2010,
  author = {Arora, Chetan and Banerjee, Subhashis and Kalra, Prem and Maheshwari, SN},
  title = {An efficient graph cut algorithm for computer vision problems},
  booktitle = ECCV,
  pages = {552--565},
  year = {2010}
}

@inproceedings{Jensen2020,
    author = {Jensen, Patrick M. and Dahl, Anders B. and Dahl, Vedrana A.},
    title = {Multi-Object Graph-Based Segmentation With Non-Overlapping Surfaces},
    booktitle = CVPRW,
    pages={976--977},
    year = {2020}
}

@book{clrs2009,
  title={Introduction to algorithms},
  author={Cormen, Thomas H and Leiserson, Charles E and Rivest, Ronald L and Stein, Clifford},
  year={2009},
  publisher={MIT press}
}

@article{Anderson1995,
  title = {A parallel implementation of the push-relabel algorithm for the maximum flow problem},
  author = {Anderson, Richard and Setubal, Joao C.},
  journal = JPDC,
  volume = {29},
  number = {1},
  pages = {17--26},
  year = {1995}
}

@techreport{Bader2006,
  title={A cache-aware parallel implementation of the push-relabel network flow algorithm and experimental evaluation of the gap relabeling heuristic},
  author={Bader, David A and Sachdeva, Vipin},
  year={2006},
  institution={Georgia Institute of Technology}
}

@article{Goldberg1991,
  title = {Processor-efficient implementation of a maximum flow algorithm},
  author = {Goldberg, Andrew V},
  journal = IPL,
  volume = {38},
  number = {4},
  pages = {179--185},
  year = {1991}
}

@article{Hong2010,
  title = {An asynchronous multithreaded algorithm for the maximum network flow problem with nonblocking global relabeling heuristic},
  author = {Hong, Bo and He, Zhengyu},
  journal = TPDS,
  volume = {22},
  number = {6},
  pages = {1025--1033},
  year = {2010}
}

@book{Boyd2004,
	title={Convex optimization},
	author={Boyd, Stephen and Vandenberghe, Lieven},
	year={2004},
	publisher={Cambridge University Press}
}
}


\begin{IEEEbiography}[{\includegraphics[width=1in,height=1.25in,clip,keepaspectratio]{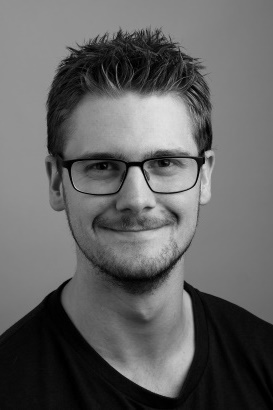}}]{Patrick M. Jensen}
was born in Copenhagen, Denmark, in 1994. He received his B.Sc.Eng degree in 2017 and M.Sc.Eng degree in 2019, both in applied mathematics, at the Technical University of Denmark (DTU), Kgs. Lyngby, Denmark. He is currently pursuing a Ph.D. in 3D image analysis at the Visual Computing group at the Department of Applied Mathematics and Computer Science, Technical University of Denmark. His research interests lie in 3D image segmentation.
\end{IEEEbiography}

\begin{IEEEbiography}[{\includegraphics[width=1in,height=1.25in,clip,keepaspectratio]{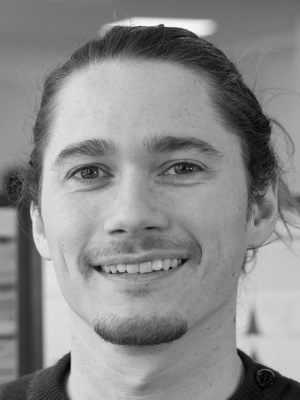}}]{Niels Jeppesen}
is an image analysis and machine learning specialist at FORCE Technology with a Ph.D. degree in image analysis of 3D structures from the Department of Applied Mathematics and Computer Science, Technical University of Denmark (DTU), Kgs. Lyngby, Denmark. His research interests lie in \maxflow algorithms and quantitative analysis of structures in 3D images. He applies these methods for automated quality control of structures and materials, in particular, in the wind turbine industry.
\end{IEEEbiography}

\begin{IEEEbiography}[{\includegraphics[width=1in,height=1.25in,clip,keepaspectratio]{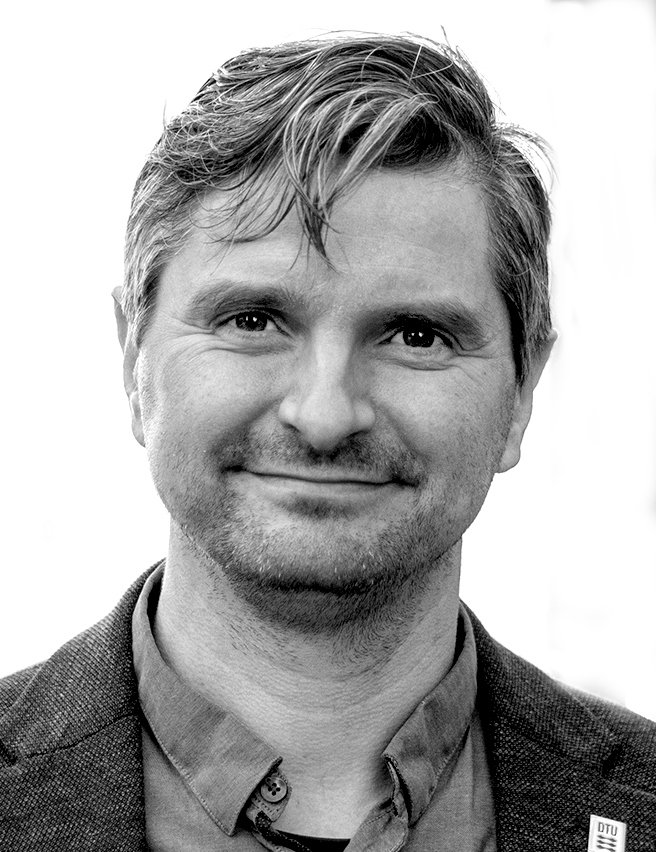}}]{Anders Bjorholm Dahl}
is professor in 3D image analysis, and a head of the Section for Visual Computing at the Department of Applied Mathematics and Computer Science, Technical University of Denmark (DTU), Kgs. Lyngby, Denmark. He is heading The Center for Quantification of Imaging Data from MAX IV, focusing on quantitative analysis of 3D images. His research is focused on image segmentation and its applications.
\end{IEEEbiography}

\begin{IEEEbiography}[{\includegraphics[width=1in,height=1.25in,clip,keepaspectratio]{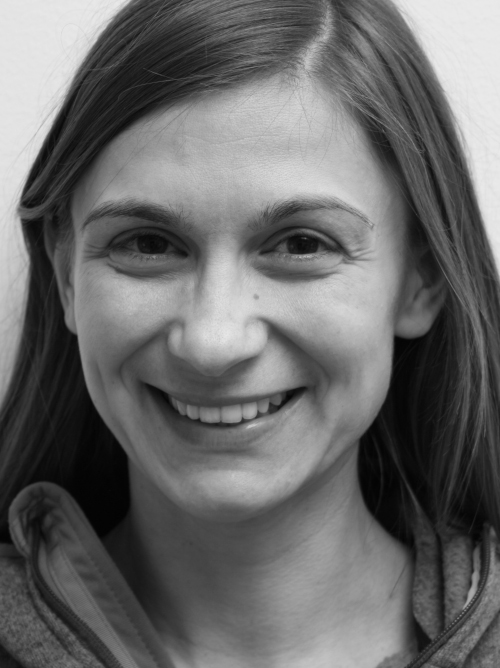}}]{Vedrana Andersen Dahl}
is an associate professor at the Department of Applied Mathematics and Computer Science, Technical University of Denmark (DTU), Kgs. Lyngby, Denmark. 
Her primary research interest is in the use of geometric models for the analysis of volumetric data. This includes volumetric segmentation and methods based on deformable meshes. She developed analysis tools with applications in material science, industrial inspection, and biomedicine.

\end{IEEEbiography}

\vfill

\end{document}